Department of Mathematics
TUM School of Computation, Information and Technology
Technical University of Munich

TUM

# Context Collapse

In-Context Learning and Model Collapse

## Josef Ott

Thesis for the attainment of the academic degree

**Master of Science**

at the TUM School of Computation, Information and Technology of the Technical University of Munich

**Examiner:**
Prof. Dr. Suvrit Sra

**Supervisors:**
Prof. Dr. Suvrit Sra
Dr. Manish Krishan Lal

**Submitted:**
Munich, 15. July 2025

I hereby declare that this thesis is entirely the result of my own work except where otherwise indicated. I have only used the resources given in the list of references. In addition, I used language tools, including Grammarly and large language models (LLMs) such as ChatGPT, to assist with editing and improving the clarity of the text. All intellectual contributions, ideas, and analyses are my own unless explicitly referenced.

Munich, 15. July 2025                                  Josef Ott

*Für meine Eltern – für all eure Unterstützung, Liebe und Geduld.*

# Zusammenfassung

Diese Arbeit untersucht zwei zentrale Phänomene in großen Sprachmodellen (LLMs): **In-Context Learning (ICL)** und **Model Collapse**.

Wir analysieren ICL anhand eines linearen Transformers mit *geteilten Gewichten (weight tying)*, der auf lineare Regressionsaufgaben trainiert wird, und zeigen, dass das Minimieren des In-Context-Loss zu einem Phasenübergang in den gelernten Parametern führt. Überschreitet die Kontextlänge einen kritischen Schwellenwert, entwickelt die Lösung eine *schiefsymmetrische Komponente*. Dies zeigen wir, indem wir den Forward Pass des linearen transformers unter Gewichtsteilung auf eine Form des *vorkonditionierten Gradientenabstiegs* zurückführen und den optimalen Vorkonditionierer analysieren. Dieser enthält eine schiefsymmetrische Komponente, die eine Rotation der Gradientenrichtung bewirkt.

Für das Phänomen des *Model Collapse* verwenden wir Martingal- und Random-Walk-Theorie, um vereinfachte Szenarien—lineare Regression und Schätzung einer Gaußschen Verteilung—unter sowohl *ersetzenden* als auch *kumulativen* Datenregimen zu analysieren. Wir erweitern bestehende Ergebnisse, indem wir *fast sichere Konvergenz* beweisen und zeigen, dass ein Kollaps eintritt, sofern die Datenmenge nicht schnell genug wächst oder langfristig erhalten bleibt.

Abschließend führen wir den Begriff des *Context Collapse* ein: eine Verschlechterung des Kontexts bei langen Generierungsprozessen, insbesondere im Chain-of-Thought-Reasoning. Dieses Konzept verbindet die Dynamik von ICL mit langfristigen Stabilitätsproblemen generativer Modelle.

# Abstract

This thesis investigates two key phenomena in large language models (LLMs): **in-context learning (ICL)** and **model collapse**.

We study ICL in a linear transformer with tied weights trained on linear regression tasks, and show that minimising the in-context loss leads to a phase transition in the learned parameters. Above a critical context length, the solution develops a *skew-symmetric component*. We prove this by reducing the forward pass of the linear transformer under weight tying to *preconditioned gradient descent*, and then analysing the optimal preconditioner. This preconditioner includes a skew-symmetric component, which induces a rotation of the gradient direction.

For *model collapse*, we use martingale and random walk theory to analyse simplified settings—linear regression and Gaussian fitting—under both *replacing* and *cumulative* data regimes. We strengthen existing results by proving *almost sure convergence*, showing that collapse occurs unless the data grows sufficiently fast or is retained over time.

Finally, we introduce the notion of *context collapse*: a degradation of context during long generations, especially in chain-of-thought reasoning. This concept links the dynamics of ICL with long-term stability challenges in generative models.

# Contents

# 1 Introduction

## 1.1 Background

### 1.1.1 Language models and their objective

A *language model* is a probability distribution over sequences of tokens. A *token* is a discrete unit of text, typically a word, subword, or character, depending on the tokenisation scheme used. For example, in subword tokenisation (such as Byte Pair Encoding [Gag94]), common word fragments like un, der, and stand may each be treated as separate tokens, allowing the model to efficiently handle rare or compound words.

Given a sequence of tokens $x_1, x_2, \ldots, x_L$, a language model assigns a probability $p(x_1, x_2, \ldots, x_L)$ that shows how likely that sequence is in natural language. These models are often autoregressive and model the joint probability using the chain rule

$$p(x_1, x_2, \ldots, x_L) = \prod_{i=1}^{L} p(x_i \mid x_1, \ldots, x_{i-1}).$$

This allows the model to predict the next token based on previous tokens and generate text by sampling one token at a time from these conditional distributions.

Early language models were based on *n-grams* [Kat87], which use a Markov assumption to predict each token based only on the preceding $n - 1$ tokens. Although computationally simple, these models are statistically inefficient due to data sparsity and an inability to capture long-range dependencies. Increasing $n$ leads to a combinatorial explosion in the number of parameters, exacerbating the problem.

### 1.1.2 Neural language models

To address the limitations of n-gram models, Bengio et al. [Ben+03] proposed *neural probabilistic language models*, which learn distributed representations of tokens (embeddings) and use them as inputs to a neural network that predicts the next token. Formally, each token $x_i$ is mapped to an embedding vector $e_i \in \mathbb{R}^d$, and the conditional probability $p(x_i \mid x_{i-n+1}, \ldots, x_{i-1})$ is computed via a neural network (NN)

$$p(x_i \mid x_{i-n+1}, \ldots, x_{i-1}) = \text{softmax}(\text{NN}(e_{i-n+1}, \ldots, e_{i-1})).$$

This approach combats the curse of dimensionality and enables generalisation to unseen sequences. However, early neural models used feedforward architectures and fixed-size contexts, limiting their ability to capture sequential structure.

### 1.1.3 Recurrent neural networks and LSTMs

*Recurrent Neural Networks* (RNNs) [Elm90; Mik+10] improved upon earlier models by maintaining a hidden state that is updated as each new token is processed. At each time step $t$, the hidden state $h_t \in \mathbb{R}^d$ is computed as

$$h_t = \phi(W_e e_t + W_h h_{t-1}),$$

where $e_t \in \mathbb{R}^d$ is the embedding of the current token $x_t$, $W_e \in \mathbb{R}^{d \times d}$ is the input weight matrix, $W_h \in \mathbb{R}^{d \times d}$ is the recurrent weight matrix, and $\phi$ is a nonlinear activation function (typically tanh or ReLU). This recurrent update allows RNNs to maintain a form of memory, giving them a theoretically unbounded context window to capture sequential dependencies in the input.

Standard RNNs struggle with long-term dependencies due to vanishing and exploding gradients. The *Long Short-Term Memory* (LSTM) network, introduced by Hochreiter and Schmidhuber [HS97], introduced gating mechanisms that allow the network to retain information over longer time horizons. Despite their gating mechanisms, LSTMs still compress all contextual information into a single hidden state and process inputs sequentially, resulting in two key limitations. First, the *hidden-state bottleneck* makes it challenging to represent fine-grained dependencies over long sequences, since LSTMs encode all relevant information into a fixed-size vector. Second, the inherently sequential nature of recurrence impedes parallelisation, making training and inference less efficient on modern hardware.

### 1.1.4 The attention mechanism

To overcome the hidden-state bottleneck and the strictly sequential computation of gated RNNs, Bahdanau, Cho, and Bengio [BCB16] introduced the *attention mechanism*. By computing content-based weights over the entire input sequence, attention enables the model to access any position directly (at constant distance). It supports highly parallel matrix operations, thereby addressing the representational and computational limitations of recurrent architectures [Vas+23].

Given query, key, and value matrices $Q$, $K$, and $V$, and the key and query dimension $d_k$, the *scaled dot-product attention* used in modern architectures [Vas+23] is defined as

$$\text{Attention}(Q, K, V) = \text{softmax}\left(\frac{QK^\top}{\sqrt{d_k}}\right)V,$$

which assigns higher weights to input positions more relevant to the current query. The scaling factor $\sqrt{d_k}$ prevents the dot products from becoming too large, stabilising the softmax. The attention mechanism eliminates the need to compress the entire sequence into a single fixed-size vector and forms the computational core of transformer models.

### 1.1.5 The transformer architecture

The 2017 paper *Attention Is All You Need* by Vaswani et al. [Vas+23] introduced the *transformer*, a neural architecture based entirely on attention mechanisms and without recurrence. The transformer uses *self-attention*, where each token attends to every other token in the same sequence, allowing for efficient parallelisation and better modelling of long-range dependencies.

As shown in Fig. 1.1, the full transformer consists of an encoder on the left and a decoder on the right. The encoder processes the input using self-attention and passes its output to the decoder. It applies masked self-attention (to handle autoregressive generation) and cross-attention over the encoder's outputs. The transformer layer consists of:

- Multi-head self-attention,

- Position-wise feedforward networks,

- Residual connections and layer normalisation,

- Positional encodings to inject sequence order information.

Each self-attention head computes attention using distinct learned projections of $Q$, $K$, and $V$. Outputs from all heads are concatenated and passed through a linear layer.

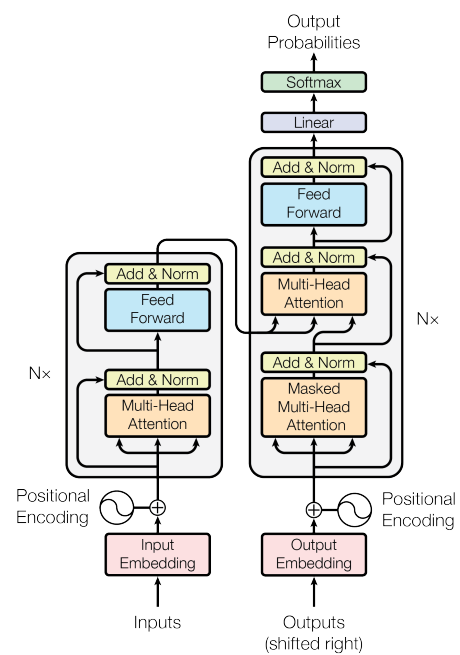

**Figure 1.1** The transformer architecture from Vaswani et al. [Vas+23].

The transformer rapidly became the foundation of modern *large language models* (LLMs). For instance, BERT (bidirectional encoder representations from transformers) [Dev+19] used a transformer encoder for language understanding, while GPT (generative pre-trained transformer) models [Bro+20; Rad+18] use autoregressive transformer decoders for generation. These models scale to billions of parameters and are trained on large corpora to perform tasks ranging from translation to reasoning.

## 1.2 Motivation

### 1.2.1 The emergence of in-context learning in language models

One of the most surprising capabilities of modern LLMs is *in-context learning* (ICL)—the ability to perform new tasks at inference time purely from textual prompts containing a few examples, without any parameter updates. This phenomenon was first widely observed with GPT-3 [Bro+20], demonstrating strong few-shot performance on diverse natural language processing (NLP) tasks such as translation, question answering, and arithmetic operations. In some cases, GPT-3 matched models that had been fine-tuned on thousands of supervised examples.

These results prompted significant interest in understanding how ICL works. Early interpretations suggested that during pretraining on massive text corpora, transformer models internalise patterns across a wide range of tasks, and can re-use this knowledge when given task demonstrations in context. Xie et al. [Xie+22] offered a Bayesian framing, modelling ICL as inference over a latent task concept. Empirical studies such as [Min+22] further showed that the format and surface-level structure of prompts are often more critical than the examples' correctness.

Mechanistic studies have provided further insight. Elhage et al. [Elh+21] identified specialised *induction heads*—attention heads that implement a copying mechanism across the context window. These circuits allow the model to continue patterns seen earlier in the prompt. Further work has shown that transformers can simulate standard learning algorithms like gradient descent purely via their forward pass [Aky+23], cementing ICL as a real and general-purpose capability of transformer models.

Crucially, practical techniques like *Chain-of-Thought* (CoT) prompting [Wei+23] have leveraged this understanding to significantly enhance LLM performance on complex reasoning tasks. By including intermediate reasoning steps in the prompt, models like PaLM achieved state-of-the-art performance on benchmarks such as GSM8K [Cob+21], validating that LLMs can generalise reasoning patterns from demonstrations. The development and success of such techniques underscore the reality and importance of ICL in modern language modelling.

### 1.2.2 The reality of model collapse in generative AI

As the use of generative models proliferates, a new challenge has come into focus: *model collapse.* This phenomenon occurs when models are trained or fine-tuned on datasets that increasingly contain content generated by other models, rather than fresh human-created data. The term was formalised by Shumailov et al. [Shu+24a], who demonstrated that iterative training on synthetic data leads to a degenerative process where the model loses diversity, accuracy, and robustness. They frame this concern with a striking question:

**What happens to GPT versions GPT-{$n$} as generation $n$ increases?**

Empirical studies showed that model collapse unfolds in stages. In the early phase, models begin to under-represent rare patterns and long-tail information, even while average performance appears stable. Over time, successive generations of models trained on synthetic data experience a narrowing of their output distribution, leading to reduced variance and factual correctness. In extreme cases, models exhibit non-sensical behaviour and forget previously mastered capabilities.

This collapse does not result from catastrophic forgetting or adversarial attacks but emerges from the statistical feedback loop inherent in training on model-generated data. As such, it presents a unique and pressing threat to the long-term viability of generative modelling, especially if future models rely on polluted data sources.

## 1.3 Contributions

This thesis contributes to two distinct but related areas: ICL dynamics in weight-tied transformers and the theory of model collapse in simplified settings. We outline them below.

**Contributions for ICL**   We consider a *weight-tied* linear transformer with $L = d = 2$, meaning it has two layers and operates in dimension two. When trained on the in-context loss $f_I$, the global minimisers of the loss function (see Theorem 2.5.7) display a distinct phase transition as the context length $n$ varies. Specifically, all minimisers are diagonal matrices for $n < 15$, reflecting purely scaling behaviour. However, once $n \geq 15$, the structure of the minimisers changes qualitatively: they acquire a nonzero skew-symmetric component, indicating the emergence of rotational dynamics in the learned transformation. We provide empirical support in Section 2.5.3. To quantify this transition, we fit a predictive model for the critical context length $n_{\mathrm{crit}}(d, L)$ at which skew symmetry first emerges, as a function of $d$ and $L$. The fitted model takes the form

$$n_{\mathrm{crit}}(d, L) = 6.95 + 0.45\,d + 12.67 \cdot \frac{d^{1.44}}{L^{2.43}}.$$

This is, to our knowledge, the first theoretical result to demonstrate the emergence of skew symmetry in optimal solutions due to weight tying, an effect not captured in prior analyses. On a more classical note, we further show that the same skew-symmetric component arises as the optimal preconditioner for gradient descent in linear regression.

**Contributions for model collapse**   We apply martingale and random walk theory to the most common proxies for model collapse—linear regression and Gaussian fitting. This framework strengthens existing results by upgrading convergence in expectation to almost sure convergence, providing a more rigorous characterisation of collapse dynamics. The *replace* scenario, where new data overwrites old data at every iteration, and the *accumulate* scenario, where data is incrementally aggregated, are analysed to investigate whether continual data accumulation mitigates model collapse. Furthermore, we observe that collapse is not limited to retraining on synthetic data, but can also occur during autoregressive text generation, particularly in multi-step reasoning tasks such as CoT prompting. This phenomenon, in which a model's internal context deteriorates as generation progresses, is referred to as *context collapse*.

# 2 In-context learning

## 2.1 What is in-context learning?

In-context learning (ICL) describes the phenomenon whereby LLMs exhibit the ability to learn from examples embedded within their input context, without undergoing any explicit parameter updates or gradient-based training [Bro+20]. When presented with a prompt containing several input-output pairs for a given task, an LLM can often generalise to a novel input from the same task and produce the appropriate output in a single forward pass.

This emergent capability—prominently observed in models like GPT-3[1] [Bro+20]—blurs the traditional boundary between training and inference. Rather than adapting its parameters, the model appears to infer a function from the input-output demonstrations and apply it to new data. Understanding the internal mechanism behind this behaviour has become a central question in machine learning theory.

## 2.2 Theoretical perspectives on in-context learning

Two main theoretical perspectives have emerged to explain how transformers achieve ICL, as identified by Dong et al. [Don+24]:

1. **Gradient Descent Emulation:** Transformers emulate gradient-based learning by internally performing updates analogous to gradient descent.

2. **Bayesian Inference:** Transformers implicitly perform posterior inference under a latent generative model. For example, Xie et al. [Xie+22] argue that transformers infer a latent task from in-context examples, while Zhang et al. [Zha+23] suggest that they predict by averaging over hypotheses consistent with the observed context.

While both perspectives offer valuable insights, our focus is on the first: the interpretation that transformers perform gradient-based optimisation during their forward pass. In particular, if each transformer layer executes a single gradient descent step, it becomes natural to study the weight-tied transformer, a model architecture in which all layers share the same weights.

**Transformers as gradient-based and closed-form learners**  Akyürek et al. [Aky+23] demonstrated that transformer models trained on linear regression tasks can emulate classical learning algorithms such as gradient descent and ridge regression. They provided constructive proofs and empirical evidence showing that transformer outputs align closely with these algorithms.

Furthermore, Oswald et al. [Osw+23] showed that transformers trained on in-context regression tasks approximate gradient descent and can learn to perform more advanced procedures such as iterative curvature corrections.

**Preconditioned gradient descent: a theoretical result**  Ahn et al. [Ahn+23] studied a simplified linear transformer trained on linear regression tasks. They showed that, at optimality, the transformer implements a single step of preconditioned gradient descent. The preconditioning matrix is learned and reflects the statistical structure of the task distribution, such as input covariances.

---

[1]ICL can be seen in different model architectures [TP25]

The authors further demonstrated that stacking multiple attention layers allows each layer to perform one gradient descent step. Thus, a transformer with $L$ layers can approximate $L$ steps of optimisation, providing a direct correspondence between network depth and iterative refinement.

## 2.3 Justification for using linear attention

Throughout the remainder of this chapter, we study the *linear attention* mechanism, later defined as

$$\text{Attn}_{P,Q}(Z) = PZM(Z^\top QZ),$$

where $Z \in \mathbb{R}^{(d+1)\times(n+1)}$ stacks $n+1$ context vectors column-wise, $P, Q \in \mathbb{R}^{(d+1)\times(d+1)}$ are learned projection matrices, and $M \in \mathbb{R}^{(n+1)\times(n+1)}$ is a diagonal masking matrix shared across positions. We adopt this formulation to better understand the optimisation view of ICL.

**Why we use linear attention** We study ICL in the setting of linear regression tasks, where linear attention is particularly well-suited. First, it is analytically tractable: the attention mechanism is purely linear, without softmax or nonlinearities. This simplicity enables us, as shown in Section 2.5.1, to explicitly compute the *in-context loss* without resorting to expectations. Empirically, linear attention also performs well on linear regression problems: a single linear attention head matches or outperforms softmax attention with two heads [Osw+23]. As demonstrated by Ahn et al. [Ahn+24], it trains faster and achieves lower loss.

Despite omitting softmax and feedforward layers, linear attention retains many essential behaviours observed in full transformers. It exhibits similar training dynamics—Adam outperforms SGD, gradient noise is heavy-tailed, and the loss landscape remains ill-conditioned [Ahn+24]. Its loss surface resembles standard transformers, and it remains expressive enough to learn and represent learning algorithms across layers, akin to softmax-based models [Ahn+23].

In summary, linear attention offers a simplified yet representative architecture. It preserves key optimisation characteristics of full transformers while enabling explicit analysis, making it a valuable proxy for studying ICL and transformer dynamics—the central focus of this chapter.

## 2.4 In-context learning via linear regression

To investigate the mechanisms by which transformers perform ICL, we adopt a controlled and analytically tractable setting: linear regression. This choice allows us to isolate the model's ability to internalise and apply an algorithm—specifically, gradient descent—over a distribution of tasks. We now formalise the components of this setup:

- The model is a multi-layer linear transformer equipped with residual connections and a simplified attention mechanism.

- The data consists of input-output pairs drawn from a linear model.

- The loss measures how well the model predicts the label of a new input based on the context examples.

This minimal setting captures the essence of ICL while enabling exact theoretical analysis. It follows the approach of prior work [Ahn+23; Aky+23; Gar+23; MHM23; Osw+23; ZFB23], and serves as a foundation for our theoretical contributions.

### 2.4.1 Linear transformer model, data, and loss

We adopt the notation used by Ahn et al. [Ahn+23].

**Definition 2.4.1** (Linear regression data). Let $D_\mathcal{X}$ be a distribution over $\mathbb{R}^d$, and $D_\mathcal{W}$ a distribution over weights in $\mathbb{R}^d$. Draw covariates $x^{(i)} \sim_{\text{i.i.d.}} D_\mathcal{X}$ for $i = 1, \ldots, n+1$, and independently draw a ground-truth weight vector $w_* \sim D_\mathcal{W}$. Define the covariate and response matrices as

$$X \coloneqq \begin{bmatrix} x^{(1)} & \cdots & x^{(n+1)} \end{bmatrix} \in \mathbb{R}^{d \times (n+1)} \quad \text{and} \quad Y \coloneqq \begin{bmatrix} y^{(1)} & \cdots & y^{(n+1)} \end{bmatrix} \coloneqq w_*^\top X \in \mathbb{R}^{1 \times (n+1)}.$$

The input matrix $Z_0 \in \mathbb{R}^{(d+1) \times (n+1)}$ is given by

$$Z_0 \coloneqq \begin{bmatrix} x^{(1)} & \cdots & x^{(n)} & x^{(n+1)} \\ y^{(1)} & \cdots & y^{(n)} & 0 \end{bmatrix},$$

where the final entry $y^{(n+1)}$ is masked with zero. The corresponding target is

$$y_{\text{test}} \coloneqq y^{(n+1)}. \tag{2.1}$$

The task is to predict the final response $y^{(n+1)}$ using the input matrix $Z_0$, which contains all covariates and responses except that $y^{(n+1)}$ is masked to zero. Each training example is thus a pair $(Z_0, y_{\text{test}})$, where the model must infer the held-out response from the preceding context.

**Definition 2.4.2** (Linear transformer model). Let $L, d, n \geq 1$, and let $\theta \coloneqq \{P_\ell, Q_\ell\}_{\ell=0}^{L-1}$ with $P_\ell, Q_\ell \in \mathbb{R}^{(d+1) \times (d+1)}$. The linear transformer is a map

$$\text{TF}_L^\theta : \mathbb{R}^{(d+1) \times (n+1)} \to \mathbb{R}, \quad Z_0 \mapsto -[Z_L]_{(d+1),(n+1)},$$

where the sequence $(Z_\ell)_{\ell=0}^L$ is defined recursively by

$$Z_{\ell+1} = Z_\ell + \frac{1}{n} \text{Attn}_{P_\ell, Q_\ell}(Z_\ell), \quad \text{for } \ell = 0, \ldots, L-1, \tag{2.2}$$

with

$$\text{Attn}_{P,Q}(Z) \coloneqq PZM(Z^\top Q Z), \quad M \coloneqq \begin{bmatrix} I_n & 0 \\ 0 & 0 \end{bmatrix}. \tag{2.3}$$

*Remark* 2.4.3. The matrix $M$ masks out the final row of the attention matrix $Z^\top Q Z$, preventing all tokens from attending to the $(n+1)$st. When desired, we write $\text{TF}_L(Z_0; \theta)$ to make the parameter dependence explicit.

Of particular interest is the case where each $P_\ell$ and $Q_\ell$ satisfies a sparsity constraint, for which, as shown in Lemma 2.4.9, the pass through the linear transformer corresponds to preconditioned gradient descent.

**Definition 2.4.4** (Sparsity structure). Let $A_\ell \in \mathbb{R}^{d \times d}$. We define

$$P_\ell = \begin{bmatrix} 0_{d \times d} & 0 \\ 0 & 1 \end{bmatrix}, \quad Q_\ell = -\begin{bmatrix} A_\ell & 0 \\ 0 & 0 \end{bmatrix}.$$

In this case, we write

$$\text{Attn}_{A_\ell}(Z) \coloneqq \text{Attn}_{P_\ell, Q_\ell}(Z)$$

and

$$\text{TF}_L\left(Z_0; \{A_\ell\}_{\ell=0}^{L-1}\right) \coloneqq \text{TF}_L\left(Z_0; \{P_\ell, Q_\ell\}_{\ell=0}^{L-1}\right).$$

Given a distribution over data pairs $(Z_0, y_{\text{test}})$, the ICL objective is to minimise the expected squared error between the true test label and the linear transformer's output. Concretely, the model is tasked with predicting

$$y_{\text{test}} = y^{(n+1)} = w_*^\top x^{(n+1)}$$

based on the matrix $Z_0$, which encodes the full sequence of covariates $x^{(1)}, \ldots, x^{(n+1)}$ and responses $y^{(1)}, \ldots, y^{(n)}$.

**Definition 2.4.5** (In-context loss). Let $\{P_\ell, Q_\ell\}_{\ell=0}^{L-1}$ denote the transformer parameters, and let $(Z_0, y_{\text{test}})$ be a pair of random variables. We define the in-context loss as the expected squared error

$$f_{(Z_0, y_{\text{test}})}\left(\{P_\ell, Q_\ell\}_{\ell=0}^{L-1}\right) \coloneqq \mathbb{E}_{(Z_0, y_{\text{test}})}\left[\left(\text{TF}_L\left(Z_0; \{P_\ell, Q_\ell\}_{\ell=0}^{L-1}\right) - y_{\text{test}}\right)^2\right].$$

*Remark* 2.4.6. In the sparse setting from Definition 2.4.4, the parameters are constrained via

$$P_\ell = \begin{bmatrix} 0 & 0 \\ 0 & 1 \end{bmatrix}, \quad Q_\ell = -\begin{bmatrix} A_\ell & 0 \\ 0 & 0 \end{bmatrix},$$

and we write the loss as $f_{(Z_0, y_{\text{test}})}\left(\{A_\ell\}_{\ell=0}^{L-1}\right)$. When the data pair $(Z_0, y_{\text{test}})$ is sampled according to the distributions $D_{\mathcal{X}}$ and $D_{\mathcal{W}}$ as in Definition 2.4.1, we write

$$f_{(D_{\mathcal{X}}, D_{\mathcal{W}})} \coloneqq f_{(Z_0, y_{\text{test}})}$$

as shorthand for the in-context loss evaluated on such samples. In the special case where $D_{\mathcal{X}} = \mathcal{N}(0, \Sigma)$ and $D_{\mathcal{W}} = \mathcal{N}(0, \Sigma^{-1})$, we further abbreviate

$$f_\Sigma \coloneqq f_{(D_{\mathcal{X}}, D_{\mathcal{W}})}.$$

We then consider training linear transformers, as defined in Definition 2.4.2, on the linear regression data from Definition 2.4.1 using the in-context loss introduced in Definition 2.4.5.

## 2.4.2 Weight tying

Empirically and theoretically (see Fact 2.4.12), the learned matrices $A_0, \ldots, A_{L-1}$ that minimize the in-context loss

$$f_\Sigma\left(\{A_\ell\}_{\ell=0}^{L-1}\right)$$

are all equal up to scalar multiplication. This suggests a natural simplification of the model by reusing the same matrix across all layers:

$$A_0 = A_1 = \cdots = A_{L-1}.$$

**Definition 2.4.7** (Weight tying). To formalise this, we define the *weight-tied linear transformer* as

$$\text{TF}_L(Z_0; A) \coloneqq \text{TF}_L\left(Z_0; \{A\}_{\ell=0}^{L-1}\right).$$

*Remark* 2.4.8. The in-context loss then simplifies to a function of a single matrix:

$$f_*(A) \coloneqq f_*(\{A\}_{\ell=0}^{L-1}),$$

where $*$ denotes any of the loss variants introduced previously, such as $f_{(Z_0, y_{\text{test}})}(A)$, $f_{(D_{\mathcal{X}}, D_{\mathcal{W}})}(A)$, or $f_\Sigma(A)$.

This weight-tied architecture is an instance of what is known in the literature as a *looped transformer* [Fan+25; Gat+24; Gia+23; Yan+24]. Later in this chapter, we will analyse the minimisers of the weight-tied in-context loss $f_\Sigma(A)$, examining how their structure depends on the parameters $d, L, n$ and the covariance matrix $\Sigma$.

## 2.4.3 Equivalence to preconditioned gradient descent

Consider solving a linear regression problem using preconditioned gradient descent. The empirical risk (loss function) is

$$R(w) \coloneqq \frac{1}{2n}\sum_{i=1}^{n}\left(w^\top x^{(i)} - y^{(i)}\right)^2,$$

where $w, x^{(i)} \in \mathbb{R}^d$ and $y^{(i)} \in \mathbb{R}$. Given a preconditioning matrix $A_\ell \in \mathbb{R}^{d \times d}$, a single gradient descent step updates the weights via

$$w \leftarrow w - A_\ell \nabla R(w).$$

We now evaluate the loss after performing this update:

$$
\begin{aligned}
R\left(w - A_\ell \nabla R(w)\right) &= \frac{1}{2n} \sum_{i=1}^{n} \left(\left(w - A_\ell \nabla R(w)\right)^\top x^{(i)} - y^{(i)}\right)^2 \\
&= \frac{1}{2n} \sum_{i=1}^{n} \left(w^\top x^{(i)} - \left(y^{(i)} + x^{(i)\top} A_\ell \nabla R(w)\right)\right)^2.
\end{aligned}
$$

This equation shows that instead of modifying the parameter vector $w$, one can equivalently interpret the update as modifying the labels $y^{(i)}$. In other words, the update shifts the error to the responses rather than the model. As we will show in Lemma 2.4.9, the linear transformer with sparse parameters operates in this manner: it does not compute explicit parameter updates but applies gradient updates directly to the response vector $Y_\ell$.

The result in Lemma 2.4.9 was first introduced by Ahn et al. [Ahn+23]. This work presents a generalisation of their original statement and a significantly simplified proof. Additionally, we note that both the original paper and several subsequent citations omit the transpose on $A_\ell$ in (2.5), which is required for correctness.

**Lemma 2.4.9** (Forward pass as preconditioned gradient descent). *Consider an $L$-layer linear transformer parametrised by $A_0, \ldots, A_{L-1}$, as defined in Definition 2.4.4. Let $Z_0$, $w_*$, and $x^{(1)}, \ldots, x^{(n+1)}$ be as in Definition 2.4.1, and let $Z_\ell$ denote the output of the $\ell$-th layer of the transformer, per Eq. (2.2). Let $X_\ell$ and $Y_\ell$ be the first $d$ rows and the last row of $Z_\ell$, respectively. Define the iterates of preconditioned gradient descent by:*

$$w_{\mathrm{gd}}^{(0)} = 0, \tag{2.4}$$

$$w_{\mathrm{gd}}^{(\ell+1)} = w_{\mathrm{gd}}^{(\ell)} - A_\ell^\top \nabla R_{w_*}\left(w_{\mathrm{gd}}^{(\ell)}\right) \quad \text{for } \ell \in \{0, \ldots, L-1\}, \tag{2.5}$$

*where the loss function with respect to the true weights $w_*$ is*

$$R_{w_*}(w) \coloneqq \frac{1}{2n} \sum_{i=1}^{n} \left(w^\top x^{(i)} - w_*^\top x^{(i)}\right)^2. \tag{2.6}$$

*Then, for all $\ell \in \{0, \ldots, L\}$, we have*

$$Y_\ell = Y_0 - \left(w_{\mathrm{gd}}^{(\ell)}\right)^\top X_0, \tag{2.7}$$

*and in particular,*

$$TF_L\left(Z_0; \{A_\ell\}_{\ell=0}^{L-1}\right) = \left\langle w_{\mathrm{gd}}^{(L)}, x^{(n+1)} \right\rangle. \tag{2.8}$$

*Proof.* See Appendix B.5.2. □

Using the characterisation of the linear transformer's output as the prediction of $y^{(n+1)}$ under preconditioned gradient descent (see Lemma 2.4.9), we can derive a simplified expression for the in-context loss $f$.

**Corollary 2.4.10** (In-context loss as prediction error of gradient descent). *The in-context loss defined in Definition 2.4.5 is equal to the expected squared prediction error incurred when using the final iterate $w_{\mathrm{gd}}^{(L)}$ from preconditioned gradient descent (as defined in Lemma 2.4.9) to predict the test label $y_{test} = w_*^\top x^{(n+1)}$. Formally,*

$$f_{(Z_0, y_{test})}\left(\{A_\ell\}_{\ell=0}^{L-1}\right) = \mathbb{E}\left[\left(\left\langle w_{\mathrm{gd}}^{(L)} - w_*, x^{(n+1)} \right\rangle\right)^2\right].$$

*Proof.* By (2.8), we have

$$\mathrm{TF}_L\left(Z_0; \{A_\ell\}\right) = \left\langle w_{\mathrm{gd}}^{(L)}, x^{(n+1)} \right\rangle,$$

and by (2.1), the true label is $y_{\mathrm{test}} = \langle w_*, x^{(n+1)} \rangle$. Therefore, the in-context loss becomes

$$f_{(Z_0, y_{\mathrm{test}})}\left(\{A_\ell\}\right) = \mathbb{E}\left[\left(\left\langle w_{\mathrm{gd}}^{(L)}, x^{(n+1)} \right\rangle - \left\langle w_*, x^{(n+1)} \right\rangle\right)^2\right],$$

which simplifies to

$$f_{(Z_0, y_{\mathrm{test}})}\left(\{A_\ell\}\right) = \mathbb{E}\left[\left(\left\langle w_{\mathrm{gd}}^{(L)} - w_*, x^{(n+1)} \right\rangle\right)^2\right].$$

□

### 2.4.4 Problem statement and simplification

We are interested in characterising the minimisers of the in-context loss functional $f_\Sigma$ in both the untied and weight-tied settings. While other authors previously studied the untied case, the weight-tied case remains largely unexplored. In particular, we are the first to observe the emergence of a skew-symmetric component in the optimal solution under weight tying.

**Problem 2.4.11** (Minimization without weight tying)*. Find the minimisers of the in-context loss functional:*

$$\arg\min_{\{A_\ell\}_{\ell=0}^{L-1}} f_\Sigma\left(\{A_\ell\}_{\ell=0}^{L-1}\right).$$

Ahn et al. [Ahn+23] addressed this problem. We restate their result below for completeness.

**Fact 2.4.12** (Near-stationarity under sparse parametrisation)*. Let $\mathcal{S} \subset \mathbb{R}^{L \times d \times d}$ denote the set of sequences $A = (A_0, \ldots, A_{L-1})$ such that for each $i = 0, \ldots, L-1$, there exists a scalar $a_i \in \mathbb{R}$ with*

$$A_i = a_i \Sigma^{-1}.$$

*Since $\Sigma$ is symmetric, every $A_i \in \mathcal{S}$ is symmetric as well. Then, the following holds:*

$$\inf_{A \in \mathcal{S}} \sum_{i=0}^{L-1} \left\|\nabla_{A_i} f(A)\right\|_{\mathrm{F}}^2 = 0,$$

*where $\nabla_{A_i} f$ denotes the gradient of $f$ with respect to $A_i$, and $\|\cdot\|_{\mathrm{F}}$ is the Frobenius norm.*

The result stated above, established initially by Ahn et al. [Ahn+23, Theorem 3] and supported by numerical experiments, shows that the learned weights $A_i$ tend to align with scalar multiples of $\Sigma^{-1}$, suggesting that the optimal transformer parameters approximately lie within the class $\mathcal{S}$.

We now turn to the weight-tying setting, where all layers share the same matrix. Our goal is to characterise the minimisers of the in-context loss in this constrained scenario.

**Problem 2.4.13** (Minimization with weight tying)*. Characterise the set of minimisers of the weight-tied in-context loss functional:*

$$\arg\min_A f_\Sigma\left(A\right).$$

Our results (see Theorem 2.5.7) show that, in contrast to the untied setting, the minimiser in the weight-tied case can exhibit a non-vanishing skew-symmetric component. Specifically, the optimal matrix $A$ may take the form

$$A = a\Sigma^{-1} + K,$$

where $a \in \mathbb{R}$ and $K \in \mathbb{R}^{d \times d}$ is skew-symmetric.

**Simplify Problem 2.4.13 by reducing it to $\Sigma = I$**

The results in this section are original, though inspired by the approach by Ahn et al. [Ahn+23].

A change of variables based on the Cholesky decomposition [Bor18, p. 29] of $\Sigma$ shows that we can reduce the problem of finding the minimisers of $f_\Sigma$ to the isotropic case $\Sigma = I$.

**Lemma 2.4.14.** *For any sequence of matrices $\{A_\ell\}_{\ell=0}^{L-1}$ with $A_\ell \in \mathbb{R}^{d \times d}$, and for any positive definite covariance matrix $\Sigma = \mathbf{L}\mathbf{L}^\top$, we have*

$$f_\Sigma\left(\{A_\ell\}_{\ell=0}^{L-1}\right) = f_I\left(\{\mathbf{L}^\top A_\ell \mathbf{L}\}_{\ell=0}^{L-1}\right).$$

*Proof.* The proof can be found in Appendix B.5.1. □

**Proposition 2.4.15.** *Let $\Sigma \in \mathbb{R}^{d \times d}$ be a covariance matrix, and let $\Sigma = \mathbf{L}\mathbf{L}^\top$ be its Cholesky decomposition. Assume that both $\arg\min f_I$ and $\arg\min f_\Sigma$ are not empty. Then*

$$\{\mathbf{L}^\top A_\ell \mathbf{L}\}_{\ell=0}^{L-1} \in \arg\min f_I \quad \text{if and only if} \quad \{A_\ell\}_{\ell=0}^{L-1} \in \arg\min f_\Sigma.$$

*Proof.* Let

$$\{\mathbf{L}^\top A_\ell^I \mathbf{L}\}_{\ell=0}^{L-1} \in \arg\min f_I \quad \text{and} \quad \{A_\ell^\Sigma\}_{\ell=0}^{L-1} \in \arg\min f_\Sigma.$$

From Lemma 2.4.14, we have

$$f_I\left(\{\mathbf{L}^\top A_\ell^\Sigma \mathbf{L}\}\right) = f_\Sigma\left(\{A_\ell^\Sigma\}\right) \le f_\Sigma\left(\{A_\ell^I\}\right) = f_I\left(\{\mathbf{L}^\top A_\ell^I \mathbf{L}\}\right).$$

Therefore,

$$\{\mathbf{L}^\top A_\ell^\Sigma \mathbf{L}\}_{\ell=0}^{L-1} \in \arg\min f_I.$$

Conversely, using the same logic:

$$f_\Sigma\left(\{A_\ell^I\}\right) = f_I\left(\{\mathbf{L}^\top A_\ell^I \mathbf{L}\}\right) \le f_I\left(\{\mathbf{L}^\top A_\ell^\Sigma \mathbf{L}\}\right) = f_\Sigma\left(\{A_\ell^\Sigma\}\right),$$

which implies

$$\{A_\ell^I\}_{\ell=0}^{L-1} \in \arg\min f_\Sigma. \qquad \square$$

**Simplifying the in-context loss via gradient descent dynamics**

The following original result enables us to express the in-context loss $f_I$ as an expectation over $X$, thereby eliminating dependence on $w_*$ and $x^{(n+1)}$, and significantly simplifying the analysis.

**Proposition 2.4.16** (Simplification of the in-context loss). *Let $A_0, \ldots, A_{L-1} \in \mathbb{R}^{d \times d}$. Then the in-context loss satisfies*

$$f_I\left(\{A_\ell\}_{\ell=0}^{L-1}\right) = \mathbb{E}_X \left\| \prod_{\ell=0}^{L-1} \left(I - \frac{1}{n} A_\ell^\top X X^\top\right) \right\|_F^2,$$

*where $X \in \mathbb{R}^{d \times n}$ has entries $X_{ij} \overset{iid}{\sim} \mathcal{N}(0, 1)$. In the case of weight tying, this further simplifies to*

$$f_I(A) = \mathbb{E}_X \left\| \left(I - \frac{1}{n} A^\top X X^\top\right)^L \right\|_F^2.$$

*Proof.* Let $w_{\text{gd}}^{(\ell)}$ be the gradient descent iterates defined in Lemma 2.4.9, via Eq. (2.4), Eq. (2.5), and define $R_{w_*}(w)$ as in Eq. (2.6). From Eq. (B.6), we have

$$\nabla R_{w_*}(w) = \frac{1}{n}\overline{X}_0 \overline{X}_0^\top (w - w_*),$$

where $\overline{X}_0 \in \mathbb{R}^{d \times n}$ is the matrix of the first $n$ covariates, i.e.,

$$\overline{X}_0 \coloneqq \begin{bmatrix} x^{(1)} & \cdots & x^{(n)} \end{bmatrix}.$$

This implies the recurrence

$$w_{\text{gd}}^{(\ell+1)} - w_* = \left( I - \frac{1}{n} A_\ell^\top \overline{X}_0 \overline{X}_0^\top \right) (w_{\text{gd}}^{(\ell)} - w_*).$$

Telescoping this recursion gives

$$w_{\text{gd}}^{(L)} - w_* = \prod_{\ell=0}^{L-1} \left( I - \frac{1}{n} A_\ell^\top \overline{X}_0 \overline{X}_0^\top \right) (w_{\text{gd}}^{(0)} - w_*) = -\prod_{\ell=0}^{L-1} \left( I - \frac{1}{n} A_\ell^\top \overline{X}_0 \overline{X}_0^\top \right) w_*,$$

where the last line follows from $w_{\text{gd}}^{(0)} = 0$ by Eq. (2.4). Substituting this into the in-context loss (cf. Corollary 2.4.10), we obtain

$$
\begin{aligned}
f_I \left( \{A_\ell\}_{\ell=0}^{L-1} \right) &= \mathbb{E} \left[ \left( \langle w_{\text{gd}}^{(L)} - w_*, x^{(n+1)} \rangle \right)^2 \right] \\
&= \mathbb{E}_{\overline{X}_0} \left[ \mathbb{E}_{w_*, x^{(n+1)}} \left[ \left\langle -\prod_{\ell=0}^{L-1} \left( I - \frac{1}{n} A_\ell^\top \overline{X}_0 \overline{X}_0^\top \right) w_*, x^{(n+1)} \right\rangle^2 \middle| \overline{X}_0 \right] \right] \\
&= \mathbb{E}_{\overline{X}_0} \left[ \mathbb{E}_{w_*, x^{(n+1)}} \left[ \left\langle \prod_{\ell=0}^{L-1} \left( I - \frac{1}{n} A_\ell^\top \overline{X}_0 \overline{X}_0^\top \right) w_*, x^{(n+1)} \right\rangle^2 \middle| \overline{X}_0 \right] \right].
\end{aligned}
\tag{2.9}
$$

Since $w_* \sim \mathcal{N}(0, I)$, $x^{(n+1)} \sim \mathcal{N}(0, I)$, and they are independent of $\overline{X}_0$, by Lemma B.4.5, the conditional inner product has expectation equal to the squared Frobenius norm:

$$\mathbb{E}_{w_*, x^{(n+1)}} \left[ \langle A w_*, x^{(n+1)} \rangle^2 \right] = \|A\|_{\text{F}}^2.$$

Applying this to Eq. (2.9) with $A = \prod_{\ell=0}^{L-1} \left( I - \frac{1}{n} A_\ell^\top \overline{X}_0 \overline{X}_0^\top \right)$ yields

$$f_I \left( \{A_\ell\}_{\ell=0}^{L-1} \right) = \mathbb{E}_{\overline{X}_0} \left\| \prod_{\ell=0}^{L-1} \left( I - \frac{1}{n} A_\ell^\top \overline{X}_0 \overline{X}_0^\top \right) \right\|_{\text{F}}^2,$$

as claimed. In the weight tying case, where $A_\ell = A$ for all $\ell$, the product becomes

$$\left( I - \frac{1}{n} A^\top X X^\top \right)^L,$$

which yields the stated simplification. $\qquad\square$

## 2.5  Emergence of skew-symmetric components in the weight-tied case

This section presents entirely original work. As demonstrated empirically in Section 2.5.3, determining the conditions under which skew-symmetric components emerge in the weight-tying setting is non-trivial. Such components do not appear universally; their emergence depends sensitively on the underlying parameter configuration.

To build analytical intuition, we examine a minimal working example with fixed parameters $d = L = 2$. In this simplified setting, we prove that for $n \geq 15$, the global minimiser of the in-context loss $f_I$ necessarily contains a nonzero skew-symmetric component. This result implies that the optimum lies outside the subspace of symmetric matrices—contrasting with the behaviour observed in the untied setting (see Fact 2.4.12). Numerical experiments further support this analytical insight at larger values of $d$ and $L$, consistently revealing the emergence of skew-symmetric components in minimisers of $f_I$.

Recall from Proposition 2.4.16 that in the weight-tied case, the in-context loss $f_I$ simplifies to

$$f_I(A) = \mathbb{E}_X \left[ \left\| \left( I - \frac{1}{n} A^\top X X^\top \right)^L \right\|_F^2 \right],$$

where $X \in \mathbb{R}^{d \times n}$ has i.i.d. entries from $\mathcal{N}(0, 1)$. Note that $W = XX^\top \sim \mathcal{W}_d(n, I_d)$ follows a Wishart distribution with $n$ degrees of freedom and identity scale matrix (see Definition A.1.1).

First, we leverage the orthogonal invariance of $f_I$ to restrict the search for stationary points to matrices whose symmetric part is diagonal.

**Lemma 2.5.1** (Orthogonal invariance of $f_I$). *Let $A \in \mathbb{R}^{d \times d}$. Then for any orthogonal matrix $Q \in \mathbb{R}^{d \times d}$ (i.e., $Q^\top Q = I$), the functional $f_I$ satisfies*

$$f_I(QAQ^\top) = f_I(A).$$

*Proof.* We use two classical invariance properties:

- The Frobenius norm is invariant under orthogonal conjugation: for any $M \in \mathbb{R}^{d \times d}$, we have $\|Q^\top M Q\|_F = \|M\|_F$.

- The Wishart distribution with identity scale matrix is invariant under orthogonal conjugation: if $W \sim \mathcal{W}_d(n, I_d)$, then $Q^\top W Q \sim \mathcal{W}_d(n, I_d)$ for any orthogonal matrix $Q$.

Let $W \sim \mathcal{W}_d(n, I_d)$. Then

$$f_I(QAQ^\top) = \mathbb{E}_W \left[ \left\| \left( I - \tfrac{1}{n}(QAQ^\top)^\top W \right)^L \right\|_F^2 \right] = \mathbb{E}_W \left[ \left\| \left( I - \tfrac{1}{n} QA^\top Q^\top W \right)^L \right\|_F^2 \right]$$

$$= \mathbb{E}_W \left[ \left\| Q \left( I - \tfrac{1}{n} A^\top Q^\top W Q \right)^L Q^\top \right\|_F^2 \right] = \mathbb{E}_W \left[ \left\| \left( I - \tfrac{1}{n} A^\top Q^\top W Q \right)^L \right\|_F^2 \right],$$

which equals $f_I(A)$ by the distributional invariance of $W$. □

**Lemma 2.5.2** (Reduction to diagonal symmetric part). *It suffices to consider matrices whose symmetric part is diagonal to find all stationary points of $f_I$. The complete set of stationary points is obtained by applying all orthogonal conjugations to this class of matrices.*

*Proof.* Let $A \in \mathbb{R}^{d \times d}$, and define its symmetric part as $S := \frac{1}{2}(A + A^\top)$. By the spectral theorem [Lan87, Ch. VIII, Thm. 4.4], there exists an orthogonal matrix $Q \in \mathbb{R}^{d \times d}$ such that

$$QSQ^\top = D,$$

where $D$ is diagonal. Define $A' := QAQ^\top$. Then the symmetric part of $A'$ is

$$\tfrac{1}{2}(A' + A'^\top) = Q \left( \tfrac{1}{2}(A + A^\top) \right) Q^\top = QSQ^\top = D,$$

which is diagonal. Let $\psi : \mathbb{R}^{d \times d} \to \mathbb{R}^{d \times d}$ be defined by $\psi(A) := QAQ^\top$. By the orthogonal invariance of $f_I$ (Lemma 2.5.1), we have

$$f_I \circ \psi = f_I,$$

which implies $\nabla(f_I \circ \psi)(A) = \nabla f_I(A)$. Calculating the left side yields, by application of the chain rule to $f_I \circ \psi$, for all $H$,

$$D(f_I \circ \psi)(A)[H] = Df_I(\psi(A))[D\psi(A)[H]] = Df_I(QAQ^\top)[QHQ^\top]$$
$$= \left\langle \nabla f_I(QAQ^\top), QHQ^\top \right\rangle = \left\langle Q^\top \nabla f_I(QAQ^\top) Q, H \right\rangle.$$

We used that the Fréchet derivative of $\psi$ at $A$ is $D\psi(A)[H] = QHQ^\top$, since $\psi$ is linear. Because this holds for all $H$, we identify

$$\nabla(f_I \circ \psi)(A) = Q^\top \nabla f_I(QAQ^\top)Q.$$

Thus, $\nabla f_I(A) = 0$ if and only if $\nabla f_I(A') = 0$. Therefore, without loss of generality, it suffices to study stationary points $A$ for which the symmetric part is diagonal. Moreover, the complete set of stationary points is given by

$$\left\{ QAQ^\top : A \in \mathcal{SP}(f_I), \text{ with } \frac{1}{2}(A + A^\top) \text{ diagonal, } Q \in O(d) \right\},$$

where $\mathcal{SP}(f_I) := \{A \in \mathbb{R}^{d \times d} : \nabla f_I(A) = 0\}$, and $O(d)$ is the set of orthogonal matrices of order $d$.     □

We next introduce a simplified reparametrisation of $f_I$ to facilitate the discussion.

**Lemma 2.5.3** (Reparametrised functional). *Define the reparametrised functional*

$$f(A) := f_I(-nA^\top) = \mathbb{E}_W \left[ \left\| (I + AW)^L \right\|_F^2 \right],$$

*where $W \sim \mathcal{W}_d(n, I_d)$. Then a matrix $A \in \mathbb{R}^{d \times d}$ is a stationary point (respectively, minimizer) of $f$ if and only if $-nA^\top$ is a stationary point (respectively, minimizer) of $f_I$.*

*Proof.* Let $\varphi : \mathbb{R}^{d \times d} \to \mathbb{R}^{d \times d}$ be the smooth bijection defined by $\varphi(A) = -nA^\top$, so $f = f_I \circ \varphi$. Since the Wishart distribution $\mathcal{W}_d(n, I_d)$ has finite moments of all orders, $f$ and $f_I$ are polynomials in $n$ and the entries of $A$ and therefore smooth. As $\varphi$ is linear, its Fréchet derivative at any point $A$ is given by $D\varphi(A)[H] = -nH^\top$ for any direction $H \in \mathbb{R}^{d \times d}$. Applying the chain rule to $f = f_I \circ \varphi$, we compute for all $H$,

$$Df(A)[H] = Df_I(\varphi(A))[D\varphi(A)[H]] = Df_I(-nA^\top)[-nH^\top]$$
$$= \left\langle \nabla f_I(-nA^\top), -nH^\top \right\rangle = \left\langle -n\nabla f_I(-nA^\top)^\top, H \right\rangle.$$

Since this holds for all $H$, we identify $\nabla f(A) = -n\nabla f_I(-nA^\top)^\top$, so $\nabla f(A) = 0$ if and only if $\nabla f_I(-nA^\top) = 0$, and stationary points correspond. To show the correspondence of minima, suppose $A^\star$ is a local minimizer of $f$, so there exists $\varepsilon > 0$ such that $f(A) \geq f(A^\star)$ for all $A$ with $\|A - A^\star\|_F < \varepsilon$. Then since $\varphi$ is a diffeomorphism, the image $\varphi(B_\varepsilon(A^\star))$ is an open neighbourhood of $B^\star := \varphi(A^\star) = -nA^{\star\top}$, and for all $B$ in this neighbourhood we have

$$f_I(B) = f(\varphi^{-1}(B)) \geq f(A^\star) = f_I(B^\star),$$

so $B^\star$ is a local minimizer of $f_I$. The converse holds similarly using $\varphi^{-1}$. Because $\varphi$ is a bijection, global minima are also preserved: $A^\star$ minimizes $f$ globally if and only if $-nA^{\star\top}$ minimizes $f_I$ globally. This completes the proof.     □

### 2.5.1 Analytic case: $L = d = 2$

Determining when a nonzero skew-symmetric component arises in minimisers of $f_I$ is generally a nontrivial problem. To gain insight, we construct a minimal working example by analysing the simplest nontrivial parameter setting in which skew-symmetry can occur.

For $d = 1$, the matrix $A \in \mathbb{R}^{1 \times 1}$ is scalar, so skew-symmetric components are not defined. Moreover, as shown in Section 2.5.2, no skew-symmetric component arises when $L = 1$. Thus, we consider the first nontrivial case where skew-symmetry can emerge: $L = d = 2$.

In this setting, we study how the behaviour of minimisers depends on $n$, the number of in-context examples. Specifically, we seek to identify whether and for which values of $n$ a skew-symmetric component appears in the minimiser of $f_I$. For these parameters, the in-context loss takes the form

$$f_I(A) = \mathbb{E}_X \left[ \left\| \left( I - \frac{1}{n}A^\top W \right)^2 \right\|_F^2 \right],$$

where $W \sim \mathcal{W}_2(n, I_2)$ is a $2 \times 2$ Wishart-distributed random matrix. All symbolic calculations used in this chapter can be found in Appendix B.1.

**Stationary point analysis**

We begin by decomposing $A$ into its symmetric and skew-symmetric parts: $A = S + K$. By Lemma 2.5.2, we may assume without loss of generality that $S$ is diagonal, and subsequently recover the complete set of solutions via orthogonal conjugation. This yields

$$S = \tfrac{1}{2}(A + A^\top) = \begin{bmatrix} a_{11} & 0 \\ 0 & a_{22} \end{bmatrix}, \quad K = \tfrac{1}{2}(A - A^\top) = \begin{bmatrix} 0 & k \\ -k & 0 \end{bmatrix},$$

so that

$$A = \begin{bmatrix} a_{11} & k \\ -k & a_{22} \end{bmatrix}, \quad \text{with } a_{11}, a_{22}, k \in \mathbb{R}.$$

To simplify the analysis, we search for stationary points of the reparametrised functional $f(A) = f_I(-nA)$. As established in Lemma 2.5.3, the stationary points of $f_I$ correspond exactly to $-n$ times the stationary points of $f$.

The functional $f(A)$ is a degree-four (quartic) polynomial in the variables $a_{11}, a_{22}$, and $k$, and satisfies $f(A) \geq 0$ for all $A \in \mathbb{R}^{2 \times 2}$. By expanding the expression for $f(A)$ and symbolically evaluating[2] the expectation, we obtain an explicit polynomial representation in terms of $a_{11}, a_{22}$, and $k$.

To identify stationary points, we compute the gradient of $f$, i.e., the partial derivatives with respect to the three parameters:

$$g_1 := \frac{\partial f}{\partial a_{11}}, \quad g_2 := \frac{\partial f}{\partial a_{22}}, \quad g_3 := \frac{\partial f}{\partial k}.$$

As a first simplification, we can show that any stationary point satisfies $a_{11} = a_{22}$, thereby reducing the number of free parameters.

**Lemma 2.5.4.** *At stationary points of $f_I$, the symmetric part of the matrix is a scalar multiple of the identity.*

*Proof.* To characterise the structure of stationary points, we analyse the condition under which the gradient of $f$ vanishes. Consider the difference between the first two partial derivatives:

$$g_1 - g_2 = 4n(a_{11} - a_{22}) \cdot P_n(a_{11}, a_{22}, k),$$

where $P_n$ is a second-degree polynomial in $a_{11}, a_{22}$, and $k$, with coefficients depending on $n$. Assume, for contradiction, that $a_{11} \neq a_{22}$. Then the vanishing of the gradient implies $P_n(a_{11}, a_{22}, k) = 0$. We now examine the third component of the gradient:

$$g_3 = 8kn \cdot Q_n(a_{11}, a_{22}, k),$$

where $Q_n$ is another quadratic polynomial. Suppose $k \neq 0$. Then $Q_n(a_{11}, a_{22}, k) = 0$ must also hold. Consider the sum of the two polynomials,

$$T_n(a_{11}, a_{22}, k) := P_n(a_{11}, a_{22}, k) + Q_n(a_{11}, a_{22}, k).$$

Symbolic computation reveals that $T_n$ is a quadratic polynomial with strictly positive coefficients. In particular, its minimum value satisfies

$$\min T_n = 6 > 0,$$

independent of $n$. Therefore, $T_n(a_{11}, a_{22}, k) \neq 0$ for all $a_{11}, a_{22}, k \in \mathbb{R}$, contradicting the assumption that both $P_n = 0$ and $Q_n = 0$ simultaneously. We conclude that $k = 0$, i.e., the skew-symmetric component vanishes. Returning to the expression for $g_1 - g_2$, we now analyse the case $k = 0$. In this setting, the polynomial simplifies to

$$P_{n,k=0}(a_{11}, a_{22}) := P_n(a_{11}, a_{22}, 0),$$

---

[2]The code for symbolic evaluation can be found in Appendix B.1.

which again has strictly positive coefficients. Symbolic minimisation yields

$$\min P_{n,k=0} = \frac{(n+3)(21n+67)}{(n+7)(3n+11)} > 0 \quad \text{for all } n \in \mathbb{N},$$

so $P_{n,k=0}(a_{11}, a_{22}) \neq 0$, contradicting the assumption $a_{11} \neq a_{22}$. Therefore, we conclude that all stationary points satisfy $a_{11} = a_{22}$, i.e., the symmetric part of $A$ is a scalar multiple of the identity. □

Let $x := a_{11} = a_{22}$. Then the matrix $A$ has the form

$$A = \begin{bmatrix} x & k \\ -k & x \end{bmatrix},$$

and the gradient of $f(A)$ can be expressed in terms of the variables $x$ and $k$. Define

$$g_x := \frac{\partial f}{\partial x}, \quad g_k := \frac{\partial f}{\partial k}.$$

**Lemma 2.5.5** (Diagonal stationary point of $f_I$ for $d = L = 2$). *For every $n \in \mathbb{N}$, the in-context loss $f_I$ admits a stationary point at the diagonal matrix $A = -nx^*_{\text{diag}}I$, where $x^*_{\text{diag}} \in \mathbb{R}$ is the unique real root of the cubic polynomial*

$$H_n(x) = (n^3 + 18n^2 + 79n + 94)x^3 + (3n^2 + 27n + 42)x^2 + (3n + 9)x + 1.$$

*Proof.* When $k = 0$, the matrix $A$ is diagonal and the skew-symmetric component vanishes. In this case, $g_k = 0$ trivially, and the stationarity condition reduces to

$$g_x(x, 0) = 4n \cdot H_n(x),$$

where $H_n(x)$ is a cubic polynomial in $x$ with coefficients depending on $n \in \mathbb{N}$. Explicitly,

$$H_n(x) = (n^3 + 18n^2 + 79n + 94)x^3 + (3n^2 + 27n + 42)x^2 + (3n + 9)x + 1.$$

To determine the number of real roots of $H_n$, we compute its discriminant

$$\Delta(H_n) = -108(n + 2)(27n^2 + 106n + 123).$$

Since all terms in the product are strictly positive for $n \in \mathbb{N}$, it follows that $\Delta(H_n) < 0$ for all $n \in \mathbb{N}$. A negative discriminant implies that $H_n(x)$ has exactly one real root (along with a pair of non-real complex conjugate roots) [Wik25a]. While this real root can, in principle, be written explicitly via Cardano's formula (for a modern treatment on cubics, see [BLW23] ), the resulting expression is algebraically cumbersome and provides little conceptual insight. For our purposes, existence and uniqueness suffice. Therefore, for each $n \in \mathbb{N}$, there exists a unique real value $x^*_{\text{diag}} \in \mathbb{R}$ such that the diagonal matrix

$$A = \begin{bmatrix} x^*_{\text{diag}} & 0 \\ 0 & x^*_{\text{diag}} \end{bmatrix}$$

is a stationary point of $f$. By Lemma 2.5.3, it follows that $-nx^*_{\text{diag}}I$ is a stationary point of $f_I$. □

**Lemma 2.5.6** (Skew-symmetric stationary points of $f_I(A)$ for $d = L = 2$). *For all $n \geq 15$, the in-context loss $f_I$ admits two additional stationary points with nonzero skew-symmetric components. These occur at*

$$A = -n \begin{bmatrix} x^*_{\text{skew}} & k^* \\ -k^* & x^*_{\text{skew}} \end{bmatrix},$$

*where $x^*_{\text{skew}} \in \mathbb{R}$ is the second-largest real root of the cubic polynomial*

$$R_n(x) = a_3 x^3 + a_2 x^2 + a_1 x + a_0, \tag{2.10}$$

*with coefficients*

$$a_3 = 32n^3 + 256n^2 + 672n + 576,$$
$$a_2 = 12n^3 + 144n^2 + 492n + 504,$$
$$a_1 = 19n^2 + 118n + 183,$$
$$a_0 = 7n + 25,$$

*and the corresponding skew-symmetric component* $k^* \in \mathbb{R} \setminus \{0\}$ *satisfies*

$$(k^*)^2 = -\frac{(n+2)(n^2+8n+23)(x_{\mathrm{skew}}^*)^2 + 2(n+2)(n+7)x_{\mathrm{skew}}^* + (n+7)}{(n-1)(n+1)(n+2)}.$$

*Proof.* Recall that the gradient component with respect to $k$ is given by

$$g_k = 8kn \cdot Q_n(x, k),$$

and we seek solutions with $k \neq 0$. Thus, $Q_n(x, k) = 0$, where

$$Q_n(x, k) = (n+2)(n^2+8n+23)x^2 + 2(n+2)(n+7)x + (n-1)(n+1)(n+2)k^2 + (n+7).$$

Solving for $k^2$ yields

$$k^2(x) = -\frac{(n+2)(n^2+8n+23)x^2 + 2(n+2)(n+7)x + (n+7)}{(n-1)(n+1)(n+2)}.$$

For $n = 1$, the denominator vanishes and $Q_n$ becomes independent of $k$:

$$Q_1(x) = 96x^2 + 48x + 8,$$

with discriminant $\Delta = 48^2 - 4 \cdot 96 \cdot 8 = -768 < 0$, so $Q_1(x) > 0$ for all $x \in \mathbb{R}$ [Wik25b]. Hence, no stationary point with $k \neq 0$ exists when $n = 1$. Now consider $n > 1$. Define the numerator of $k^2(x)$ as

$$p_n(x) := (n+2)(n^2+8n+23)x^2 + 2(n+2)(n+7)x + (n+7),$$

so that $k^2(x) > 0 \iff p_n(x) < 0$. This polynomial is quadratic in $x$ with a positive leading coefficient and is negative only between its real roots. A discriminant calculation shows $p_n$ has two distinct real roots if and only if $n > 9$. Thus, for $n \leq 9$, there exists no $x \in \mathbb{R}$ with $k^2(x) > 0$, and hence no stationary point with nonzero skew-symmetric component. Assume now $n > 9$. Let $\alpha_n < \beta_n$ be the two real roots of $p_n$. Then $k^2(x) > 0$ for all $x \in (\alpha_n, \beta_n)$, so any such $x$ yields $k \neq 0$. We now analyse $g_x = \partial f / \partial x$. Substituting the expression for $k^2(x)$ into the symbolic form of $g_x$ yields

$$g_x(x) = -\frac{16n}{n^2 - 1} \cdot R_n(x),$$

where $R_n(x)$ is the cubic polynomial defined in (2.10). The prefactor is nonzero for all $n \geq 2$, so $g_x = 0 \iff R_n(x) = 0$. The discriminant analysis confirms that $R_n(x)$ has three real roots when $n \geq 9$. For $n \geq 15$, symbolic evaluation shows

$$R_n(\alpha_n) > 0, \quad R_n(\beta_n) < 0,$$

so by the Intermediate Value Theorem, there exists a root $x^* \in (\alpha_n, \beta_n)$ such that $R_n(x^*) = 0$. Since this root lies in the admissible interval, $k^2(x^*) > 0$, yielding a nonzero skew component $k^* \in \mathbb{R} \setminus \{0\}$. Thus, for $n \geq 15$, the matrix

$$A = \begin{bmatrix} x^* & k^* \\ -k^* & x^* \end{bmatrix} \quad \text{and its transpose} \quad A^\top = \begin{bmatrix} x^* & -k^* \\ k^* & x^* \end{bmatrix}$$

are distinct stationary points of $f$ with nonzero skew-symmetric component. For $10 \leq n \leq 14$, numerical analysis confirms that no real root of $R_n$ lies in the interval $(\alpha_n, \beta_n)$, and thus no admissible stationary point with $k \neq 0$ exists in this range. Lastly, all stationary points are given by orthogonal conjugation (Lemma 2.5.2). In the skew-symmetric case,

$$A = x_n^* I + k^* J, \quad \text{where } J := \begin{bmatrix} 0 & 1 \\ -1 & 0 \end{bmatrix},$$

and for any $Q \in \mathrm{O}(2)$,

$$QAQ^\top = x_n^* I + k^* \det(Q) J = \begin{cases} A & \text{if } \det(Q) = +1, \\ A^\top & \text{if } \det(Q) = -1. \end{cases}$$

Therefore, the complete set of skew-symmetric stationary points is $\{A, A^\top\}$, and no further distinct minimisers arise under orthogonal conjugation (Lemma 2.5.2). $\qquad\square$

**Global minimizers**

**Theorem 2.5.7** (Global minimizers of $f_I(A)$ for $d = L = 2$)**.** *Let $d = L = 2$. Then:*

- *If $n < 15$, the unique global minimizer of $f_I$ is the diagonal matrix $A = -n x_{\mathrm{diag}}^* I$, where $x_{\mathrm{diag}}^* \in \mathbb{R}$ is the unique real root of the polynomial $H_n(x)$ defined in Lemma 2.5.5.*

- *If $n \geq 15$, $f_I$ admits two global minimizers with nonzero skew-symmetric components:*

$$A = -n \begin{bmatrix} x_{\mathrm{skew}}^* & k^* \\ -k^* & x_{\mathrm{skew}}^* \end{bmatrix} \quad and \quad A^\top = -n \begin{bmatrix} x_{\mathrm{skew}}^* & -k^* \\ k^* & x_{\mathrm{skew}}^* \end{bmatrix},$$

*where $x_{\mathrm{skew}}^*, k^* \in \mathbb{R} \setminus \{0\}$ are as described in Lemma 2.5.6.*

*Proof.* We analyse the Hessian of the reparametrised functional $f(A)$ at the stationary points identified previously. It suffices to study the case $k = 0$, corresponding to the diagonal stationary point. For a twice-differentiable function, a stationary point is a strict local minimum if the Hessian is positive definite at that point [Wik25c]. In the reduced parametrisation $(x, k)$, the Hessian takes the form

$$H(x, k) := \begin{bmatrix} \partial_x^2 f & \partial_x \partial_k f \\ \partial_k \partial_x f & \partial_k^2 f \end{bmatrix} \in \mathbb{R}^{2 \times 2}.$$

At the diagonal point $A = xI$, the off-diagonal terms vanish, and we have:

$$H(x, 0) = \begin{bmatrix} \partial_x^2 f(x, 0) & 0 \\ 0 & \partial_k^2 f(x, 0) \end{bmatrix}.$$

Let $x_n^* \in \mathbb{R}$ denote the unique root of $H_n(x) = 0$. The second derivative $\partial_x^2 f(x, 0)$ is a quadratic polynomial in $x$ with a strictly positive leading coefficient and a negative discriminant:

$$\Delta = -2304 n^2 (n + 2)(3n^2 + 18n + 43) < 0 \quad \text{for all } n \in \mathbb{N},$$

so it is strictly positive for all $x$ [Wik25b], and in particular $\partial_x^2 f(x_n^*, 0) > 0$. We now examine $\partial_k^2 f(x_n^*, 0)$, denoted $g_n(x_n^*)$, where

$$g_n(x) = A_n x^2 + B_n x + C_n,$$

with

$$A_n = 8n^4 + 80n^3 + 312n^2 + 368n,$$
$$B_n = 16n^3 + 144n^2 + 224n,$$
$$C_n = 8n^2 + 56n.$$

This is a quadratic polynomial with $A_n > 0$. The discriminant is

$$\Delta_n = 256 n^2 (n - 9)(\cdots),$$

which is nonnegative if and only if $n \geq 9$. Thus:

- If $n \leq 8$, then $g_n(x) > 0$ for all $x$, and in particular $g_n(x_n^*) > 0$.

- If $n \geq 9$, then $g_n(x)$ has two real roots $\ell_n < u_n < 0$, and is negative only on the interval $(\ell_n, u_n)$.

To determine whether $x_n^* \in (\ell_n, u_n)$, remember that $H_n(x)$ is a degree three polynomial with unique root $x_n^*$. Since the leading coefficient of $H_n$ is positive,

$$x_n^* \in (\ell_n, u_n) \iff H_n(\ell_n) < 0 < H_n(u_n).$$

To determine if this is the case, express $H_n(x)$ in terms of $g_n(x)$ via symbolic division:

$$H_n(x) = (k_1(n)x + k_0(n))g_n(x) + r_1(n)x + r_0(n).$$

The remainder has positive coefficients:

$$r_1(n) = \frac{2(5n^4 + 130n^3 + 900n^2 + 2014n + 1047)}{n^4 + 16n^3 + 110n^2 + 368n + 529} > 0,$$

$$r_0(n) = \frac{2(5n^3 + 99n^2 + 555n + 877)}{n^4 + 16n^3 + 110n^2 + 368n + 529} > 0.$$

Hence, if $x \in \{\ell_n, u_n\}$ is a root of $g_n$, then

$$H_n(x) = r_1(n)x + r_0(n),$$

which is negative exactly when $x < t_n := -\frac{r_0(n)}{r_1(n)} < 0$. So:

$$x_n^* \in (\ell_n, u_n) \iff H_n(\ell_n) < 0 < H_n(u_n) \iff \ell_n < t_n < u_n \iff g_n(t_n) < 0.$$

We evaluate $g_n(t_n)$ to determine when this occurs. Symbolic computation yields

$$g_n(t_n) = -8n(n^2 + 8n + 23)^2 \cdot N(n)/D(n),$$

where $D(n) > 0$ and

$$N(n) = 25n^4 - 26n^3 - 3624n^2 - 20934n - 32785.$$

Symbolic calculation yields that

$$n < 15 \iff N(n) < 0 \iff g_n(t_n) > 0 \iff x_n^* \notin (\ell_n, u_n) \iff g_n(x_n^*) > 0.$$

In summary

$$g_n(x_n^*) > 0 \iff n < 15, \quad \text{and} \quad g_n(x_n^*) < 0 \iff n \geq 15.$$

For $n < 15$, the diagonal stationary point $A = x_n^* I$ is the only stationary point of $f_I$. Since the Hessian at this point is positive definite, it is a strict local minimum. Being the only stationary point, it must also be the global minimiser.

For $n \geq 15$, the Hessian at the diagonal point becomes indefinite, as $g_n(x_n^*) < 0$, so $A = x_n^* I$ is a saddle point [Wik25c]. In this regime, the only remaining stationary points are the skew-symmetric ones identified in Lemma 2.5.6. Since $f_I$ must attain a global minimum and no other stationary points exist, these skew-symmetric points must be global minimisers. In particular, they are local minima by necessity. $\quad\square$

### 2.5.2 Analytic case: $L = 1$

For $L = 1$, $f : \mathbb{R}^{d \times d} \to \mathbb{R}$ takes the form

$$f(A) = \mathbb{E}_W \left[ \|I + AW\|_F^2 \right],$$

where $W \sim \mathcal{W}_d(n, I_d)$ is a Wishart-distributed random matrix with $n$ degrees of freedom and identity scale matrix.

**Theorem 2.5.8** (Minimiser of $f$ for $L = 1$). *$f$ is uniquely minimised at*

$$A^* = -\frac{1}{n + d + 1} I_d.$$

*Proof.* We expand the squared Frobenius norm using $\|M\|_F^2 = \text{Tr}(M^\top M)$.

$$f(A) = \mathbb{E}_W \left[ \text{Tr} \left( (I + AW)^\top (I + AW) \right) \right] = \mathbb{E}_W \left[ \text{Tr}(I + (AW)^\top + AW + WA^\top AW) \right].$$

Taking expectations and using linearity and symmetric invariance of the trace:

$$f(A) = \text{Tr}(I) + 2\mathbb{E}_W[\text{Tr}(AW)] + \mathbb{E}_W[\text{Tr}(WA^\top AW)].$$

For $W \sim \mathcal{W}_d(n, I_d)$, it holds by [Lemma A.2.1]{.green} that $\mathbb{E}[W] = nI_d$ and $\mathbb{E}[W^2] = n(n + d + 1)I_d$. Therefore,

$$\mathbb{E}_W[\text{Tr}(AW)] = \text{Tr}(A \cdot \mathbb{E}[W]) = n \text{Tr}(A),$$

and

$$\mathbb{E}_W[\text{Tr}(WA^\top AW)] = \text{Tr}(A^\top A \cdot \mathbb{E}[W^2]) = n(n + d + 1) \text{Tr}(A^\top A) = n(n + d + 1)\|A\|_F^2.$$

Combining all terms, we have

$$f(A) = d + 2n \text{Tr}(A) + n(n + d + 1)\|A\|_F^2.$$

This is a convex quadratic function in $A$. Taking the gradient with respect to $A$, we get

$$\nabla_A f(A) = 2nI_d + 2n(n + d + 1)A.$$

Setting $\nabla_A f(A) = 0$, we solve

$$2nI_d + 2n(n + d + 1)A = 0 \quad \Rightarrow \quad A = -\frac{1}{n + d + 1} I_d.$$

This is the unique minimiser since $f$ is strictly convex. □

**Corollary 2.5.9.** *The unique minimizer of $f_l$ is $\frac{n}{n+d+1}I$.*

This recovers [Ahn+23], Theorem 1 in the case $\Sigma = I_d$.

### 2.5.3 Empirical analysis

We empirically study the structural properties of global minimisers of the in-context loss function $f_l(A)$ across various parameter regimes. Our investigation serves three primary purposes:

- To verify the theoretical predictions derived for the analytically tractable case $d = L = 2$,

- To explore how the structure of minimisers evolves as a function of the sample size $n$, number of layers $L$, and model dimension $d$,

- To identify qualitative trends and potential phase transitions in the emergence of skew-symmetric components in the minimiser.

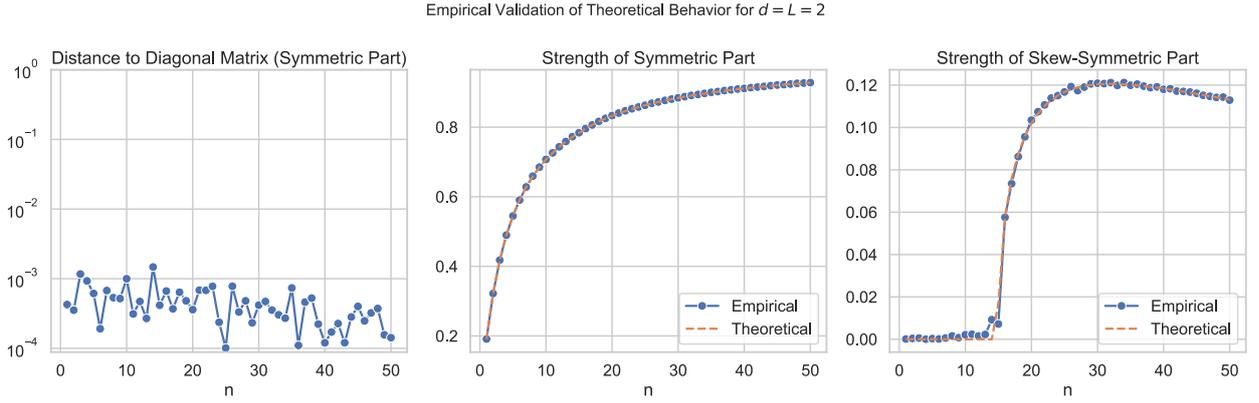

**Figure 2.1** Empirical validation for $d = L = 2$. Left: Distance to a diagonal symmetric matrix. Centre: Symmetric strength and theoretical prediction. Right: Skew-symmetric strength and theoretical curve.

**Baseline verification:** $d = L = 2$

We begin by validating our symbolic analysis in the special case $d = L = 2$. As shown in Fig. 2.1, the empirical behaviour aligns closely with the closed-form expressions derived in Theorem 2.5.7.

For $n < 15$, the minimiser is a scalar multiple of the identity, leading to a nearly diagonal symmetric part and negligible skew-symmetric strength. At $n = 15$, a sharp transition occurs: the skew-symmetric component becomes nontrivial, while the symmetric part remains close to diagonal and continues to match the theoretical strength curve. These results strongly support the symbolic expressions and offer a baseline for interpreting more complex settings.

*Remark* 2.5.10 (Quantifying spectral isotropy in the skew-symmetric part). To assess the isotropy of the skew-symmetric part $K = \frac{1}{2}(A - A^\top)$, we define the *isotropy distance* as the Frobenius norm distance to the set of skew-symmetric matrices with equal singular values:

$$\text{dist}_{\text{iso}}(K) := \sqrt{\sum_{i=1}^{d}(\sigma_i - \bar{\sigma})^2}, \quad \text{where} \quad \bar{\sigma} = \frac{1}{d}\sum_{i=1}^{d}\sigma_i.$$

This quantity vanishes if all singular values of $K$ are equal. We prove in Theorem B.4.1 that this distance corresponds to the minimal Frobenius distance to the isotropic class. For $d = 2$, this value is always zero and thus omitted.

**Varying sample size $n$**

Next, we fix $d = 5$, $L = 6$, and vary the sample size $n \in \{1, \ldots, 30\}$. Results are shown in Fig. 2.2.

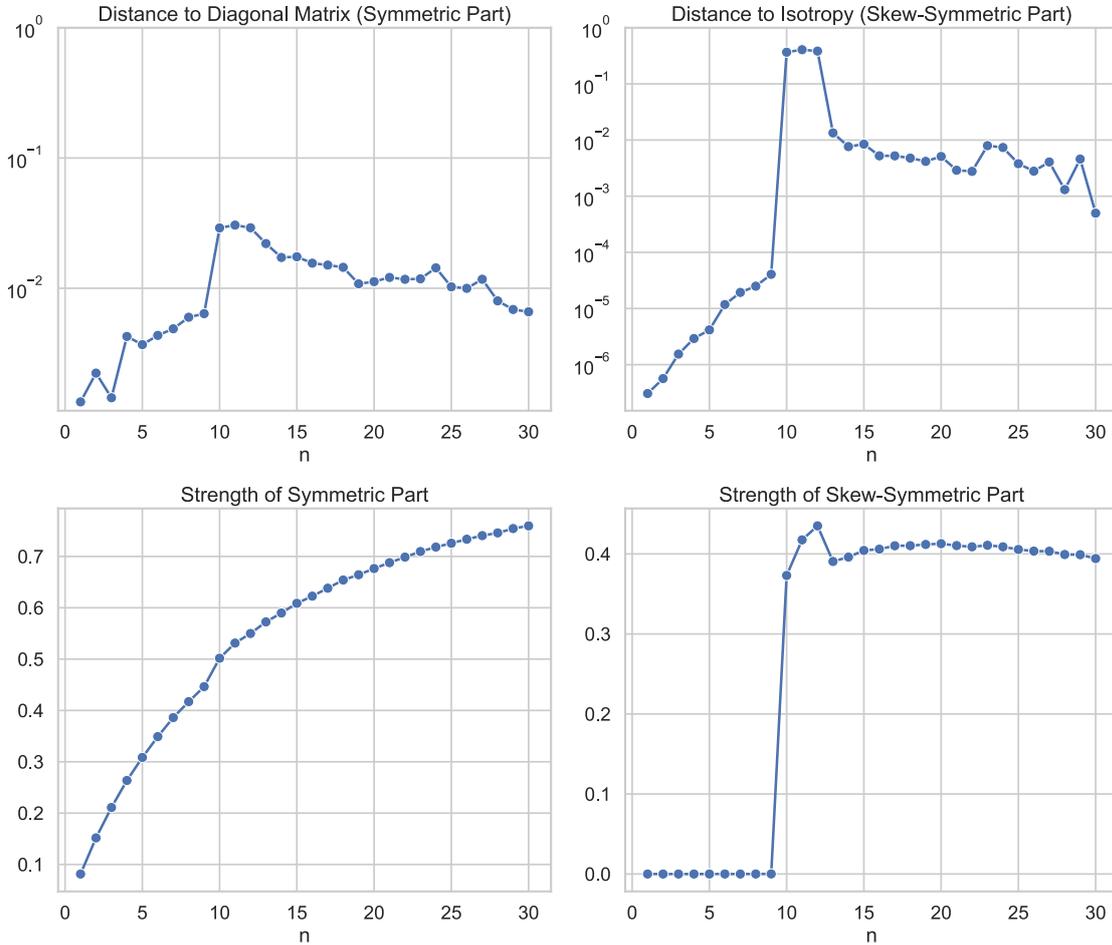

**Figure 2.2** Empirical behaviour for $d = 5$, $L = 6$, as a function of sample size $n$.

A sharp transition occurs at $n = 10$: the skew-symmetric strength $\|K\|$ jumps from nearly zero to approximately 0.4, while the isotropy distance peaks at similar values for $n = 10$–$12$. Afterwards, the isotropy distance rapidly decays to near-zero, indicating a transition to a spectrally isotropic skew-symmetric part. Meanwhile, the symmetric strength $\|S\|$ grows monotonically with $n$, suggesting increasing confidence with more data. Interestingly, the onset of skew symmetry coincides with a noticeable increase in symmetric strength, implying that the skew-symmetric component enables more aggressive optimisation steps.

**Varying number of layers $L$**

We now fix $d = 5$, $n = 10$, and vary the number of layers $L \in \{1, \ldots, 30\}$, interpreted as the number of in-context gradient steps. Results are presented in Fig. 2.3.

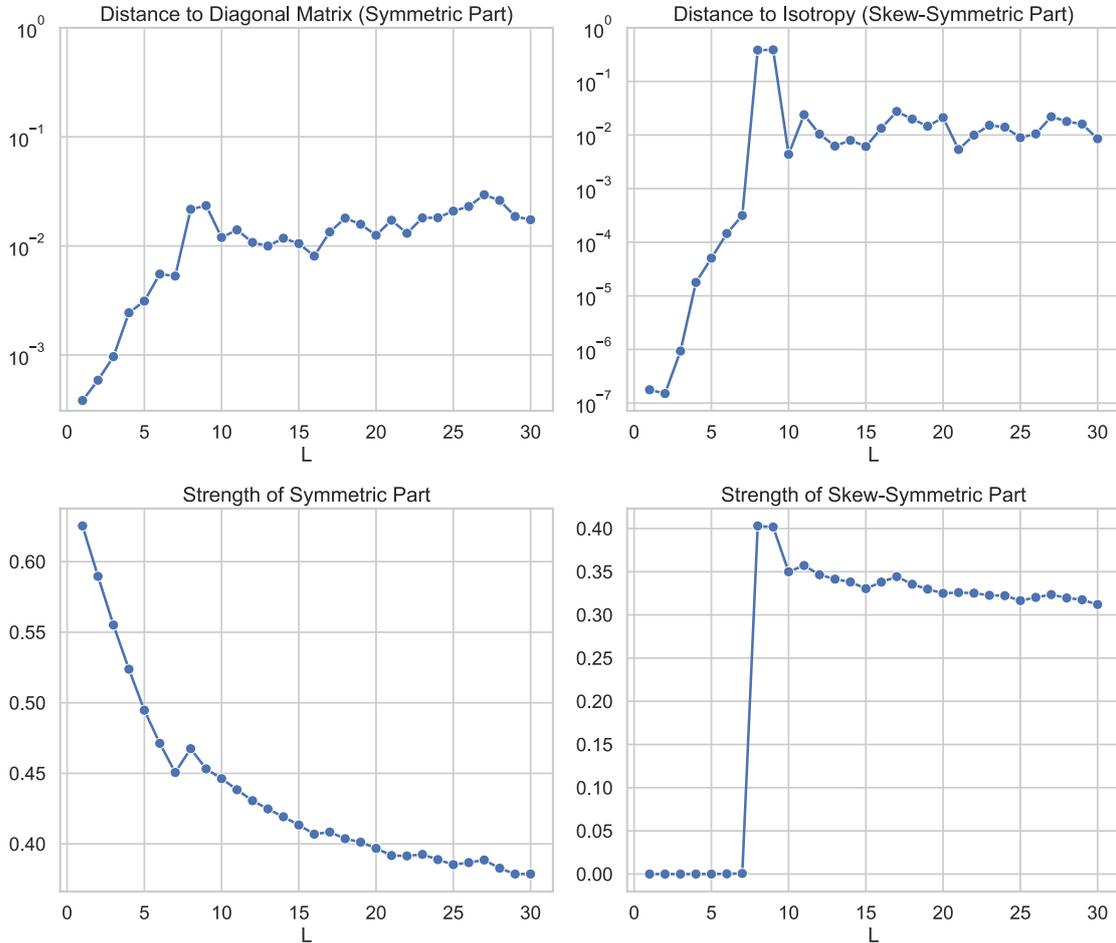

**Figure 2.3** Empirical results for $d = 5$, $n = 10$, and variable number of layers $L$.

A phase transition emerges at $L = 8$, where the skew-symmetric strength sharply rises to approximately 0.4. As in the $n$-sweep, this jump is accompanied by a peak in the isotropy distance, which later diminishes to near-zero. The symmetric part remains close to diagonal throughout, but its strength weakens with increasing $L$. Notably, the strength temporarily increases at the symmetry-breaking point, mirroring behaviour observed in the sample size sweep.

**Varying model dimension** $d$

Finally, we fix $n = 10$, $L = 6$, and vary the model dimension $d \in \{1, \ldots, 20\}$. The results are displayed in Fig. 2.4.

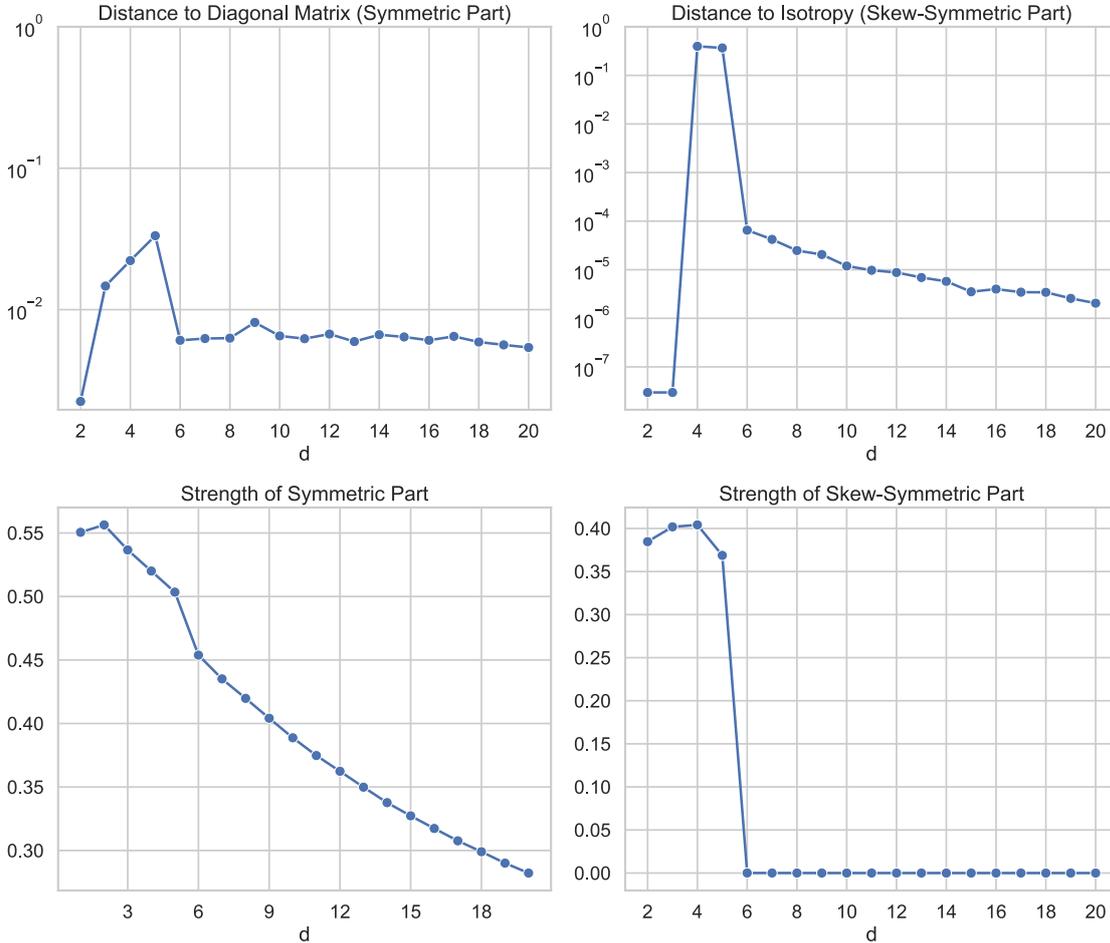

**Figure 2.4** Empirical results for $n = 10$, $L = 6$, as a function of model dimension $d$.

For $d = 1$, the model is scalar and has no skew-symmetric part. For $d = 2$–5, the skew-symmetric strength remains high ($\|K\| \approx 0.4$) and an isotropy plateau appears at $d = 4, 5$. At $d = 6$, the skew-symmetric part collapses to zero. A corresponding drop in symmetric strength is also observed at this transition point. Overall, symmetric strength decreases with $d$, while the symmetric part remains close to diagonal.

**Synthesis across experiments**

Across all settings, we observe consistent structural phenomena:

1. **Symmetry-breaking transition:** Each parameter sweep exhibits a sharp threshold where the skew-symmetric component becomes nonzero:

$$n_{\mathrm{crit}} = 10,$$
$$L_{\mathrm{crit}} = 8,$$
$$d_{\mathrm{crit}} = 5,$$
$$n_{\mathrm{crit}} = 15 \quad \text{for } d = L = 2.$$

2. **Anisotropy plateau:** Immediately following each transition, the skew-symmetric part is highly anisotropic:

$$\text{dist}_{\text{iso}} \approx 0.4 \quad \text{for} \quad \begin{cases} n = 10\text{--}12 \\ L = 8, 9 \\ d = 4, 5 \end{cases}$$

This plateau disappears as the spectrum equilibrates.

3. **Universal jump height:** The critical transition consistently yields

$$\|K\| \approx 0.4, \qquad \text{dist}_{\text{iso}} \approx 0.4,$$

suggesting a universal loss geometry that governs the emergence and shape of skew-symmetric structures.

**Post-transition dynamics.** The behaviour after the transition point differs across sweeps:

**$n$-sweep:** $\|K\|$ stabilizes near 0.4; $\|S\|$ continues to grow with $n$.

**$L$-sweep:** $\|K\|$ gradually declines; $\|S\|$ decreases with $L$.

**$d$-sweep:** $\|K\|$ vanishes for $d \geq 6$; $\|S\|$ decreases with $d$.

**Transition heatmap: when does skew symmetry emerge?**

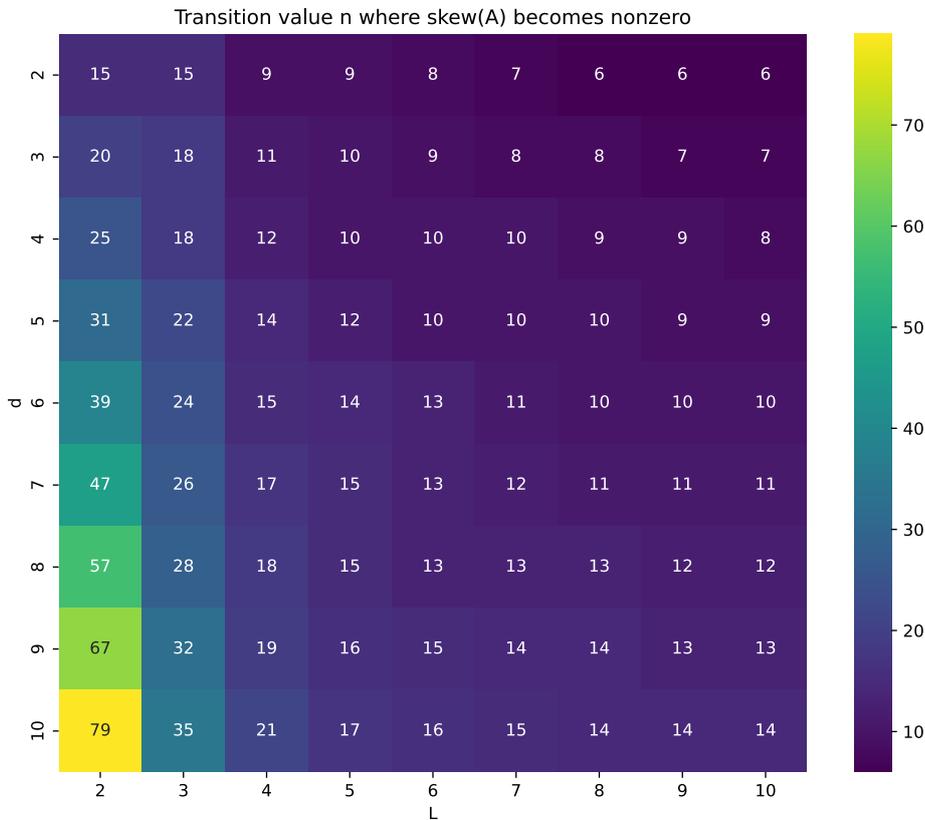

**Figure 2.5** Heatmap of the critical sample size $n$ at which the skew-symmetric part becomes nonzero, as a function of model dimension $d$ and number of layers $L$. Lower values indicate earlier emergence of skew symmetry.

To synthesise the threshold behaviour across a broader range of parameter settings, we construct a heatmap that records the critical value of $n$ at which the skew-symmetric component $K = \frac{1}{2}(A - A^\top)$ becomes nonzero, for various combinations of model dimension $d$ and number of layers $L$. The result is shown in Fig. 2.5. We define the transition point as the smallest sample size $n$ for which the skew-symmetric strength

$$\texttt{skew\_strength}(A) \coloneqq \left\| \tfrac{1}{2}(A - A^\top) \right\|_2$$

exceeds a threshold of $10^{-2}$. Due to numerical noise and optimisation imprecision, the empirical transition may slightly differ from the theoretical phase boundary. Nevertheless, the chosen criterion robustly captures the onset of symmetry breaking across settings. Several key patterns emerge:

- **Monotonic dependence on model dimension:** For fixed $L$, the transition value $n_{\text{crit}}$ increases roughly linearly with $d$, suggesting that larger models require more data before breaking symmetry and developing skew-symmetric structure.

- **Depth accelerates the transition, but with diminishing returns:** For fixed $d$, increasing the number of layers $L$ significantly reduces the required sample size for symmetry breaking, particularly in the low-to-moderate range of $L$. For instance, when $d = 6$, increasing $L$ from 2 to 4 lowers $n_{\text{crit}}$ from 39 to 15. However, the benefits of additional depth taper off beyond this point, with further increases in $L$ yielding progressively smaller reductions in $n_{\text{crit}}$.

- **Exact linear regime for large $L$:** Remarkably, for $L = 10$, the critical threshold follows an exact affine relation:

$$n_{\text{crit}}(d, L = 10) = d + 4,$$

suggesting that a predictable and straightforward scaling law governs the symmetry-breaking threshold beyond a certain depth.

- **Diagonal alignment with empirical observations:** The values along the slice $(d = 5, L = 6)$, $(d = 5, n = 10)$, and $(L = 6, n = 10)$ are consistent with the individual sweep experiments discussed previously, confirming that the heatmap captures the same underlying transition phenomenon.

This transition map offers a compact summary of how inductive bias, expressivity (through $d$), and optimisation capacity (through $L$) jointly govern the onset of skew-symmetric structure in the minimisers of the in-context loss. The pattern also supports the hypothesis of a *universal symmetry-breaking threshold surface* in the space of model and task parameters.

**Predictive model for symmetry-breaking threshold**

To capture the dependency of the symmetry-breaking threshold on model dimension $d$ and number of layers $L$, we fit a parametric model to the empirical transition map shown in Fig. 2.5. We chose the model to reflect both linear and inverse-depth scaling effects:

$$n_{\text{crit}}(d, L) = c_0 + c_1 d + c_2 \cdot \frac{d^p}{L^q},$$

where $c_0, c_1, c_2, p, q \in \mathbb{R}$ are learnable parameters. This structure captures both the baseline ($c_0 + c_1 d$) and the nonlinear influence of depth via a regularised power-law decay in $L$. Fitting this model to the empirical transition values yields the following coefficients:

$$c_0 = 6.95, \quad c_1 = 0.45, \quad c_2 = 12.67, \quad p = 1.44, \quad q = 2.43.$$

The model achieves strong predictive performance:

$$\text{RMSE} = 1.40, \qquad \text{MAE} = 0.95, \qquad \max |\text{error}| = 5.4.$$

The predicted transition surface closely approximates the empirical heatmap (see Fig. B.1), with most errors falling below 1.0. The most significant deviations occur at small $L$, where the data shows more variability and greater sensitivity to initialisation and optimisation dynamics.

Notably, the learned exponents $p \approx 1.44$ and $q \approx 2.43$ indicate that the effect of increasing depth decays super-quadratically, further supporting the earlier observation of diminishing returns in the depth dimension. While the model approaches an affine dependence on $d$ as $L$ increases, the learned slope and intercept do not match the empirical rule $n_{\text{crit}} = d+4$, suggesting a slight discrepancy in asymptotic scaling.

This model offers a compact and interpretable approximation of the transition behaviour across the parameter space. Full visualisations of the predicted surface and signed error map are provided in Figs. B.1 and B.2.

## 2.6 Classical analysis of the optimal preconditioner

Our earlier analysis showed that under weight tying and sparsity assumptions, minimising the in-context loss corresponds to minimising prediction loss under preconditioned gradient descent. In this section, we shift to a more classical setting, setting aside the interpretative lens of ICL to directly study the optimal preconditioner for gradient descent. We formulate the problem as follows.

**Problem 2.6.1** (Optimal Preconditioner for Gradient Descent in Linear Regression)**.** *Let* $x^{(i)}, x^{\text{test}} \overset{i.i.d.}{\sim} \mathcal{N}(0, \Sigma)$ *for* $i = 1, \ldots, n$, *and let* $w_* \sim \mathcal{N}(0, \Sigma^{-1})$. *Define the gradient descent iterates* $\{w_{\text{gd}}^{(\ell)}\}_{\ell=0}^{L}$ *by*

$$w_{\text{gd}}^{(0)} = 0,$$
$$w_{\text{gd}}^{(\ell+1)} = w_{\text{gd}}^{(\ell)} - A \nabla R_{w_*}\left(w_{\text{gd}}^{(\ell)}\right) \quad \text{for } \ell = 0, \ldots, L-1,$$

*where the loss with respect to the true weight vector* $w_*$ *is defined as*

$$R_{w_*}(w) \coloneqq \frac{1}{2n} \sum_{i=1}^{n} \left(w^{\top} x^{(i)} - w_*^{\top} x^{(i)}\right)^2.$$

*What choice of preconditioner* $A \in \mathbb{R}^{d \times d}$ *minimises the expected squared prediction error*

$$f_{\Sigma}(A) = \mathbb{E}\left[\left(\left\langle w_{\text{gd}}^{(L)}, x^{\text{test}}\right\rangle - \left\langle w_*, x^{\text{test}}\right\rangle\right)^2\right]?$$

The results established earlier remain applicable. In particular, Lemma 2.4.14 showed that the optimal preconditioner for general $\Sigma$ can be reduced to the case $\Sigma = I$, via the transformation $\Sigma = \mathbf{L}^{\top}\mathbf{L}$. Specifically,

$$\mathbf{L}^{\top} A \mathbf{L} \in \arg\min f_I \quad \Longleftrightarrow \quad A \in \arg\min f_{\Sigma}.$$

In Section 2.5, our analytic and empirical analysis demonstrated that for $\Sigma = I$, the minimizer takes the form $A^* = aI + W$, where $W$ is skew-symmetric.[3] While the skew component $W$ vanishes in some regimes, it appears across a wide range of model parameters. This corresponds to a gradient descent update with step size $a$, augmented by a rotational component induced by $W$. The update rule then takes the form

$$w_{\text{gd}}^{\ell+1} = w_{\text{gd}}^{\ell} - (aI + W) \nabla R_{w_*}(w_{\text{gd}}^{\ell})$$
$$= w_{\text{gd}}^{\ell} - a \nabla R_{w_*}(w_{\text{gd}}^{\ell}) - W \nabla R_{w_*}(w_{\text{gd}}^{\ell}),$$

where the second term induces a rotation in the update direction. Fig. 2.6 visualises the effect of the skew-symmetric component on the optimisation trajectory.

This discrete-time update corresponds to an explicit Euler discretisation of the continuous-time dynamics:

$$\frac{dw_{\text{gd}}(t)}{dt} = -a \nabla R(w_{\text{gd}}(t)) - W \nabla R(w_{\text{gd}}(t)),$$

where a skew-symmetric term perturbs standard gradient flow, we now relate this behaviour to recent theoretical results on skew-perturbed dynamics.

---

[3]For general $\Sigma$, the minimizer becomes $A^* = a\Sigma^{-1} + \mathbf{L}^{\top} W \mathbf{L}$, with $\mathbf{L}^{\top} W \mathbf{L}$ still skew-symmetric.

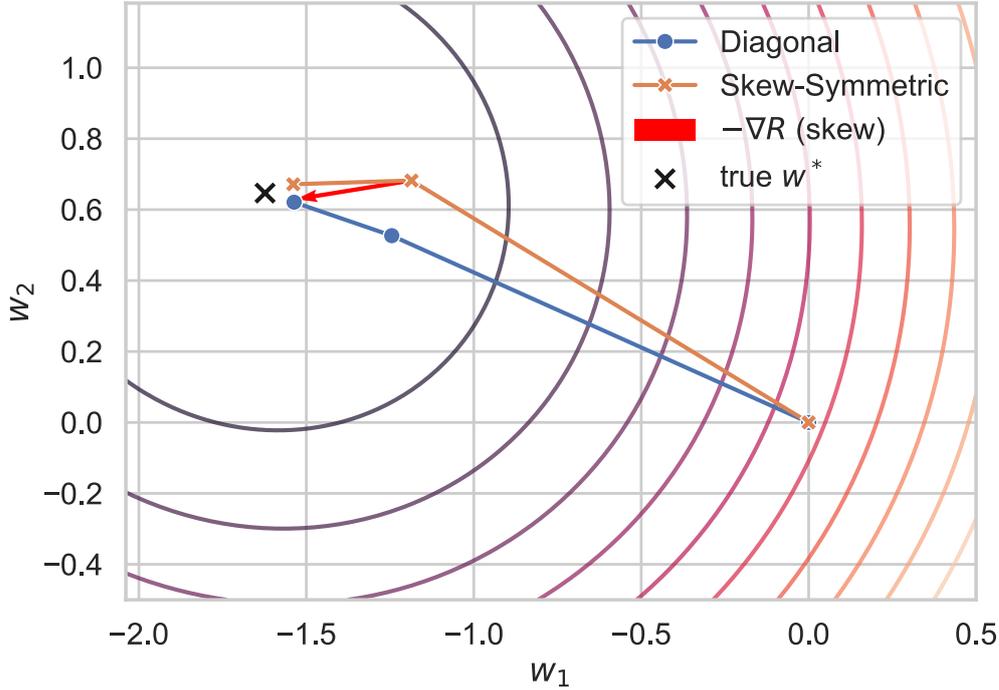

**Figure 2.6** Effect of skew-symmetric preconditioning on gradient descent trajectories for $n = 20$. Contour lines show the loss landscape in parameter space. The blue trajectory corresponds to a diagonal preconditioner, while the orange trajectory reflects a skew-augmented preconditioner. The red arrow indicates the direction of the skew-adjusted gradient at the final iterate. The true minimizer $w_*$ is marked with a black ×. The presence of a skew component induces a rotational dynamic.

**Theoretical motivation and acceleration via skew symmetry**  Futami et al. [Fut+21] considered dynamics of the form

$$\frac{dx(t)}{dt} = -(I + \alpha W)\nabla f(x(t)), \quad \text{with } W^\top = -W.$$

They showed that introducing a skew-symmetric matrix $J$ can accelerate convergence by improving the flow's spectral properties, specifically, by enhancing the effective condition number through rotational dynamics.

**Discretisation and stability considerations**  While the continuous-time skew-perturbed dynamics demonstrate accelerated convergence, Futami et al. [Fut+21] also noted that this benefit does not necessarily carry over to the naive Euler discretisation; in some cases, the discretised dynamics may even converge more slowly than standard gradient descent. To address this, they proposed a hybrid Euler–Leapfrog discretisation scheme that preserves the convergence advantages of the continuous flow.

**Our contribution**  We provide analytic and empirical evidence that skew-symmetric perturbations can enhance convergence, even without applying advanced discretisation schemes. In particular, we identify cases where naive Euler discretisation —despite its limitations—retains the beneficial effects of the skew-symmetric term.

## 2.7  Looped transformers for in-context learning

Recent research has introduced looped transformers—architectures with a single transformer block applied iteratively across multiple internal steps—as a compelling alternative to standard deep transformers for algorithmic and ICL tasks. This section reviews the key theoretical and empirical contributions that have positioned looped transformers as an efficient, interpretable, and provably capable class of models.

**Empirical evidence for algorithmic induction**  The foundational work by Yang et al. [Yan+24] demonstrates that looped transformers can effectively learn iterative algorithms across various in-context tasks. Their architecture consists of a single transformer block applied repeatedly to the hidden state, with shared parameters across all iterations. Through tasks such as sparse linear regression, decision tree fitting, and shallow neural network training, they show that looped models match the performance of much deeper unrolled transformers and often do so with significantly fewer parameters.

**Theoretical characterisation of the optimal looped learner**  Gatmiry et al. [Gat+24] give the first complete description of how a *trained* looped transformer behaves in in-context linear regression. Their Theorem 3.2 shows that the global minimiser $A_{\text{opt}}$ of the population loss has a very specific structure:

$$(1-c)\Sigma^{-1} \preceq A_{\text{opt}} \preceq (1+c)\Sigma^{-1}, \qquad c = 8\delta \, d^{1/(2L)}, \quad \delta = \left(\frac{8Ld}{\sqrt{n}}\right)^{1/(2L)},$$

provided that $\frac{8Ld^2}{\sqrt{n}} \leq \frac{1}{2^{2L}}$. Hence, the learned looped block *necessarily* realises exactly $L$ iterations of gradient descent with a preconditioner spectrally close to the ideal $\Sigma^{-1}$. As $n \to \infty$, the gap $c$ shrinks to zero, so the model converges to the statistically optimal linear estimator. Our results complement this finding. In the analytically tractable case $L = d = 2$, we also observe that $A_{\text{opt}} \to \Sigma^{-1}$ as $n \to \infty$, in agreement with the above convergence.

## 2.8 Limitations and outlook

The results in this chapter rely on a specific structural assumption: the sparsity of the linear transformer weight matrices. As such, it remains an open question whether similar phenomena arise in the non-sparse case. Preliminary checks indicate that skew symmetry does not persist in linear transformers after removing sparsity, suggesting that this structure may be specific to the sparse regime.

We also examined whether skew-symmetric components appear in real transformer architectures. In particular, we compared the self-attention weight matrices in BERT and ALBERT to investigate whether weight tying across layers, as implemented in ALBERT, leads to a larger skew-symmetric component in $Q^\top K$. However, we observed no significant difference in skew-symmetric strength between the two models. This result suggests that the emergence of skew-symmetric components may be specific to the theoretical setup studied here and might not generalise directly to practical transformer training or ICL.

Nevertheless, the appearance of skew symmetry remains interesting from a theoretical standpoint, particularly in connection with preconditioned gradient descent. While our analysis focused on linear regression with clean data, future work could explore whether similar dynamics emerge in the presence of noise or more complex tasks. Additionally, it may be instructive to investigate architectures with different activation functions, such as those studied in the context of functional gradient descent by Cheng, Chen, and Sra [CCS24]. Such work could help clarify whether skew symmetry is a byproduct of specific simplifying assumptions or indicative of a deeper optimisation principle.

# 3 Model collapse

## 3.1 Literature survey

Model collapse refers to the degenerative phenomenon where generative models trained on their own synthetic outputs begin to "forget" the true underlying data distribution. This issue was highlighted by Shumailov et al. [Shu+24a], who showed that indiscriminate use of model-generated content in training causes "irreversible defects"—rare events in the original data vanish and model behaviour narrows significantly. They demonstrated this effect across various generative paradigms, including LLMs, variational autoencoders (VAEs), and Gaussian mixture models (GMMs). A model's performance progressively degrades over successive generations of training on polluted (synthetic) data until it becomes effectively useless [Ale+23]. This raises concerns that widespread reliance on generative models may pollute data ecosystems and undermine the benefits of scaling laws [Doh+24a].

Shumailov et al. [Shu+24a] distinguish between *early collapse*—loss of tail information—and *late collapse*, where the model eventually produces outputs that no longer reflect the diversity or structure of the original data, often collapsing into a narrow, low-variance distribution. The root causes include (1) *statistical approximation error*, which arises due to finite sample sizes and causes rare events to be lost during resampling; (2) *functional expressivity error*, which results from the limited representational capacity of the model—e.g., fitting a mixture of Gaussians with a single Gaussian—and leads to mismatches even with perfect data; and (3) *functional approximation error*, which reflects biases or limitations in the learning process, such as those introduced by the optimisation algorithm or objective function. These errors compound across generations, causing cascading degradation. Alemohammad et al. [Ale+23] draw an analogy to "Model Autophagy Disorder (MAD)," where models consume their own outputs and degenerate over time without fresh real data.

**Collapse as a shift in scaling laws**   Recent theoretical work has begun to clarify the mechanisms by which synthetic data undermines model generalisation. Dohmatob, Feng, and Kempe [DFK24] study high-dimensional ridge regression in a self-training loop, deriving exact bias–variance decompositions that include an additional "collapse" term. This term captures a departure from classical scaling laws, predicting that test performance will stagnate once a sufficient portion of the training data is synthetic. Dohmatob et al. [Doh+24a] extend this analysis by viewing collapse as a structural shift in neural scaling laws, particularly in heavy-tailed regimes. They identify three distinct effects: (i) a breakdown of scaling, (ii) a shift in scaling exponents, and (iii) the unlearning of previously acquired skills. Notably, their framework accounts for grokking-like behaviour, in which improvements are delayed until real data is reintroduced. Dohmatob et al. [Doh+24b] go further, proving a form of *strong model collapse*, whereby even small fractions of synthetic data (as little as 0.1%) can halt generalisation progress. Larger models tend to exacerbate this effect below the interpolation threshold, overfitting to imperceptible artefacts in synthetic data. Collectively, these results suggest that the long-term use of synthetic data – even in small amounts– can fundamentally limit the benefits of scale.

**Avoiding or mitigating collapse**   Various ideas to alleviate the danger represented by model collapse have been proposed.

- **Mix, don't replace.** Retaining the original human-authored snapshot and *appending*—rather than overwriting with—synthetic text provably bounds distributional drift and has been validated on transformers, VAEs and diffusion models [Ger+24].

- **Provenance filtering.** Collapse can be mitigated by identifying synthetic content. Techniques include embedding robust watermarks at generation time [Dat+24] and using detectors during continued pre-training to down-weight machine-generated tokens [DYL25].

- **Quality-gated synthesis.** Token-level editing that resamples only "easy" or "hard" positions [Zhu+25] and verifier-gated pipelines that accept or reject each generated chunk before it enters the corpus [Fen+24] maintain diversity while scaling cheap data generation.

**Ongoing debates and outlook**   Schaeffer et al. [Sch+25] argue that the term "model collapse" has been inconsistently used, encompassing several distinct failure modes. They evaluate literature assumptions and suggest that many catastrophic collapse scenarios rely on unrealistic training setups. They believe the field can better assess risk and mitigation strategies by standardising definitions and assumptions.

**Conclusion**   Model collapse is a real and theoretically grounded phenomenon, with significant implications for the scalability of generative models. Although catastrophic outcomes can be avoided with careful data accumulation, synthetic data continues to pose challenges for scaling and generalisation. Maintaining data fidelity will be key to sustainable progress as generative AI becomes more prevalent.

## 3.2  What is model collapse?

Model collapse refers to the degradation of a generative model's output quality when it is repeatedly trained on data generated by previous versions of itself, rather than on independently-sampled real data.

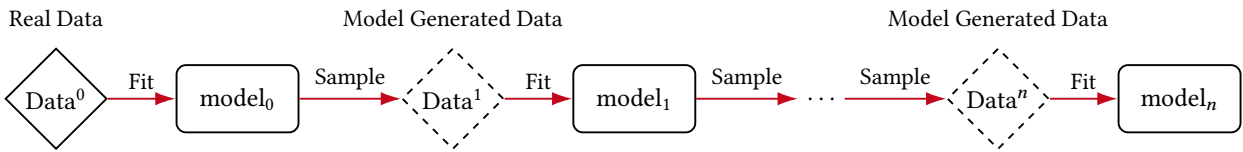

**Figure 3.1** Illustration of iterative training where each model is trained on data generated by its predecessor. Concept inspired by Figure 2 in [Shu+24b].

The process illustrated in Fig. 3.1 begins with real data, denoted $\text{Data}^0$, which is sampled from the true distribution $P_{\text{real}}$. A model $\text{model}_0$ is trained (*Fit*) on this data to approximate $P_{\text{real}}$. The trained model is then used to generate synthetic samples $\text{Data}^1$ (*Sample*), which serve as training data for the next model $\text{model}_1$. This cycle repeats: each model $\text{model}_k$ is trained on data generated by its predecessor $\text{model}_{k-1}$. Mathematically, this iterative process defines a sequence of training distributions:

$$P_0 := P_{\text{real}}, \quad P_k := P_{\text{model}_{k-1}} \quad \text{for } k \geq 1.$$

To analyse this process more precisely, we decompose it into two components: a *deterministic fitting step* and a *stochastic sampling step*. Specifically:

- Let $\mathcal{A}$ denote a deterministic learning algorithm that maps a dataset to a model. That is, for any dataset $\text{Data}^k$, the corresponding model is given by $\text{model}_k = \mathcal{A}(\text{Data}^k)$.

- The dataset $\text{Data}^{k+1}$ is randomly sampled from the distribution defined by $\text{model}_k$, i.e., $\text{Data}^{k+1} \sim P_{\text{model}_k}$.

This formalisation introduces stochasticity into the sequence of models and training distributions. For each $k \geq 1$, we obtain

$$\text{Data}^k \sim P_{\text{model}_{k-1}}, \quad \text{and} \quad \text{model}_k = \mathcal{A}(\text{Data}^k).$$

As a result, $\text{model}_k$ becomes a *random variable*, since it depends on the randomly sampled data $\text{Data}^k$. Consequently, the distribution $P_k := P_{\text{model}_k}$ is also a random variable. This recursive process can be abstracted as a stochastic recurrence:

$$P_0 := P_{\text{real}}, \quad P_k := \Phi(P_{k-1}, \xi_k), \quad k \geq 1,$$

where $\xi_k$ encapsulates the randomness in the sampling step, and $\Phi$ represents the composition of the sampling and fitting operations. Understanding the behaviour of this stochastic process—its convergence, divergence, or long-term stability—is essential for characterising the onset and progression of model collapse.

There are several ways to evaluate the quality of a model $\text{model}_k$, depending on the nature of the task. If the model defines a predictive function $f_k$, a common choice is the *population risk*, defined as the expected loss over the true data distribution:

$$\mathbb{E}_{x \sim P_{\text{real}}}[\ell(f_k(x))],$$

where $\ell$ is a task-specific loss function that quantifies the discrepancy between the model's output and the target. This quantity reflects how well the model generalises to real, unseen data drawn from $P_{\text{real}}$.

Alternatively, if the model defines a probability distribution over the data—such as in generative modeling or density estimation tasks—then quality can be assessed by comparing the learned distribution $P_k := P_{\text{model}_k}$ to the true data distribution $P_{\text{real}}$. A natural metric in this setting is the *Wasserstein distance*:

$$W(P_k, P_{\text{real}}),$$

which measures how much "mass" must be transported to transform one distribution into the other.

In the context of repeated model fitting, what exactly constitutes model collapse? The answer varies across the literature. As noted by Schaeffer et al. [Sch+25], at least eight definitions are in use—see Appendix C.3 for a survey.

In this work, we focus on two mathematically precise and analytically central definitions: *asymptotically diverging population risk* (see Item 3) and *collapsing variance* of the learned distribution (see Item 4). While much of the existing literature on model collapse emphasises convergence or divergence in expectation, we argue that almost sure convergence provides a more informative perspective. Notably, we observe cases where the population risk diverges in expectation, even though the model parameters converge almost surely. This highlights a subtle but important phenomenon: with probability one, the population risk may remain bounded, despite its expectation diverging.

We summarise our findings in Table 3.1.

## 3.3 Linear regression

This section investigates model collapse in linear regression under recursive self-training. The key idea is to model a stylised interaction loop between a model and its environment. At each iteration $i$, a fixed population of users queries the model $\hat{w}_i$ by submitting questions or prompts $X_i$. The model responds with answers $y_i$, which are used to retrain it. The process then repeats.

Crucially, we assume that the distribution of user prompts does not change over time. The rows of $X_i$ are sampled i.i.d. from a fixed distribution, modelling a stable environment or user base. The outputs $y_i$ are generated by the current model $\hat{w}_{i-1}$ plus noise, and the next model $\hat{w}_i$ is obtained by ordinary least squares (OLS) regression on the synthetic data $(X_i, y_i)$.

Our goal is to understand how the sequence of models $\{\hat{w}_i\}$ evolves over time, particularly how the prediction error behaves as the model recursively trains on its own outputs. We begin by analysing the case where a fresh dataset is sampled at every step ("resampling"). Later, we will study a variant where all data is accumulated over time ("cumulative data").

**Table 3.1** Summary of model collapse under different settings and data regimes.

| Setting | Replacing data | Cumulative data |
|---|---|---|
| **Gaussian fitting** | • **Wasserstein distance:** Converges almost surely $\Rightarrow$ *No collapse* in the sense of diverging population risk.<br><br>• **Variance** $\Sigma_i$**:** Collapses to 0 if and only if $M_i$ grows linearly or sublinearly $\Rightarrow$ *Collapse* in the sense of collapsing variance (unless $M_i$ grows superlinearly). | • **Wasserstein distance:** Converges almost surely $\Rightarrow$ *No collapse.*<br><br>• **Variance** $\Sigma_i$**:** Remains bounded away from 0; converges to non-degenerate limit $\Rightarrow$ *No collapse.* |
| **Linear regression** | • **Test error:** Diverges almost surely and in expectation $\Rightarrow$ *Collapse.*<br><br>• **Model parameters:** Diverge almost surely as a random walk $\Rightarrow$ *Collapse.* | • **Test error:** Converges almost surely; bounded in expectation $\Rightarrow$ *No collapse.*<br><br>• **Model parameters:** Converge almost surely $\Rightarrow$ *No collapse.* |

### 3.3.1 Relation to existing work

Our analysis builds upon and refines recent theoretical studies of model collapse in self-training loops. In the resampling regime, where each model iteration uses only freshly generated synthetic data, we recover the linear growth of test error first observed by Dohmatob, Feng, and Kempe [DFK24], who study kernel and ridge regression. We additionally show that the model parameters diverge almost surely, strengthening their expectation-based result. In contrast, in the cumulative regime—where all synthetic data is retained—we prove that the model sequence remains uniformly bounded in test error and converges both in $L^2$ and almost surely. This complements findings by Gerstgrasser et al. [Ger+24], demonstrating bounded risk under a fixed design matrix reused at each step. Our analysis generalises this by allowing the design matrices $X_i$ to be drawn independently from the same distribution, thus capturing a more realistic user-query setting.

### 3.3.2 Recursive self-training with resampled data

We now describe the formal setup for the resampling regime. Fix a true underlying model parameter $w^* \in \mathbb{R}^d$. We define an initial model $\hat{w}_0 = w^*$, and then recursively generate a sequence of models $(\hat{w}_i)_{i \geq 0}$ as follows:

- At step $i \geq 1$, draw a fresh matrix of user queries $X_i \in \mathbb{R}^{T \times d}$ with i.i.d. rows:

$$X_i^{(t)} \sim \mathcal{N}(0, \Sigma), \quad t = 1, \dots, T.$$

- Generate synthetic responses from the current model:

$$y_i = X_i \hat{w}_{i-1} + E_i, \quad E_i \sim \mathcal{N}(0, \sigma^2 I_T).$$

- Fit a new model by OLS:

$$\hat{w}_i = (X_i^\top X_i)^{-1} X_i^\top y_i.$$

We define the test error as the expected squared prediction loss on fresh data drawn from the same generative model used for training. Specifically, test inputs are sampled as $x \sim \mathcal{N}(0, \Sigma)$, and corresponding targets as $y = x^\top w^* + \varepsilon$, with $\varepsilon \sim \mathcal{N}(0, \sigma^2)$. The test error of $\hat{w}_n$ is given by

$$E_{\text{test}}(\hat{w}_n) := \mathbb{E}_{x,\varepsilon}\left[(x^\top \hat{w}_n - y)^2\right] - \sigma^2.$$

Writing $y = x^\top w^* + \varepsilon$ and expanding gives

$$\mathbb{E}_{x,\varepsilon}\left[(x^\top \hat{w}_n - y)^2\right] = \mathbb{E}_x\left[(x^\top (\hat{w}_n - w^*))^2\right] + \mathbb{E}[\varepsilon^2] = (\hat{w}_n - w^*)^\top \Sigma (\hat{w}_n - w^*) + \sigma^2.$$

Subtracting $\sigma^2$ yields the simplified form

$$E_{\text{test}}(\hat{w}_n) = (\hat{w}_n - w^*)^\top \Sigma (\hat{w}_n - w^*) = \|\hat{w}_n - w^*\|_\Sigma^2.$$

In contrast to Dohmatob, Feng, and Kempe [DFK24], we define the test error solely as an expectation over the test data $x$ and noise $\varepsilon$, and not over the model parameter $\hat{w}_n$. Their definition, $E_{\text{test}}(\hat{w}_n)$, corresponds to the expectation of our test error, i.e., $\mathbb{E}[E_{\text{test}}(\hat{w}_n)]$. We adopt this formulation to analyse almost sure convergence and divergence of the test error, which provides a more precise characterisation of model collapse than convergence in expectation. The following proposition corresponds to [DFK24, Theorem 4.1].

**Proposition 3.3.1** (Test error divergence). *Let $\hat{w}_n$ be defined by the recursion above. Then, under the assumptions that $T > d + 1$, the expected test error grows linearly with $n$*

$$\mathbb{E}[E_{\text{test}}(\hat{w}_n)] = \frac{n\sigma^2 d}{T - d - 1} \longrightarrow \infty \quad \text{as } n \to \infty.$$

*Proof.* We begin by rewriting the update rule as

$$\hat{w}_i = \hat{w}_{i-1} + (X_i^\top X_i)^{-1} X_i^\top E_i.$$

Unrolling the recursion and using the initial condition $\hat{w}_0 = w^*$, we obtain

$$\hat{w}_n = w^* + \sum_{i=1}^n (X_i^\top X_i)^{-1} X_i^\top E_i.$$

It follows that

$$\mathbb{E}[\|\hat{w}_n - w^*\|_\Sigma^2] = \sum_{i=1}^n \mathbb{E}\left[E_i^\top X_i (X_i^\top X_i)^{-1} \Sigma (X_i^\top X_i)^{-1} X_i^\top E_i\right],$$

where the cross terms vanish due to the independence of the $E_i$. Since $E_i$ is independent of $X_i$, we can take the expectation over $E_i$ to obtain

$$\mathbb{E}\left[E_i^\top X_i (X_i^\top X_i)^{-1} \Sigma (X_i^\top X_i)^{-1} X_i^\top E_i\right] = \sigma^2 \mathbb{E}\left[\text{tr}\left(X_i (X_i^\top X_i)^{-1} \Sigma (X_i^\top X_i)^{-1} X_i^\top\right)\right] = \sigma^2 \mathbb{E}\left[\text{tr}\left(\Sigma (X_i^\top X_i)^{-1}\right)\right].$$

Since $X_i^\top X_i \sim \mathcal{W}_d(\Sigma, T)$ and $T > d + 1$, by Proposition A.2.4 we have

$$\mathbb{E}[(X_i^\top X_i)^{-1}] = \frac{\Sigma^{-1}}{T - d - 1},$$

so

$$\mathbb{E}[\text{tr}(\Sigma (X_i^\top X_i)^{-1})] = \frac{\text{tr}(\Sigma \Sigma^{-1})}{T - d - 1} = \frac{d}{T - d - 1}.$$

Putting it all together:

$$\mathbb{E}[E_{\text{test}}(\hat{w}_n)] = n\sigma^2 \frac{d}{T - d - 1} \longrightarrow \infty \quad \text{as } n \to \infty. \qquad \square$$

The following theorem and corollary present original contributions. They establish the almost sure divergence of both the estimated weights $\hat{w}_n$ and the test error $E_{\text{test}}(\hat{w}_n)$, contrasting the cumulative case discussed in Section 3.3.3, where the weights and test error converge almost surely.

**Theorem 3.3.2** (Almost-sure divergence). *Assume $d \geq 3$, $T > d + 1$, and that the design matrices $X_i$ have independent rows $X_i^{(t)} \sim \mathcal{N}(0, \Sigma)$ with $\Sigma$ positive definite. Then*

$$\|\hat{w}_n - w^*\| \xrightarrow{a.s.} \infty.$$

*Proof.* Let $S_n := \hat{w}_n - w^* = \sum_{i=1}^n \Delta_i$, where $\Delta_i := (X_i^\top X_i)^{-1} X_i^\top E_i$. The increments $\Delta_i$ for $i \geq 1$ are i.i.d., centred, and have positive-definite covariance

$$\text{Cov}(\Delta_i) = \sigma^2 \mathbb{E}[(X_i^\top X_i)^{-1}] = \frac{\sigma^2}{T - d - 1} \Sigma^{-1},$$

where we calculate the expectation of the inverse Wishart distribution by Proposition A.2.4. Hence $S_n$ forms a *truly $d$-dimensional* random walk in the sense of Durrett [Dur19, Chapter 5, §5.4], because its increment distribution is not supported on any proper linear subspace. Any truly three-dimensional random walk is *transient*; that is, it does not return to any bounded set infinitely often [Dur19, Theorem 5.4.14]. Thus, for $d \geq 3$, any three-dimensional sub-random walk is transient, implying that the entire $d$-dimensional walk is also transient. This means the walk visits every bounded set only finitely many times; equivalently,

$$\lim_{n \to \infty} \|S_n\| = \infty \quad \text{almost surely.} \qquad \square$$

*Remark* 3.3.3. For $d = 1, 2$ we can check recurrence with [Dur19, Theorem 5.4.8/9].

**Corollary 3.3.4** (Almost sure divergence of test error). *Under the assumptions of Theorem 3.3.2, the test error diverges almost surely:*

$$E_{\text{test}}(\hat{w}_n) = \|\hat{w}_n - w^*\|_\Sigma^2 \xrightarrow{a.s.} \infty.$$

*Proof.* By positive-definiteness of $\Sigma$, there exists $c > 0$ such that

$$\|v\|_\Sigma^2 = v^\top \Sigma v \geq c \|v\|^2 \quad \text{for all } v \in \mathbb{R}^d.$$

Applying this to $v = \hat{w}_n - w^*$, we get

$$E_{\text{test}}(\hat{w}_n) = \|\hat{w}_n - w^*\|_\Sigma^2 \geq c \|\hat{w}_n - w^*\|^2.$$

Since $\|\hat{w}_n - w^*\| \to \infty$ a.s. by Theorem 3.3.2, the test error diverges almost surely as well. $\qquad \square$

We have shown that in the case of linear regression with replacing data, the test error diverges both almost surely and in expectation. Thus, model collapse occurs with probability one. We now contrast this with linear regression using cumulative data, where we will see that model collapse is almost surely averted.

### 3.3.3 Recursive self-training with cumulative data

We now consider a modified setting in which all previously generated data is retained. Instead of retraining the model using only the most recent synthetic dataset at each step, we use the entire history of all data collected so far. As before, we initialise $\hat{w}_0 = w^*$. At each iteration $i \geq 1$:

- A new batch of inputs $X_i \in \mathbb{R}^{T \times d}$ is sampled, with i.i.d. rows $X_i^{(t)} \sim \mathcal{N}(0, \Sigma)$.

- Responses are generated as $y_i = X_i \hat{w}_{i-1} + E_i$, with $E_i \sim \mathcal{N}(0, \sigma^2 I_T)$.

- Define the full accumulated dataset:

$$X_{1:i} = \begin{bmatrix} X_1 \\ \vdots \\ X_i \end{bmatrix}, \quad y_{1:i} = \begin{bmatrix} y_1 \\ \vdots \\ y_i \end{bmatrix}.$$

- Ordinary least squares (OLS) fits the updated model on the entire accumulated dataset:

$$\hat{w}_i = (X_{1:i}^\top X_{1:i})^{-1} X_{1:i}^\top y_{1:i}.$$

This cumulative scheme reflects a scenario in which synthetic data is never discarded, and each new model sees a growing set of training examples over time.

As previously mentioned, we extend the analysis of Gerstgrasser et al. [Ger+24] in two key ways: first, by considering the more general setting where the input matrices $X_i$ may vary across iterations; and second, by strengthening the analysis from convergence in expectation to almost sure convergence, which provides a sharper and more informative characterisation of model collapse.

**Theorem 3.3.5** (Martingale property of $(\hat{w}_i)$). *In the cumulative data regime described above, the sequence $(\hat{w}_i)$ is a martingale with respect to the filtration $\mathcal{F}_i := \sigma(X_1, E_1, \ldots, X_i, E_i)$.*

*Proof.* Define the accumulated Gram matrices $S_i := X_{1:i}^\top X_{1:i}$ and note that

$$S_i = S_{i-1} + X_i^\top X_i.$$

From the definition of $\hat{w}_i$, we can write the update as

$$\hat{w}_i = S_i^{-1} \left( S_{i-1} \hat{w}_{i-1} + X_i^\top y_i \right).$$

This identity holds because $S_{i-1} \hat{w}_{i-1} = X_{1:i-1}^\top y_{1:i-1}$ by definition of $\hat{w}_{i-1}$. Expanding this, we obtain

$$\hat{w}_i = \hat{w}_{i-1} + S_i^{-1} X_i^\top \left( y_i - X_i \hat{w}_{i-1} \right).$$

By the data generation process, we have $y_i - X_i \hat{w}_{i-1} = E_i$, so

$$\hat{w}_i = \hat{w}_{i-1} + S_i^{-1} X_i^\top E_i \tag{3.1}$$

Now take the conditional expectation of $\hat{w}_i$ given $\mathcal{F}_{i-1}$:

$$\mathbb{E}[\hat{w}_i \mid \mathcal{F}_{i-1}] = \mathbb{E}\left[\hat{w}_{i-1} + S_i^{-1} X_i^\top E_i \big| \mathcal{F}_{i-1}\right] = \hat{w}_{i-1} + \mathbb{E}\left[S_i^{-1} X_i^\top E_i \big| \mathcal{F}_{i-1}\right]$$
$$= \hat{w}_{i-1} + \mathbb{E}_{X_i}\left[S_i^{-1} X_i^\top \mathbb{E}[E_i \mid X_i, \mathcal{F}_{i-1}]\right] = \hat{w}_{i-1},$$

where the final step uses $\mathbb{E}[E_i \mid X_i, \mathcal{F}_{i-1}] = 0$ by independence. Hence, $\mathbb{E}[\hat{w}_i \mid \mathcal{F}_{i-1}] = \hat{w}_{i-1}$, so $(\hat{w}_i)_{i \geq 0}$ is a martingale with respect to the filtration $(\mathcal{F}_i)_{i \geq 0}$. $\square$

**Theorem 3.3.6** (Uniform boundedness of the expected test error). *Assume $T > d + 1$. Then the test error*

$$E_{\text{test}}(\hat{w}_i) = (\hat{w}_i - w^*)^\top \Sigma (\hat{w}_i - w^*)$$

*is uniformly bounded in expectation over all iterations $i \geq 0$, that is,*

$$\sup_{i \geq 0} \mathbb{E}\left[E_{\text{test}}(\hat{w}_i)\right] < \infty.$$

*Proof.* We denote by $Z_i := \hat{w}_i - w^*$ the estimation error at iteration $i$, and define the martingale increments $\Delta_k := \hat{w}_k - \hat{w}_{k-1}$. As shown in (3.1), $\Delta_k$ satisfies

$$\Delta_k = S_k^{-1} X_k^\top E_k,$$

where $S_k := X_{1:k}^\top X_{1:k}$ and $E_k \sim \mathcal{N}(0, \sigma^2 I_T)$ is independent of all previous data and of $X_k$. Then we have $Z_i = \sum_{k=1}^i \Delta_k$, and hence

$$\mathbb{E}[E_{\text{test}}(\hat{w}_i)] = \mathbb{E}\left[Z_i^\top \Sigma Z_i\right] = \text{tr}\left(\Sigma \mathbb{E}[Z_i Z_i^\top]\right).$$

To evaluate the covariance $\mathbb{E}[Z_i Z_i^\top]$, we expand

$$Z_i Z_i^\top = \sum_{k=1}^{i} \sum_{\ell=1}^{i} \Delta_k \Delta_\ell^\top.$$

Now observe that for $k \neq \ell$, we have

$$\mathbb{E}[\Delta_k \Delta_\ell^\top] = \mathbb{E}\left[S_k^{-1} X_k^\top E_k E_\ell^\top X_\ell S_\ell^{-1}\right] = 0,$$

since $E_k$ and $E_\ell$ are independent and mean zero. It follows that the cross-covariance terms vanish, and we obtain

$$\mathbb{E}[Z_i Z_i^\top] = \sum_{k=1}^{i} \mathbb{E}[\Delta_k \Delta_k^\top].$$

We now compute the covariance of each $\Delta_k$. Conditioning on the design matrices,

$$\mathbb{E}[\Delta_k \Delta_k^\top \mid X_{1:k}] = \sigma^2 S_k^{-1} A_k S_k^{-1}, \qquad \text{where } A_k := X_k^\top X_k.$$

Taking expectations and using the exchangeability of the batches $A_1, \ldots, A_k$, we observe that all terms $S_k^{-1} A_j S_k^{-1}$ contribute equally to the sum $S_k^{-1} S_k S_k^{-1} = \sum_{j=1}^{k} S_k^{-1} A_j S_k^{-1}$. Therefore,

$$\mathbb{E}[S_k^{-1} A_k S_k^{-1}] = \frac{1}{k} \mathbb{E}[S_k^{-1}].$$

Since $S_k \sim \mathcal{W}_d(kT, \Sigma)$ is a Wishart-distributed matrix with $kT > d + 1$, we may use <span style="color:green">Proposition A.2.4</span> to obtain

$$\mathbb{E}[S_k^{-1}] = \frac{\Sigma^{-1}}{kT - d - 1},$$

which yields

$$\mathbb{E}[\Delta_k \Delta_k^\top] = \sigma^2 \frac{\Sigma^{-1}}{k(kT - d - 1)}.$$

Substituting into the expression for the expected test error, we obtain

$$\mathbb{E}[E_{\text{test}}(\hat{w}_i)] = \text{tr}\left(\Sigma \sum_{k=1}^{i} \mathbb{E}[\Delta_k \Delta_k^\top]\right) = \sigma^2 \sum_{k=1}^{i} \frac{\text{tr}(\Sigma \Sigma^{-1})}{k(kT - d - 1)} = \sigma^2 d \sum_{k=1}^{i} \frac{1}{k(kT - d - 1)}.$$

For $T \geq d + 2$, the denominator satisfies $kT - d - 1 \geq k$, so the series $\sum_{k=1}^{\infty} \frac{1}{k(kT-d-1)} \leq \sum_{k=1}^{\infty} \frac{1}{k^2}$ converges. It follows that

$$\sup_{i \geq 0} \mathbb{E}[E_{\text{test}}(\hat{w}_i)] = \sigma^2 d \sum_{k=1}^{\infty} \frac{1}{k(kT - d - 1)} < \infty. \qquad \square$$

**Corollary 3.3.7** (Uniform $L^2$-boundedness). *In the cumulative data regime, the sequence $(\hat{w}_i)$ is uniformly bounded in $L^2$, i.e.,*

$$\sup_{i \geq 0} \mathbb{E}[\|\hat{w}_i\|^2] < \infty.$$

*Proof.* By the boundedness of the test error <span style="color:green">Theorem 3.3.6</span>,

$$\mathbb{E}[\|\hat{w}_i - w^*\|_\Sigma^2] \leq C < \infty \quad \text{for all } i.$$

Let $\lambda_{\min} > 0$ be the smallest eigenvalue of $\Sigma$. Then

$$\mathbb{E}[\|\hat{w}_i - w^*\|^2] \leq \frac{1}{\lambda_{\min}} \mathbb{E}[\|\hat{w}_i - w^*\|_\Sigma^2] \leq \frac{C}{\lambda_{\min}}.$$

Using the triangle inequality

$$\mathbb{E}[\|\hat{w}_i\|^2] \leq 2\|w^*\|^2 + 2\mathbb{E}[\|\hat{w}_i - w^*\|^2] \leq 2\|w^*\|^2 + \frac{2C}{\lambda_{\min}}.$$

Thus, $\sup_i \mathbb{E}[\|\hat{w}_i\|^2] < \infty$. $\qquad \square$

**Corollary 3.3.8** (Almost sure and $L^2$ convergence). *In the cumulative data regime, the sequence $(\hat{w}_i)$ converges almost surely and in $L^2$ to a limiting random vector $\hat{w}_\infty$.*

*Proof.* From Theorem 3.3.5, $(\hat{w}_i)$ is a martingale with respect to the filtration $\mathcal{F}_i$ and is uniformly bounded in $L^2$ by Corollary 3.3.7. By the martingale convergence theorem Theorem C.2.4, it follows that $\hat{w}_i$ converges almost surely and in $L^2$ to a limit $\hat{w}_\infty$. $\qquad\square$

We have shown that in the cumulative case, the test error remains uniformly bounded in expectation, and moreover, that the model parameters $\hat{w}_i$ converge almost surely. As a result, the test error $E_{\text{test}}(\hat{w}_i)$ also converges almost surely. This implies that model collapse is almost surely averted—marking a clear contrast with the replacing data case.

## 3.4 Gaussian fitting

In this section, we study the iterative fitting of Gaussian models. We begin with an initial distribution

$$\mathcal{P}_0 = \mathcal{N}(\mu_0, \Sigma_0),$$

where $\mu_0 \in \mathbb{R}^d$, and $\Sigma_0 \in \mathbb{R}^{d \times d}$ is symmetric, positive semidefinite, and nonzero. Subsequent distributions

$$\mathcal{P}_i = \mathcal{N}(\mu_i, \Sigma_i)$$

are generated recursively as follows: given $\mathcal{P}_{i-1}$, we draw $M_i \geq 2$ independent samples $p_1^i, \ldots, p_{M_i}^i$ from $\mathcal{P}_{i-1}$. Based on these samples, we estimate the parameters for the next iteration. The new mean $\mu_i$ is set as the sample average

$$\mu_i = \frac{1}{M_i} \sum_{j=1}^{M_i} p_j^i.$$

The new covariance matrix $\Sigma_i$ is estimated using the unbiased sample covariance

$$\Sigma_i = \frac{1}{M_i - 1} \sum_{j=1}^{M_i} (p_j^i - \mu_i)(p_j^i - \mu_i)^\top.$$

Our goal is to understand the behaviour of the sequence of distributions $\{\mathcal{P}_i\}$ as $i \to \infty$.

The following theorem shows that the Wasserstein distance $W_2(\mathcal{P}_0, \mathcal{P}_i)$ converges almost surely, indicating that model collapse is averted in the sense of *Asymptotically Diverging Population Risk* (see Item 3), as measured by the Wasserstein distance. However, the second notion of model collapse considered in this chapter—*Collapsing Variance* (see Item 4)—requires separate treatment. We prove that the variance collapses to zero if and only if the growth of artificial data is linear or sublinear, characterised by the finiteness of the series $S_\infty = \sum_{i=1}^{\infty} \frac{1}{M_i}$. In other words, model collapse in the sense of collapsing variance is averted if and only if the dataset grows superlinearly. This stands in contrast to the cumulative Gaussian fitting setup in Section 3.5, where the steady addition of new data leads to linear growth, sufficient to prevent collapse.

The results in this section appear to be largely novel. While Alemohammad et al. [Ale+23] consider collapsing variance, they assert that $\Sigma_i \to 0$ almost surely, regardless of the growth rate of the sample size $M_i$. Our analysis reveals a more nuanced picture: variance collapse depends critically on the asymptotic behaviour of $M_i$, occurring only when growth is linear or sublinear. Furthermore, existing analyses such as [Ale+23; Shu+24a] focus on the divergence of the Wasserstein distance in expectation, taking this as evidence of model collapse. In contrast, our results demonstrate that the Wasserstein distance converges almost surely, implying that model collapse is almost surely averted under the assumptions studied here.

**Theorem 3.4.1** (Main theorem: Gaussian fitting). *Let $(\mathcal{P}_i)_{i \geq 0}$ be the sequence of Gaussian distributions generated by the iterative procedure described above, where $\mathcal{P}_i = \mathcal{N}(\mu_i, \Sigma_i)$. Then the sequences $(\mu_i)_{i \geq 0}$ and* $(\Sigma_i)_{i \geq 0}$ converge almost surely to finite random limits $\mu_\infty \in \mathbb{R}^d$ and $\Sigma_\infty \in S_d^+$. Consequently, the Wasserstein-2 distance satisfies

$$W_2(\mathcal{P}_0, \mathcal{P}_i) \xrightarrow[i \to \infty]{a.s.} W_2(\mathcal{P}_0, \mathcal{P}_\infty), \quad \text{where } \mathcal{P}_\infty = \mathcal{N}(\mu_\infty, \Sigma_\infty).$$

Define

$$S_\infty := \sum_{i=1}^{\infty} \frac{1}{M_i}.$$

Then the limiting behaviour of the sequence depends on the value of $S_\infty$:

(i) **If** $S_\infty < \infty$:

- The sequences $(\mu_i)$, $(\Sigma_i)$, and $(W_2(\mathcal{P}_0, \mathcal{P}_i))$ converge in $L^2$. In particular, $\mu_\infty \in L^2(\Omega; \mathbb{R}^d)$, $\Sigma_\infty \in L^2(\Omega; S_d^+)$, and $W_2(\mathcal{P}_0, \mathcal{P}_\infty) \in L^2(\Omega)$.

- The limiting covariance is not almost surely degenerate:

$$\mathbb{P}(\Sigma_\infty = 0) < 1.$$

(ii) **If** $S_\infty = \infty$:

- The sequences $(\mu_i)$ and $(W_2(\mathcal{P}_0, \mathcal{P}_i))$ do not converge in $L^2$.

- The limiting covariance is almost surely degenerate:

$$\mathbb{P}(\Sigma_\infty = 0) = 1.$$

*Proof.* We establish each claim in separate parts:

- Almost sure convergence of $\Sigma_i$ is shown in Proposition 3.4.6.

- Almost sure convergence of $\mu_i$ is addressed in
  - Theorem 3.4.15 for the case $S_\infty < \infty$,
  - Proposition 3.4.17 for the case $S_\infty = \infty$.

- Almost sure convergence of $W_2(\mathcal{P}_0, \mathcal{P}_i)$ follows from the continuity of the Wasserstein metric; see Fact 3.4.18.

- Assuming $S_\infty < \infty$, convergence in $L^2$ of
  - $\mu_i$ follows from Theorem 3.4.15,
  - $\text{tr}(\Sigma_i)$ follows from Proposition 3.4.8,
  - $W_2(\mathcal{P}_0, \mathcal{P}_i)$ follows from the $L^2$-convergence of its components and continuity; see Fact 3.4.18.

- $\mathbb{P}(\Sigma_\infty = 0) < 1$ under $S_\infty < \infty$ follows from Proposition 3.4.8 and $\mathbb{P}(\Sigma_\infty = 0) = 1$ under $S_\infty = \infty$ follows from Theorem 3.4.9.

- The lack of $L^2$ convergence of $(\mu_i)$ and $(W_2(\mathcal{P}_0, \mathcal{P}_i))$ in the case $S_\infty = \infty$ is established in Proposition 3.4.19. $\qquad\square$

We introduce an equivalent construction to eliminate the presence of random variables as parameters of distributions. Let $Z_j^i \sim \mathcal{N}(0, I)$ be standard normal random vectors, independent across $i$ and $j$. Define

$$p_j^i = \mu_{i-1} + \Sigma_{i-1}^{1/2} Z_j^i,$$

so that $p_j^i \sim \mathcal{N}(\mu_{i-1}, \Sigma_{i-1})$ by construction. The updated parameters are then given by

$$\mu_i = \mu_{i-1} + \Sigma_{i-1}^{1/2} \frac{1}{M_i} \sum_{j=1}^{M_i} Z_j^i,$$

$$\Sigma_i = \frac{1}{M_i - 1} \Sigma_{i-1}^{1/2} \left( \sum_{j=1}^{M_i} (Z_j^i - \bar{Z}^i)(Z_j^i - \bar{Z}^i)^\top \right) \Sigma_{i-1}^{1/2},$$

where $\Sigma^{1/2}$ denotes the principal square root of $\Sigma$ and

$$\bar{Z}^i = \frac{1}{M_i} \sum_{j=1}^{M_i} Z_j^i$$

is the empirical mean of the $Z_j^i$ over $j$. This representation expresses $\mu_i$ and $\Sigma_i$ as functions of the previous parameters and newly sampled standard Gaussian noise, facilitating the analysis of the sequence $\{\mathcal{P}_i\}$ as $i \to \infty$. We now observe that the matrix

$$\sum_{j=1}^{M_i} (Z_j^i - \bar{Z}^i)(Z_j^i - \bar{Z}^i)^\top$$

follows a Wishart distribution (Definition A.1.1). In our setting, the random vectors $Z_1^i, \ldots, Z_{M_i}^i$ are i.i.d. $\mathcal{N}_d(0, I)$. Invoking [Mui05, Theorem 3.1.2], which states that if

$$X_1, \ldots, X_n \overset{\text{i.i.d.}}{\sim} \mathcal{N}_d(0, \Sigma),$$

then the centred scatter matrix

$$S = \sum_{i=1}^{n} (X_i - \bar{X})(X_i - \bar{X})^\top$$

follows a Wishart distribution

$$S \sim \mathcal{W}_d(n-1, \Sigma),$$

we conclude that in our setting (with $\Sigma = I$ and $n = M_i$),

$$W_i := \sum_{j=1}^{M_i} (Z_j^i - \bar{Z}^i)(Z_j^i - \bar{Z}^i)^\top \sim \mathcal{W}_d(M_i - 1, I).$$

Here, the "$-1$" in the degrees of freedom accounts for the loss of one degree of freedom due to centring by the empirical mean $\bar{Z}^i$. Moreover, the matrices $W_1, W_2, \ldots$ are independent. Thus, we express the update of the covariance as

$$\Sigma_i = \frac{1}{M_i - 1} \Sigma_{i-1}^{1/2} W_i \Sigma_{i-1}^{1/2}.$$

We will define $k_i := M_i - 1$ and note that

$$\sum_{i=1}^{\infty} \frac{1}{k_i} < \infty \iff \sum_{i=1}^{\infty} \frac{1}{M_i} < \infty.$$

### 3.4.1 $\mu_i$ and $\Sigma_i$ are martingales

Recall from the previous construction that the update for the mean is given by

$$\mu_i = \mu_{i-1} + \Sigma_{i-1}^{1/2} \frac{1}{M_i} \sum_{j=1}^{M_i} Z_j^i,$$

where $Z_j^i \sim \mathcal{N}(0, I)$ are independent standard Gaussian random variables. Define the filtration $\mathcal{F}_i$ as the $\sigma$-algebra generated by all randomness up to step $i$

$$\mathcal{F}_i = \sigma \left( \left\{ Z_j^k \right\}_{1 \le k \le i, 1 \le j \le M_k} \right).$$

**Proposition 3.4.2.** *The sequence $\{\mu_i\}_{i \geq 0}$ is a martingale with respect to the filtration $\{\mathcal{F}_i\}$, i.e.,*

$$\mathbb{E}[\mu_i \mid \mathcal{F}_{i-1}] = \mu_{i-1}.$$

*Proof.* We first verify that $\mu_i$ is adapted and integrable:

- **Adaptedness:** By construction, $\mu_i$ is measurable with respect to $\mathcal{F}_i$, since it is obtained from $\mu_{i-1}$, $\Sigma_{i-1}$ and the newly sampled $Z_j^i$, which are $\mathcal{F}_i$-measurable.

- **Integrability:** Each $\mu_i$ is a linear combination of finitely many Gaussian random variables with finite second moments. Since $\Sigma_{i-1}$ is itself positive semidefinite and built through Wishart-distributed updates, an inductive argument shows that $\mathbb{E}[\|\mu_i\|] < \infty$.

We now prove the martingale property. Expanding $\mu_i$ using the update rule, we have

$$\mu_i = \mu_{i-1} + \Sigma_{i-1}^{1/2} \frac{1}{M_i} \sum_{j=1}^{M_i} Z_j^i.$$

Taking conditional expectation given $\mathcal{F}_{i-1}$ and using the independence and centring of $Z_j^i$, we obtain

$$\mathbb{E}\left[\mu_i \mid \mathcal{F}_{i-1}\right] = \mu_{i-1} + \Sigma_{i-1}^{1/2} \frac{1}{M_i} \sum_{j=1}^{M_i} \mathbb{E}\left[Z_j^i \mid \mathcal{F}_{i-1}\right] = \mu_{i-1} + \Sigma_{i-1}^{1/2} \frac{1}{M_i} \sum_{j=1}^{M_i} 0 = \mu_{i-1}.$$

Thus, $\{\mu_i\}$ is a martingale with respect to $\{\mathcal{F}_i\}$. $\qquad\square$

Next, we establish a similar martingale property for $\Sigma_i$.

**Definition 3.4.3.** The set $S_d$ denotes the space of $d \times d$ real symmetric matrices, and $S_d^+ \subset S_d$ denotes the cone of positive semidefinite matrices. The trace norm $\|A\|_{\mathrm{tr}}$ of a matrix $A \in S_d$ is defined as the sum of the absolute values of its eigenvalues:

$$\|A\|_{\mathrm{tr}} = \sum_{i=1}^{d} |\lambda_i|,$$

where $\lambda_1, \ldots, \lambda_d$ are the eigenvalues of $A$. Equipped with this norm, the space $S_d$ is a Banach space.

By construction, for each $i \geq 0$, the matrix $\Sigma_i \in S_d^+$. This follows inductively: $\Sigma_0$ is symmetric and positive semidefinite by assumption, and the recursive update

$$\Sigma_i = \frac{1}{k_i} \Sigma_{i-1}^{1/2} W_i \Sigma_{i-1}^{1/2}$$

preserves symmetry and positive semidefiniteness, since $W_i \sim \mathcal{W}_d(k_i, I)$ and $\Sigma_{i-1}^{1/2}$ are symmetric and positive semidefinite.

**Proposition 3.4.4.** *The sequence $\{\Sigma_i\}_{i \geq 0}$ is a martingale in the Banach space $(S_d, \|\cdot\|_{\mathrm{tr}})$ with respect to the filtration $\{\mathcal{F}_i\}$. In particular, $\{\mathrm{tr}(\Sigma_i)\}_{i \geq 0}$ is a real-valued martingale.*

*Proof.* We proceed in steps.

**Step 1: Adaptedness and integrability.** By construction, each $\Sigma_i$ is measurable with respect to $\mathcal{F}_i$. Moreover, we show by induction that

$$\mathbb{E}[\|\Sigma_i\|_{\mathrm{tr}}] = \mathbb{E}[\mathrm{tr}(\Sigma_i)] = \mathrm{tr}(\Sigma_0) < \infty.$$

Indeed, using the update rule

$$\Sigma_i = \frac{1}{k_i} \Sigma_{i-1}^{1/2} W_i \Sigma_{i-1}^{1/2},$$

and the cyclic property of the trace, we have

$$\text{tr}(\Sigma_i) = \frac{1}{k_i} \text{tr}(W_i \Sigma_{i-1}).$$

Taking expectations and using that $W_i$ and $\Sigma_{i-1}$ are independent, and that by Corollary A.2.3 $\mathbb{E}[W_i] = k_i I$, we obtain

$$\mathbb{E}[\text{tr}(\Sigma_i)] = \frac{1}{k_i} \text{tr}(\mathbb{E}[W_i]\mathbb{E}[\Sigma_{i-1}]) = \text{tr}(\mathbb{E}[\Sigma_{i-1}]).$$

Thus, by induction,

$$\mathbb{E}[\text{tr}(\Sigma_i)] = \text{tr}(\Sigma_0).$$

**Step 2: Martingale property.** We compute

$$\mathbb{E}[\Sigma_i \mid \mathcal{F}_{i-1}] = \frac{1}{k_i} \Sigma_{i-1}^{1/2} \mathbb{E}[W_i] \Sigma_{i-1}^{1/2}.$$

Since $W_i \sim \mathcal{W}_d(k_i, I)$, we have by Corollary A.2.3 that $\mathbb{E}[W_i] = k_i I$, so

$$\mathbb{E}[\Sigma_i \mid \mathcal{F}_{i-1}] = \Sigma_{i-1}.$$

Thus, $\{\Sigma_i\}$ is a martingale in $(S_d, \|\cdot\|_{\text{tr}})$.

**Step 3: Scalar martingale.** Since the trace is a linear operator, $\{\text{tr}(\Sigma_i)\}$ is also a martingale. □

### 3.4.2 Behaviour of $\Sigma_i$

We begin by analysing the behaviour of the sequence $\{\Sigma_i\}$, which evolves independently of $\{\mu_i\}$. Recall that $W_i \sim W_d(I, k_i)$ are independent Wishart matrices and that $\Sigma_0 \neq 0$ is a fixed deterministic, symmetric, positive semidefinite matrix. The sequence $\{\Sigma_n\}$ is defined recursively by

$$\Sigma_n = \frac{1}{k_n} \Sigma_{n-1}^{1/2} W_n \Sigma_{n-1}^{1/2}, \quad n = 1, 2, \dots$$

We need the following concept to show almost sure convergence of $\Sigma_i$.

**Definition 3.4.5.** A sequence of random symmetric matrices $(X_n)_{n \geq 0}$, adapted to a filtration $(\mathcal{F}_n)_{n \geq 0}$, is called a *matrix supermartingale* with respect to the Löwner order if it satisfies

$$\mathbb{E}[X_{n+1} \mid \mathcal{F}_n] \preceq X_n \quad \text{for all } n \geq 0,$$

where $A \preceq B$ denotes that $B - A$ is positive semidefinite.

**Proposition 3.4.6.** *The sequence $\Sigma_n$ converges almost surely to a positive semidefinite random matrix $\Sigma_\infty$.*

*Proof.* In Proposition 3.4.4 we already showed that $\Sigma_n$ is a martingale. Consequently, $\Sigma_n$ is also a matrix supermartingale with respect to the Löwner order. Almost sure convergence of $\Sigma_n$ to a positive semidefinite random matrix $\Sigma_\infty$ follows immediately from Theorem C.2.8. □

**Data growth is superlinear**

**Lemma 3.4.7.** *For all $n \in \mathbb{N}$ we have that*

$$\mathbb{E}[\text{tr}(\Sigma_n)^2] \leq \left( \prod_{i=1}^{\infty} \left( 1 + \frac{2}{k_i} \right) \right) \text{tr}(\Sigma_0)^2.$$

*Proof.* Set $Y_n = \text{tr}(\Sigma_n)$ for brevity. Using the recursive definition

$$\Sigma_n = \frac{1}{k_n} \Sigma_{n-1}^{1/2} W_n \Sigma_{n-1}^{1/2},$$

and the cyclicity of the trace, we get

$$Y_n = \frac{1}{k_n} \text{tr}(W_n \Sigma_{n-1}).$$

Since $W_n \sim W_d(I, k_n)$ is independent of $\mathcal{F}_{n-1}$, and $\Sigma_{n-1}$ is $\mathcal{F}_{n-1}$-measurable, by applying Lemma A.2.2, we obtain

$$\text{Var}(Y_n \mid \mathcal{F}_{n-1}) = \frac{2}{k_n} \text{tr}(\Sigma_{n-1}^2).$$

Now expand the second moment and use that $\mathbb{E}[Y_n \mid \mathcal{F}_{n-1}] = Y_{n-1}$ by Proposition 3.4.4:

$$\mathbb{E}[Y_n^2] = \mathbb{E}\left[\mathbb{E}[Y_n^2 \mid \mathcal{F}_{n-1}]\right] = \mathbb{E}\left[\text{Var}(Y_n \mid \mathcal{F}_{n-1}) + (\mathbb{E}[Y_n \mid \mathcal{F}_{n-1}])^2\right]$$

$$= \mathbb{E}\left[\frac{2}{k_n}\text{tr}(\Sigma_{n-1}^2) + Y_{n-1}^2\right] = \frac{2}{k_n}\mathbb{E}\left[\text{tr}(\Sigma_{n-1}^2)\right] + \mathbb{E}[Y_{n-1}^2].$$

Since $\Sigma_{n-1}$ is positive semidefinite, we have by Lemma C.2.14 that

$$\text{tr}(\Sigma_{n-1}^2) \leq (\text{tr}(\Sigma_{n-1}))^2 = Y_{n-1}^2,$$

thus

$$\mathbb{E}[Y_n^2] \leq \left(1 + \frac{2}{k_n}\right)\mathbb{E}[Y_{n-1}^2].$$

Iterating this inequality yields

$$\mathbb{E}[Y_n^2] \leq Y_0^2 \prod_{i=1}^{n}\left(1 + \frac{2}{k_i}\right) < Y_0^2 \prod_{i=1}^{\infty}\left(1 + \frac{2}{k_i}\right),$$

where $Y_0 = \text{tr}(\Sigma_0)$ is deterministic. $\qquad\square$

**Proposition 3.4.8.** *Assume $\sum_{i=1}^{\infty} \frac{1}{k_i} < \infty$. Then $\text{tr}(\Sigma_n)$ converges to $\text{tr}(\Sigma_\infty)$ in $L^2$. Furthermore, $\Sigma_\infty$ is not almost surely $0$.*

*Proof.* By Lemma C.2.5 the assumption $\sum_{i=1}^{\infty} \frac{1}{k_i} < \infty$ implies that

$$\prod_{i=1}^{\infty}\left(1 + \frac{2}{k_i}\right) < \infty$$

and thus Lemma 3.4.7 implies that $\{\text{tr}(\Sigma_n)\}$ is bounded in $L^2$. By Theorem C.2.4, $\text{tr}(\Sigma_n)$ converges in $L^2$ and by the uniqueness of limits and the continuity of the trace, the limit is $\text{tr}(\Sigma_\infty)$. This implies that

$$\mathbb{E}[\text{tr}(\Sigma_\infty)] = \lim_{n\to\infty} \mathbb{E}[\text{tr}(\Sigma_n)] = \text{tr}(\Sigma_0) > 0,$$

since $\Sigma_0$ is nonzero and positive semidefinite. Since $\Sigma_\infty$ is positive semidefinite as well, we have $\text{tr}(\Sigma_\infty) \geq 0$, and it follows that $\mathbb{P}(\text{tr}(\Sigma_\infty) = 0) < 1$, which implies $\mathbb{P}(\Sigma_\infty = 0) < 1$. $\qquad\square$

**Data growth is at most linear**

**Theorem 3.4.9.** *Assume $\sum_{i=1}^{\infty} \frac{1}{k_i} = \infty$. Then $\Sigma_n$ converges to $0$ almost surely.*

*Proof.* **1. Reduction to a product of $\chi^2$-random variables.** Write the spectral decomposition [Lan87, Ch. VIII, Thm. 4.4]

$$\Sigma_{n-1} = U_{n-1} \operatorname{diag}(\lambda_1^{n-1}, \ldots, \lambda_d^{n-1}) U_{n-1}^\top,$$

with $U_{n-1}$ orthogonal and $\lambda_j^{n-1} \geq 0$ the eigenvalues of $\Sigma_{n-1}$. By cyclic invariance of the trace,

$$\operatorname{tr}(\Sigma_{n-1}^{1/2} W_n \Sigma_{n-1}^{1/2}) = \operatorname{tr}(W_n \Sigma_{n-1}) = \operatorname{tr}\left(U_{n-1}^\top W_n U_{n-1} \operatorname{diag}(\lambda_1^{n-1}, \ldots, \lambda_d^{n-1})\right),$$

which expands to

$$\operatorname{tr}(W_n \Sigma_{n-1}) = \sum_{j=1}^d \lambda_j^{n-1} \left(U_{n-1}^\top W_n U_{n-1}\right)_{jj},$$

and thus

$$\operatorname{tr}\Sigma_n = \frac{1}{k_n} \sum_{j=1}^d \lambda_j^{n-1} \left(U_{n-1}^\top W_n U_{n-1}\right)_{jj}.$$

Since $W_n \sim W_d(I, k_n)$, we may represent it as $W_n = Z_n^\top Z_n$ with $Z_n$ having independent standard normal entries. Setting $\tilde{Z}_n := Z_n U_{n-1}$, the rotated matrix $\tilde{Z}_n$ also has i.i.d. $N(0,1)$ entries, and

$$U_{n-1}^\top W_n U_{n-1} = \tilde{Z}_n^\top \tilde{Z}_n.$$

In particular,

$$\left(U_{n-1}^\top W_n U_{n-1}\right)_{jj} = \left\|\tilde{Z}_n^{(j)}\right\|_2^2,$$

where $\tilde{Z}_n^{(j)}$ denotes the $j$-th column of $\tilde{Z}_n$. Each term $\left\|\tilde{Z}_n^{(j)}\right\|_2^2$ is independent and distributed as $\chi_{k_n}^2$. Thus, the diagonal entries $\left(U_{n-1}^\top W_n U_{n-1}\right)_{jj}$ are independent $\chi_{k_n}^2$-distributed random variables, and we obtain

$$\operatorname{tr}\Sigma_n = \frac{1}{k_n} \sum_{j=1}^d \lambda_j^{n-1} \chi_{k_n}^{2,(j)},$$

where the $\chi_{k_n}^{2,(j)}$ are independent $\chi^2$-variables with $k_n$ degrees of freedom. If $\Sigma_{n-1} = 0$, then $\Sigma_n = 0$ and the recursion remains identically zero from that point onward, so assume $\Sigma_{n-1} \neq 0$. Define

$$T_n := \sum_{j=1}^d \chi_{k_n}^{2,(j)} \sim \chi_{dk_n}^2, \quad w_j^{n-1} := \frac{\lambda_j^{n-1}}{\sum_m \lambda_m^{n-1}},$$

and set

$$R_n := d \sum_j w_j^{n-1} \frac{\chi_{k_n}^{2,(j)}}{T_n}.$$

Then

$$\operatorname{tr}\Sigma_n = \frac{1}{k_n} \sum_{j=1}^d \lambda_j^{n-1} \chi_{k_n}^{2,(j)} = \frac{1}{k_n} \operatorname{tr}(\Sigma_{n-1}) \sum_{j=1}^d w_j^{n-1} \chi_{k_n}^{2,(j)} = \operatorname{tr}(\Sigma_{n-1}) \frac{T_n}{dk_n} R_n,$$

and iterating from 1 to $n$ yields

$$\operatorname{tr}\Sigma_n = \operatorname{tr}\Sigma_0 \left(\prod_{i=1}^n \frac{T_i}{dk_i}\right) \left(\prod_{i=1}^n R_i\right).$$

**2. "Shrinking" product.** Since

$$\sum_{i=1}^\infty \frac{1}{dk_i} = \frac{1}{d} \sum_{i=1}^\infty \frac{1}{k_i} = \infty$$

and $T_i \sim \chi^2_{dk_i}$, Theorem C.1.1 ensures

$$\prod_{i=1}^{n} \frac{T_i}{dk_i} \xrightarrow{\text{a.s.}} 0.$$

**3. "Martingale" product.** The random vector

$$\left( \frac{\chi^{2,(1)}_{k_n}}{T_n}, \ldots, \frac{\chi^{2,(d)}_{k_n}}{T_n} \right)$$

is independent of $\mathcal{F}_{n-1}$, where $\mathcal{F}_n := \sigma(W_1, \ldots, W_n)$, and follows a Dirichlet distribution with parameters $(k_n/2, \ldots, k_n/2)$. In particular, the entries satisfy

$$\mathbb{E}\left[ \frac{\chi^{2,(j)}_{k_n}}{T_n} \right] = \frac{1}{d} \quad \text{for all } j = 1, \ldots, d.$$

Hence,

$$\mathbb{E}[R_n \mid \mathcal{F}_{n-1}] = d \sum_{j=1}^{d} w_j^{n-1} \mathbb{E}\left[ \frac{\chi^{2,(j)}_{k_n}}{T_n} \Big| \mathcal{F}_{n-1} \right] = d \sum_{j=1}^{d} w_j^{n-1} \frac{1}{d} = \sum_{j=1}^{d} w_j^{n-1} = 1,$$

because the weights $w_j^{n-1}$ sum to one. Thus, setting

$$M_n := \prod_{i=1}^{n} R_i,$$

the sequence $(M_n)_{n \geq 1}$ is a nonnegative martingale with respect to the filtration $(\mathcal{F}_n)$. By Theorem C.2.3, the martingale $M_n$ converges almost surely to a finite random variable $M_\infty \geq 0$.

**4. Limit of** $\operatorname{tr} \Sigma_n$. Combining the two factors,

$$\operatorname{tr} \Sigma_n = \operatorname{tr} \Sigma_0 \left[ \prod_{i=1}^{n} \frac{T_i}{d} \right] \left[ \prod_{i=1}^{n} R_i \right] \xrightarrow{\text{a.s.}} 0 \cdot M_\infty = 0.$$

This implies that $\Sigma_\infty = 0$ almost surely. $\qquad \square$

**Corollary 3.4.10** (Uniform exponential decay of $\operatorname{tr}(\Sigma_n)$). *Assume*

$$\sum_{i=1}^{\infty} \frac{1}{k_i} = \infty, \qquad S_n := \sum_{i=1}^{n} \frac{1}{k_i}.$$

*Then there exists an almost surely finite random constant $C = C(\omega)$ such that*

$$\operatorname{tr}(\Sigma_n)(\omega) \leq C(\omega) e^{-S_n/(2d)} \qquad \text{for all } n \in \mathbb{N}.$$

*Proof.* From the decomposition in the backwards direction proof,

$$\operatorname{tr}(\Sigma_n) = \operatorname{tr}(\Sigma_0) \cdot \prod_{i=1}^{n} \frac{T_i}{dk_i} \cdot \prod_{i=1}^{n} R_i,$$

where $T_i \sim \chi^2_{dk_i}$ are independent and $R_i$ is a nonnegative martingale factor satisfying $\mathbb{E}[R_i \mid \mathcal{F}_{i-1}] = 1$. By Theorem C.1.1 applied to $T_i \sim \chi^2_{dk_i}$, the product

$$\prod_{i=1}^{n} \frac{T_i}{dk_i} \xrightarrow{\text{a.s.}} 0$$

and satisfies the exponential bound: there exists an a.s. finite constant $C_1(\omega)$ such that

$$\prod_{i=1}^{n} \frac{T_i}{dk_i} \leq C_1(\omega)e^{-S_n/(2d)} \quad \text{for all } n \in \mathbb{N}.$$

From the proof above, $M_n := \prod_{i=1}^{n} R_i$ is a nonnegative martingale and thus converges a.s. to some finite $M_\infty \geq 0$. Let

$$C_2(\omega) := \sup_n M_n(\omega) = \sup_n \prod_{i=1}^{n} R_i(\omega) < \infty \quad \text{a.s.}$$

Combining, we obtain

$$\text{tr}(\Sigma_n)(\omega) \leq \text{tr}(\Sigma_0) \cdot C_1(\omega) \cdot C_2(\omega) \cdot e^{-S_n/(2d)} = C(\omega)e^{-S_n/(2d)},$$

where $C(\omega) := \text{tr}(\Sigma_0)C_1(\omega)C_2(\omega)$ is a.s. finite. □

### 3.4.3 Behaviour of $\mu_i$

We next look at the behaviour of $\mu_i$. Recall that

$$\mu_i = \mu_{i-1} + \Sigma_{i-1}^{1/2} \frac{1}{M_i} \sum_{j=1}^{M_i} Z_j^i.$$

**Lemma 3.4.11.** *The conditional variance of the mean update is given by*

$$\text{Var}(\mu_i \mid \mathcal{F}_{i-1}) = \frac{\Sigma_{i-1}}{M_i}.$$

*Proof.* Since $Z_j^i$ are i.i.d. $\mathcal{N}(0, I)$ and independent of $\mathcal{F}_{i-1}$, the conditional variance is

$$\text{Var}\left(\mu_i \mid \mathcal{F}_{i-1}\right) = \text{Var}\left(\mu_{i-1} + \Sigma_{i-1}^{1/2} \frac{1}{M_i} \sum_{j=1}^{M_i} Z_j^i \Big| \mathcal{F}_{i-1}\right) = \Sigma_{i-1}^{1/2} \text{Var}\left(\frac{1}{M_i} \sum_{j=1}^{M_i} Z_j^i\right) \Sigma_{i-1}^{1/2}$$

$$= \Sigma_{i-1}^{1/2} \left(\frac{1}{M_i} I\right) \Sigma_{i-1}^{1/2} = \frac{\Sigma_{i-1}}{M_i}.$$ □

Let us define $S_i := \sum_{k=1}^{i} \frac{1}{M_k}$ and $S_\infty := \sum_{k=1}^{\infty} \frac{1}{M_k}$.

**Proposition 3.4.12.** *The variance of $\mu_i$ evolves as*

$$\text{Var}(\mu_i) = \Sigma_0 \sum_{k=1}^{i} \frac{1}{M_k} = \Sigma_0 S_i.$$

*Proof.* Applying the law of total variance and the martingale property Proposition 3.4.2 of $\{\mu_i\}$, we have

$$\text{Var}(\mu_i) = \mathbb{E}\left[\text{Var}(\mu_i \mid \mathcal{F}_{i-1})\right] + \text{Var}\left(\mathbb{E}[\mu_i \mid \mathcal{F}_{i-1}]\right) = \mathbb{E}\left[\text{Var}(\mu_i \mid \mathcal{F}_{i-1})\right] + \text{Var}(\mu_{i-1}),$$

where we used $\mathbb{E}[\mu_i \mid \mathcal{F}_{i-1}] = \mu_{i-1}$. Iterating this recursion and noting $\text{Var}(\mu_0) = 0$, we obtain

$$\text{Var}(\mu_i) = \sum_{k=1}^{i} \mathbb{E}\left[\text{Var}(\mu_k \mid \mathcal{F}_{k-1})\right].$$

Using Lemma 3.4.11, we substitute

$$\text{Var}(\mu_k \mid \mathcal{F}_{k-1}) = \frac{\Sigma_{k-1}}{M_k}.$$

Now, because $\{\Sigma_i\}$ is a martingale by Proposition 3.4.4,

$$\text{Var}(\mu_i) = \sum_{k=1}^{i} \frac{1}{M_k} \Sigma_0 = \Sigma_0 \sum_{k=1}^{i} \frac{1}{M_k}.$$ □

**Proposition 3.4.13.** *The squared $L^2$-distance of $\mu_i$ and $\mu_0$ is*

$$\mathbb{E}\left[\|\mu_i - \mu_0\|_2^2\right] = \text{Tr}(\Sigma_0)\sum_{k=1}^{i}\frac{1}{M_k} = \text{Tr}(\Sigma_0)S_i.$$

*Proof.* Since $\mathbb{E}[\mu_i] = \mu_0$, we have

$$\mathbb{E}\left[\|\mu_i - \mu_0\|_2^2\right] = \mathbb{E}\left[\|\mu_i - \mathbb{E}[\mu_i]\|_2^2\right].$$

Writing the squared norm as a trace,

$$\|\mu_i - \mathbb{E}[\mu_i]\|_2^2 = \text{Tr}\left((\mu_i - \mathbb{E}[\mu_i])(\mu_i - \mathbb{E}[\mu_i])^\top\right),$$

and taking the expectation yields

$$\mathbb{E}\left[\|\mu_i - \mathbb{E}[\mu_i]\|_2^2\right] = \text{Tr}\left(\text{Var}(\mu_i)\right).$$

Using Proposition 3.4.12,

$$\text{Var}(\mu_i) = \Sigma_0\sum_{k=1}^{i}\frac{1}{M_k},$$

thus

$$\text{Tr}(\text{Var}(\mu_i)) = \text{Tr}(\Sigma_0)\sum_{k=1}^{i}\frac{1}{M_k}. \qquad \square$$

**Corollary 3.4.14.** *The squared $L^2$-norm of $\mu_i$ is given by*

$$\mathbb{E}[\|\mu_i\|_2^2] = \|\mu_0\|_2^2 + \text{Tr}(\Sigma_0)S_i.$$

*Proof.* By the identity

$$\|\mu_i\|_2^2 = \|\mu_i - \mu_0 + \mu_0\|_2^2 = \|\mu_i - \mu_0\|_2^2 + 2\langle\mu_i - \mu_0, \mu_0\rangle + \|\mu_0\|_2^2,$$

and taking the expectation, we obtain

$$\mathbb{E}[\|\mu_i\|_2^2] = \mathbb{E}[\|\mu_i - \mu_0\|_2^2] + 2\mathbb{E}[\langle\mu_i - \mu_0, \mu_0\rangle] + \|\mu_0\|_2^2.$$

Since $\mathbb{E}[\mu_i - \mu_0] = 0$, the cross term vanishes

$$\mathbb{E}[\langle\mu_i - \mu_0, \mu_0\rangle] = \langle\mathbb{E}[\mu_i - \mu_0], \mu_0\rangle = 0.$$

Thus,

$$\mathbb{E}[\|\mu_i\|_2^2] = \mathbb{E}[\|\mu_i - \mu_0\|_2^2] + \|\mu_0\|_2^2.$$

Using Proposition 3.4.13, we substitute

$$\mathbb{E}[\|\mu_i - \mu_0\|_2^2] = \text{tr}(\Sigma_0)S_i,$$

which yields the desired result:

$$\mathbb{E}[\|\mu_i\|_2^2] = \|\mu_0\|_2^2 + \text{tr}(\Sigma_0)S_i. \qquad \square$$

**Theorem 3.4.15.** *Assume that*

$$S_\infty := \sum_{k=1}^{\infty}\frac{1}{M_k} < \infty.$$

*Then the sequence $(\mu_i)_{i\geq 0}$ converges both almost surely and in $L^2$ to a random vector $\mu_\infty$.*

*Proof.* Recall from Proposition 3.4.2 that $(\mu_i)_{i \geq 0}$ is a martingale adapted to the filtration $(\mathcal{F}_i)$, and that by Corollary 3.4.14,

$$\mathbb{E}\left[\|\mu_i\|_2^2\right] = \|\mu_0\|_2^2 + \operatorname{tr}(\Sigma_0)S_i.$$

Since $S_i \leq S_\infty$ for all $i$, it follows that

$$\sup_{i \geq 0} \mathbb{E}\left[\|\mu_i\|_2^2\right] \leq \|\mu_0\|_2^2 + \operatorname{tr}(\Sigma_0)S_\infty < \infty.$$

Thus $(\mu_i)_{i \geq 0}$ is an $L^2$-bounded martingale. Now, write each vector $\mu_i$ in coordinates:

$$\mu_i = (\mu_i^{(1)}, \ldots, \mu_i^{(d)}).$$

Each coordinate sequence $(\mu_i^{(k)})_{i \geq 0}$ is itself a real-valued martingale (since conditional expectations and measurability properties hold coordinate-wise) and is $L^2$-bounded because

$$\sup_{i \geq 0} \mathbb{E}\left[|\mu_i^{(k)}|^2\right] \leq \sup_{i \geq 0} \mathbb{E}\left[\|\mu_i\|_2^2\right] < \infty.$$

Applying Theorem C.2.4 with $p = 2$ to each coordinate martingale $(\mu_i^{(k)})_{i \geq 0}$, we conclude that for each $k = 1, \ldots, d$, there exists a random variable $\mu_\infty^{(k)} \in L^2$ such that

$$\mu_i^{(k)} \xrightarrow[i \to \infty]{\text{a.s. and in } L^2} \mu_\infty^{(k)}.$$

Define the random vector

$$\mu_\infty := (\mu_\infty^{(1)}, \ldots, \mu_\infty^{(d)}).$$

Since convergence holds in each coordinate, it follows that

$$\|\mu_i - \mu_\infty\|_2^2 = \sum_{k=1}^d |\mu_i^{(k)} - \mu_\infty^{(k)}|^2 \xrightarrow[i \to \infty]{\text{a.s.}} 0,$$

and similarly

$$\mathbb{E}\left[\|\mu_i - \mu_\infty\|_2^2\right] = \sum_{k=1}^d \mathbb{E}\left[|\mu_i^{(k)} - \mu_\infty^{(k)}|^2\right] \xrightarrow[i \to \infty]{} 0,$$

because convergence holds term-by-term in each coordinate. Thus, $\mu_i \to \mu_\infty$ both almost surely and in $L^2$. $\square$

**Proposition 3.4.16.** *If $\sum_i 1/M_i = \infty$, we have almost surely*

$$\sum_{i=1}^\infty \frac{\operatorname{tr}\Sigma_{i-1}}{M_i} < \infty.$$

*Proof.* By Corollary 3.4.10, there exists an almost surely finite random constant $C(\omega)$ such that

$$\operatorname{tr}(\Sigma_{i-1}) \leq C(\omega)e^{-S_{i-1}/(2d)}, \qquad S_n := \sum_{j=1}^n \frac{1}{k_j}.$$

Moreover, since $M_i = k_i + 1 > k_i$ for all $i$, we obtain

$$\frac{1}{M_i} < \frac{1}{k_i} = S_i - S_{i-1},$$

so

$$\sum_{i=1}^\infty \frac{\operatorname{tr}(\Sigma_{i-1})}{M_i} < C(\omega) \sum_{i=1}^\infty e^{-S_{i-1}/(2d)}(S_i - S_{i-1}).$$

For every $s \in [S_{i-1}, S_i]$ we have for $f(s) := e^{-s/(2d)}$

$$f(S_{i-1}) = f(s) \cdot e^{(s-S_{i-1})/(2d)} \leq f(s) \cdot e^{1/(2d)},$$

since $s - S_{i-1} \leq S_i - S_{i-1} = \frac{1}{k_i} \leq 1$. Hence

$$f(S_{i-1})(S_i - S_{i-1}) = \int_{S_{i-1}}^{S_i} f(S_{i-1})ds \leq e^{1/(2d)} \int_{S_{i-1}}^{S_i} f(s)ds.$$

Summing over $i$ and using the telescoping property of the integral,

$$\sum_{i=1}^{\infty} f(S_{i-1})(S_i - S_{i-1}) \leq e^{1/(2d)} \int_0^{\infty} f(s)ds = e^{1/(2d)} 2d.$$

Therefore

$$\sum_{i=1}^{\infty} \frac{\text{tr}(\Sigma_{i-1})}{M_i} < C(\omega)e^{1/(2d)} 2d < \infty \qquad \text{a.s..} \qquad \square$$

We now prove the following:

**Proposition 3.4.17.** *If $S_\infty := \sum_{i=1}^{\infty} \frac{1}{M_i} = \infty$, then the mean sequence $(\mu_i)_{i \geq 0}$ converges almost surely.*

*Proof.* By Lemma 3.4.11, we have

$$\text{Var}(\mu_i \mid \mathcal{F}_{i-1}) = \frac{\Sigma_{i-1}}{M_i}.$$

Fix any unit vector $v \in \mathbb{R}^d$. Then $X_i := v^\top \mu_i$ is a scalar martingale with conditional variance

$$\text{Var}(X_i \mid \mathcal{F}_{i-1}) = v^\top \text{Var}(\mu_i \mid \mathcal{F}_{i-1})v = \frac{v^\top \Sigma_{i-1} v}{M_i} \leq \frac{\text{tr}(\Sigma_{i-1})}{M_i}.$$

Hence, by Proposition 3.4.16,

$$\sum_{i=1}^{\infty} \text{Var}(X_i \mid \mathcal{F}_{i-1}) < \infty \quad \text{a.s.,}$$

so $X_i = v^\top \mu_i$ converges almost surely by Corollary C.2.13. Taking $v$ to be each coordinate vector $e_\ell (\ell = 1, \ldots, d)$ shows that every component of $\mu_i$ converges a.s.; this implies that $\mu_i$ itself converges almost surely. $\square$

### 3.4.4 Behaviour of the Wasserstein distance

**Fact 3.4.18.** *Let $\mathcal{N}(\mu_1, \Sigma_1)$ and $\mathcal{N}(\mu_2, \Sigma_2)$ be Gaussian distributions. Then the squared Wasserstein-2 distance between them is given by*

$$W_2^2(\mathcal{N}(\mu_1, \Sigma_1), \mathcal{N}(\mu_2, \Sigma_2)) = \|\mu_1 - \mu_2\|_2^2 + \text{Tr}\left(\Sigma_1 + \Sigma_2 - 2\left(\Sigma_1^{1/2} \Sigma_2 \Sigma_1^{1/2}\right)^{1/2}\right).$$

As a consequence, the squared Wasserstein distance between the distributions $\mathcal{P}_0 = \mathcal{N}(\mu_0, \Sigma_0)$ and $\mathcal{P}_i = \mathcal{N}(\mu_i, \Sigma_i)$ satisfies the inequality:

$$\mathbb{E}\left[W_2^2(\mathcal{P}_0, \mathcal{P}_i)\right] \geq \mathbb{E}\left[\|\mu_i - \mu_0\|_2^2\right] = \text{tr}(\Sigma_0)S_i,$$

where the equality follows from Proposition 3.4.13. We now use this lower bound to prove the following.

**Proposition 3.4.19.** *If $S_\infty = \infty$, then neither $(\mu_i)$ nor $(W_2(\mathcal{P}_0, \mathcal{P}_i))$ converge in $L^2$.*

*Proof.* From Corollary 3.4.14, we know that

$$\mathbb{E}\left[\|\mu_i\|_2^2\right] = \|\mu_0\|_2^2 + \mathrm{tr}(\Sigma_0)S_i.$$

Since $S_\infty = \sum_{k=1}^\infty \frac{1}{M_k} = \infty$, it follows that

$$\lim_{i\to\infty} \mathbb{E}\left[\|\mu_i\|_2^2\right] = \infty,$$

so $(\mu_i)$ does not converge in $L^2$. Now consider the Wasserstein distance. By Fact 3.4.18 we have the lower bound

$$\mathbb{E}\left[W_2^2(\mathcal{P}_0, \mathcal{P}_i)\right] \geq \mathbb{E}\left[\|\mu_i - \mu_0\|_2^2\right] = \mathrm{tr}(\Sigma_0)S_i.$$

Again, since $S_i \to \infty$, the expected squared Wasserstein distance diverges,

$$\lim_{i\to\infty} \mathbb{E}\left[W_2^2(\mathcal{P}_0, \mathcal{P}_i)\right] = \infty.$$

Therefore, $(W_2(\mathcal{P}_0, \mathcal{P}_i))$ does not converge in $L^2$. $\qquad\square$

### 3.4.5 Experimental verification

We now present a numerical experiment that illustrates and confirms the dichotomy stated in Theorem 3.4.1.

**General setup**   All simulations start with an initial Gaussian distribution $\mathcal{P}_0 = \mathcal{N}(\mu_0, \Sigma_0)$, where $\mu_0 \in \mathbb{R}^d$ is sampled from a standard Gaussian and $\Sigma_0$ is constructed as $\Sigma_0 = AA^\top$ for a random matrix $A \in \mathbb{R}^{d\times d}$, normalized so that $\mathrm{tr}(\Sigma_0) = 1$. The iterative process then follows the procedure defined earlier. At each step $i$, a set of $M_i$ independent samples is drawn from $\mathcal{P}_{i-1}$, and used to compute $\mu_i$ and $\Sigma_i$ as the empirical mean and unbiased sample covariance. We consider four sample size schedules:

- Constant: $M_i = 20$

- Logarithmic: $M_i = 20 + \lfloor\log(\max(i,1))\rfloor$

- Linear: $M_i = 20 + i$

- $n\log n$: $M_i = 20 + \lfloor i\log(\max(i,1))\rfloor$

The first three schedules result in a divergent sum $S_\infty = \sum_i 1/M_i = \infty$, while the last one ensures convergence of this series, i.e., $S_\infty < \infty$. According to Theorem 3.4.1, this difference determines whether the limiting covariance matrix $\Sigma_\infty$ is almost surely degenerate.

All results are averaged over 500 independent replicates, each running for $n = 2000$ iterations in dimension $d = 5$.

**Convergence behaviour of $\Sigma_i$:**   Figs. 3.2a and 3.2b display the empirical probability that the trace of the covariance matrix $\mathrm{tr}(\Sigma_i)$ falls below a small threshold—specifically, 0.01 and 0.05 respectively—as a function of the iteration index $i$. These plots illustrate the dichotomy described in Theorem 3.4.1. For the constant, logarithmic, and linear sampling schedules (where $S_\infty = \infty$), the probability of collapse to near-zero trace increases with iteration and ultimately approaches 1 for the constant and logarithmic cases, indicating almost sure degeneration of $\Sigma_\infty$. In contrast, the $n\log n$ schedule (where $S_\infty < \infty$) remains bounded away from 1, with nearly zero probability of degeneration even after 2000 iterations, consistent with a non-degenerate limiting covariance. The distinction becomes even sharper in the stricter threshold of 0.01, where only the divergent schedules show a high probability of collapse. This experimental evidence corroborates the theoretical dichotomy and highlights how the choice of sample size schedule governs the stability and degeneracy of the fitted distributions.

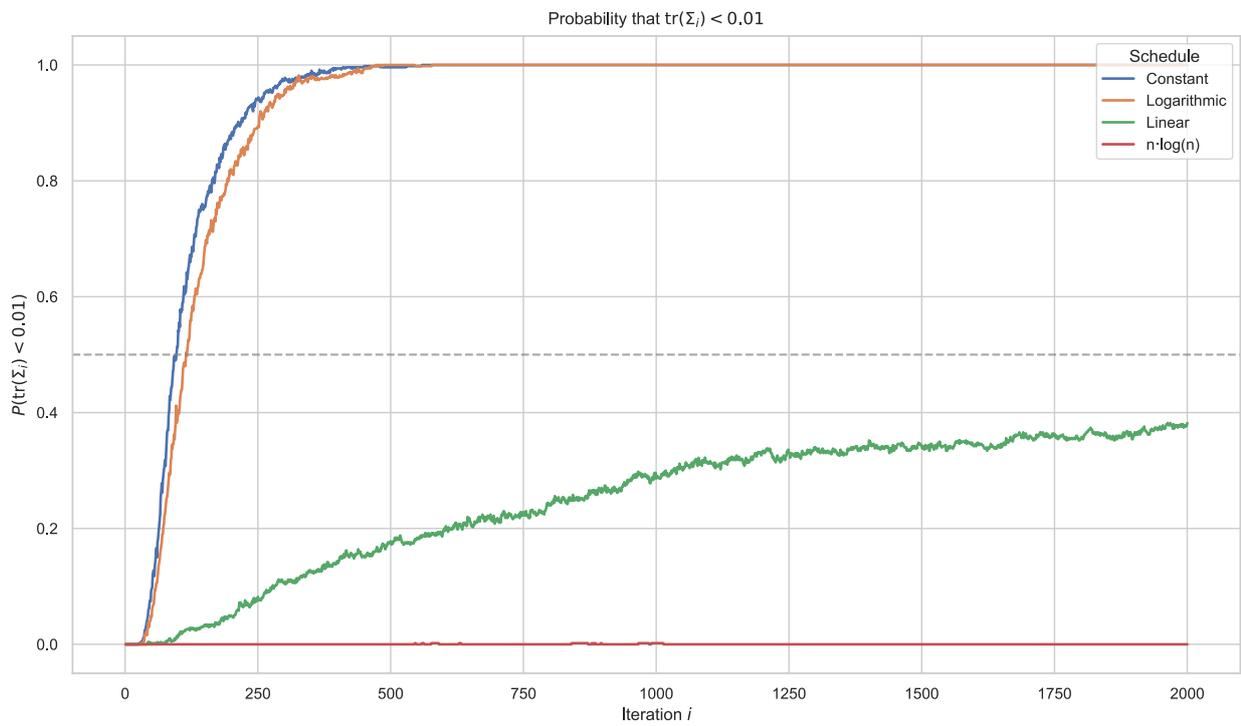

**(a)** Probability that $\mathrm{tr}(\Sigma_i) < 0.01$

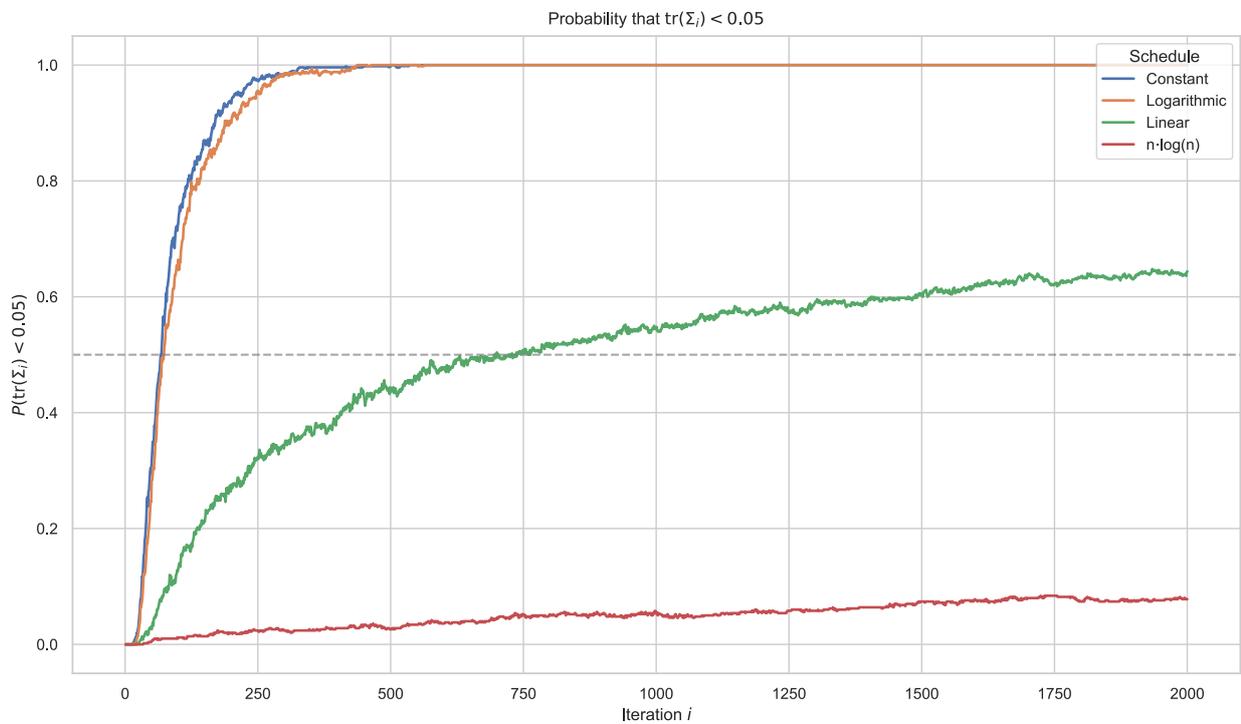

**(b)** Probability that $\mathrm{tr}(\Sigma_i) < 0.05$

**Figure 3.2** Empirical probability that the covariance trace falls below a threshold, shown for different sampling schedules. Each curve is averaged over 500 trials.

## 3.5 Cumulative Gaussian fitting

In this section, we study a variant of Gaussian parameter estimation in which data is accumulated over time, as has been studied by Kazdan et al. [Kaz+25] for $d = 1$. We generalise their discussion to all $d$. Unlike the *replacing−data* approach, which discards past samples at each iteration, we instead fit parameters using *all* data observed so far. For clarity, we focus on maximum likelihood estimation (MLE) rather than unbiased estimation, as the MLE yields more concise expressions.

Since the setup largely mirrors the replacing-data case, we keep the exposition brief and emphasise the key differences. Let $d \in \mathbb{N}$ be fixed. At each iteration $i \geq 1$, we receive a new mini-batch of samples

$$p_1^i, \ldots, p_M^i \overset{\text{i.i.d.}}{\sim} \mathcal{N}(\mu_{i-1}, \Sigma_{i-1}),$$

with fixed batch size $M \geq 2$. The total number of samples available after iteration $i$ is

$$N_i := iM.$$

Throughout this section, we let $\mathcal{F}_i$ denote the $\sigma$-field generated by all randomness observed up to and including step $i$. We now define the maximum likelihood estimators for the mean and covariance matrix based on the full dataset up to iteration $i$.

**Definition 3.5.1** (Maximum likelihood estimators for mean and covariance). Given observations $p_1^k, \ldots, p_M^k$ for $k = 1, \ldots, i$, the maximum likelihood estimators for the mean and covariance are given by:

$$\mu_i := \frac{1}{N_i} \sum_{k=1}^{i} \sum_{j=1}^{M} p_j^k, \tag{3.2}$$

$$\Sigma_i^{(\text{MLE})} := \frac{1}{N_i} \sum_{k=1}^{i} \sum_{j=1}^{M} \left(p_j^k - \mu_i\right) \left(p_j^k - \mu_i\right)^\top. \tag{3.3}$$

**Proposition 3.5.2** (Alternative representations of $\mu_i$). *The following identities hold:*

1. *Batch-mean representation:*

$$\mu_i = \frac{1}{i} \sum_{k=1}^{i} \bar{p}^k, \quad \text{where} \quad \bar{p}^k := \frac{1}{M} \sum_{j=1}^{M} p_j^k.$$

2. *Recursive averaging:*

$$\mu_i = (1 - \alpha_i)\mu_{i-1} + \alpha_i \bar{p}^i, \quad \text{where} \quad \alpha_i := \frac{1}{i}.$$

3. *Recursive update via mean shift:*

$$\mu_i = \mu_{i-1} - \alpha_i \Delta_i, \quad \text{with} \quad \Delta_i := \mu_{i-1} - \bar{p}^i.$$

*Proof.* **Item 1:** Using the definition of the empirical batch mean $\bar{p}^k$, we have:

$$\mu_i = \frac{1}{N_i} \sum_{k=1}^{i} \sum_{j=1}^{M} p_j^k = \frac{1}{iM} \sum_{k=1}^{i} \sum_{j=1}^{M} p_j^k = \frac{1}{i} \sum_{k=1}^{i} \left(\frac{1}{M} \sum_{j=1}^{M} p_j^k\right) = \frac{1}{i} \sum_{k=1}^{i} \bar{p}^k.$$

**Item 2:** From **Item 1**, write

$$\mu_i = \frac{1}{i} \sum_{k=1}^{i-1} \bar{p}^k + \frac{1}{i} \bar{p}^i = (1 - \alpha_i)\, \mu_{i-1} + \alpha_i \bar{p}^i,$$

since $\mu_{i-1} = \frac{1}{i-1} \sum_{k=1}^{i-1} \bar{p}^k$, and the coefficient matches after factoring.

**Item 3:** The update in **Item 2** can be rewritten as

$$\mu_i = \mu_{i-1} - \alpha_i(\mu_{i-1} - \bar{p}^i) = \mu_{i-1} - \alpha_i \Delta_i,$$

using the definition $\Delta_i = \mu_{i-1} - \bar{p}^i$. $\qquad \square$

The empirical covariance defined in (3.3) can also be expressed recursively. This formulation separates contributions from the within-batch variation and the drift in the batch mean.

**Proposition 3.5.3** (Recursive MLE covariance update). *Let us define the within-batch empirical covariance as*

$$V_i := \frac{1}{M} \sum_{j=1}^{M} (p_j^i - \bar{p}^i)(p_j^i - \bar{p}^i)^\top.$$

*Then the cumulative covariance satisfies the recursion*

$$\Sigma_i^{(\text{MLE})} = (1 - \alpha_i)\Sigma_{i-1} + \alpha_i V_i + \alpha_i(1 - \alpha_i)\Delta_i \Delta_i^\top. \tag{3.4}$$

*Proof.* Starting from the definition (3.3), split the total sum into past and current batches

$$\Sigma_i^{(\text{MLE})} = \frac{1}{N_i} \left[ \sum_{k=1}^{i-1} \sum_{j=1}^{M} (p_j^k - \mu_i)(p_j^k - \mu_i)^\top + \sum_{j=1}^{M} (p_j^i - \mu_i)(p_j^i - \mu_i)^\top \right].$$

For the first term, use the identity $p_j^k - \mu_i = (p_j^k - \mu_{i-1}) + \alpha_i \Delta_i$, and note that the cross-terms vanish due to zero-mean summation. This yields

$$\sum_{k=1}^{i-1} \sum_{j=1}^{M} (p_j^k - \mu_i)(p_j^k - \mu_i)^\top = N_{i-1} \left( \Sigma_{i-1} + \alpha_i^2 \Delta_i \Delta_i^\top \right).$$

For the second term, apply the identity $p_j^i - \mu_i = (p_j^i - \bar{p}^i) - (1 - \alpha_i)\Delta_i$. Expanding and again discarding cross-terms yields

$$\sum_{j=1}^{M} (p_j^i - \mu_i)(p_j^i - \mu_i)^\top = MV_i + M(1 - \alpha_i)^2 \Delta_i \Delta_i^\top.$$

Combining both contributions and dividing by $N_i = iM$ gives

$$\begin{aligned}
\Sigma_i^{(\text{MLE})} &= \frac{1}{N_i} \left( N_{i-1}\Sigma_{i-1} + N_{i-1}\alpha_i^2 \Delta_i \Delta_i^\top + MV_i + M(1 - \alpha_i)^2 \Delta_i \Delta_i^\top \right) \\
&= (1 - \alpha_i)\Sigma_{i-1} + \alpha_i V_i + \left[ (1 - \alpha_i)\alpha_i^2 + \alpha_i(1 - \alpha_i)^2 \right] \Delta_i \Delta_i^\top \\
&= (1 - \alpha_i)\Sigma_{i-1} + \alpha_i V_i + \alpha_i(1 - \alpha_i)\Delta_i \Delta_i^\top,
\end{aligned}$$

where in the last step we used $(1 - \alpha_i)\alpha_i^2 + \alpha_i(1 - \alpha_i)^2 = \alpha_i(1 - \alpha_i)(\alpha_i + 1 - \alpha_i) = \alpha_i(1 - \alpha_i)$. This establishes the recursion (3.4). □

### 3.5.1 Martingale and supermartingale properties

In this subsection, we study the evolution of the mean and covariance estimators as stochastic processes adapted to the filtration $\mathcal{F}_i$. We show that the sequence of sample means $(\mu_i)_{i \geq 1}$ forms a martingale, while the sequence of MLE covariances $(\Sigma_i^{(\text{MLE})})_{i \geq 1}$ is a supermartingale with respect to the same filtration.

**Proposition 3.5.4** (Mean estimator is a martingale). *The sequence $(\mu_i)_{i \geq 1}$ defined in Proposition 3.5.2 is a martingale with respect to the filtration $(\mathcal{F}_i)_{i \geq 1}$, i.e.,*

$$\mathbb{E}[\mu_i \mid \mathcal{F}_{i-1}] = \mu_{i-1}, \quad \text{for all } i \geq 1.$$

*Proof.* From the recursive update

$$\mu_i = \mu_{i-1} - \alpha_i \Delta_i,$$

and the definition $\Delta_i := \mu_{i-1} - \bar{p}^i$, we have

$$\mathbb{E}[\mu_i \mid \mathcal{F}_{i-1}] = \mu_{i-1} - \alpha_i \mathbb{E}[\Delta_i \mid \mathcal{F}_{i-1}] = \mu_{i-1} - \alpha_i(\mu_{i-1} - \mathbb{E}[\bar{p}^i \mid \mathcal{F}_{i-1}]).$$

Since the batch samples $p^i_j \sim \mathcal{N}(\mu_{i-1}, \Sigma_{i-1})$ are i.i.d. conditioned on $\mathcal{F}_{i-1}$, we have $\mathbb{E}[\bar{p}^i \mid \mathcal{F}_{i-1}] = \mu_{i-1}$. Therefore,

$$\mathbb{E}[\mu_i \mid \mathcal{F}_{i-1}] = \mu_{i-1},$$

confirming the martingale property. $\qquad \square$

**Proposition 3.5.5** (Covariance estimator is a supermartingale). *The sequence $(\Sigma_i^{(\text{MLE})})_{i \geq 1}$ is a matrix-valued supermartingale with respect to $(\mathcal{F}_i)$. In particular, for all $i \geq 1$,*

$$\mathbb{E}\left[\Sigma_i^{(\text{MLE})} \mid \mathcal{F}_{i-1}\right] = \left(1 - \frac{1}{Mi^2}\right)\Sigma_{i-1}^{(\text{MLE})}. \tag{3.5}$$

*Consequently, the scalar trace process $\operatorname{tr}(\Sigma_i)$ is a nonnegative real supermartingale.*

*Proof.* Using the recursion from Proposition 3.5.3:

$$\Sigma_i^{(\text{MLE})} = (1 - \alpha_i)\Sigma_{i-1} + \alpha_i V_i + \alpha_i(1 - \alpha_i)\Delta_i \Delta_i^\top,$$

we take conditional expectations given $\mathcal{F}_{i-1}$, using $\alpha_i = 1/i$ and the identities

$$\mathbb{E}\left[V_i \mid \mathcal{F}_{i-1}\right] = \left(1 - \frac{1}{M}\right)\Sigma_{i-1}, \quad \mathbb{E}\left[\Delta_i \Delta_i^\top \mid \mathcal{F}_{i-1}\right] = \frac{1}{M}\Sigma_{i-1}.$$

Plugging in, we get

$$\mathbb{E}\left[\Sigma_i^{(\text{MLE})} \mid \mathcal{F}_{i-1}\right] = (1 - \alpha_i)\Sigma_{i-1} + \alpha_i\left(1 - \frac{1}{M}\right)\Sigma_{i-1} + \alpha_i(1 - \alpha_i)\frac{1}{M}\Sigma_{i-1}$$

$$= \left[(1 - \alpha_i) + \alpha_i\left(1 - \frac{1}{M}\right) + \alpha_i(1 - \alpha_i)\frac{1}{M}\right]\Sigma_{i-1}.$$

Simplifying the coefficient

$$(1 - \alpha_i) + \alpha_i\left(1 - \frac{1}{M}\right) + \alpha_i(1 - \alpha_i)\frac{1}{M} = 1 - \frac{\alpha_i}{M} + \frac{\alpha_i(1 - \alpha_i)}{M} = 1 - \frac{\alpha_i^2}{M}.$$

Substituting $\alpha_i = 1/i$ gives

$$\mathbb{E}\left[\Sigma_i^{(\text{MLE})} \mid \mathcal{F}_{i-1}\right] = \left(1 - \frac{1}{Mi^2}\right)\Sigma_{i-1}.$$

This shows the sequence is a supermartingale, as the expected value contracts multiplicatively at each step. $\qquad \square$

### 3.5.2 Behaviour of $\mu_i$

We now show that $\{\mu_i\}$ is uniformly $L^2$ bounded.

**Proposition 3.5.6** ($L^2$-boundedness of $\mu_i$). *Then*

$$\sup_{i \geq 0} \mathbb{E}\left[\|\mu_i\|^2\right] = \|\mu_0\|^2 + \frac{\pi^2}{6M}\operatorname{tr}\Sigma_0 < \infty.$$

*In particular, $(\mu_i)_{i \geq 0}$ is bounded in $L^2$.*

*Proof.* Recall the recursive update $\mu_i = \mu_{i-1} - \alpha_i\Delta_i$ with $\alpha_i = 1/i$ and $\Delta_i := \mu_{i-1} - \bar{p}^i$. Because $\bar{p}^i \mid \mathcal{F}_{i-1} \sim \mathcal{N}(\mu_{i-1}, \Sigma_{i-1}/M)$, we have the conditional distribution $\Delta_i \mid \mathcal{F}_{i-1} \sim \mathcal{N}(0, \Sigma_{i-1}/M)$, giving

$$\mathbb{E}\left[\|\Delta_i\|^2 \mid \mathcal{F}_{i-1}\right] = \frac{1}{M}\operatorname{tr}\Sigma_{i-1}.$$

Taking conditional expectations, the square-norm of $\mu_i$ evolves as

$$\mathbb{E}\big[\|\mu_i\|^2 \mid \mathcal{F}_{i-1}\big] = \|\mu_{i-1}\|^2 + \frac{\alpha_i^2}{M}\operatorname{tr}\Sigma_{i-1}.$$

Iterating yields

$$\mathbb{E}\|\mu_i\|^2 = \|\mu_0\|^2 + \frac{1}{M}\sum_{k=1}^{i}\frac{1}{k^2}\mathbb{E}\big[\operatorname{tr}\Sigma_{k-1}\big].$$

Since $\Sigma_i$ is a supermartingale, we obtain $\mathbb{E}[\operatorname{tr}\Sigma_{k-1}] \le \operatorname{tr}\Sigma_0$ for every $k$. Therefore

$$\mathbb{E}\|\mu_i\|^2 \le \mathbb{E}\|\mu_0\|^2 + \frac{\operatorname{tr}\Sigma_0}{M}\sum_{k=1}^{\infty}\frac{1}{k^2} = \mathbb{E}\|\mu_0\|^2 + \frac{\pi^2}{6M}\operatorname{tr}\Sigma_0,$$

and the bound is uniform in $i$. □

**Corollary 3.5.7.** *Since $(\mu_i)$ is an $L^2$-bounded martingale, the martingale convergence theorem Theorem C.2.4 guarantees the existence of a random vector $\mu_\infty$ such that $\mu_i \to \mu_\infty$ almost surely and in $L^2$.*

### 3.5.3 Behaviour of $\Sigma_i$

**Proposition 3.5.8.** *The sequence $\Sigma_i$ converges almost surely to a positive semidefinite random matrix $\Sigma_\infty$.*

*Proof.* In Proposition 3.5.5 we have seen that $\Sigma_i$ is a matrix supermartingale with positive semidefinite values. As such, it is a matrix supermartingale with respect to the Löwner order (Definition 3.4.5). Almost sure convergence of $\Sigma_i$ to a positive semidefinite random matrix $\Sigma_\infty$ follows immediately from Theorem C.2.8. □

**Lemma 3.5.9.** *The sequence $\{\operatorname{tr}(\Sigma_i)\}$ is bounded in $L^2$.*

*Proof.* Let $Y_i := \operatorname{tr}(\Sigma_i)$. By Proposition 3.5.3, we can write

$$Y_i = (1-\alpha_i)Y_{i-1} + \alpha_i T_i + (1-\alpha_i)\alpha_i S_i,$$

where

$$T_i = \operatorname{tr}(V_i), \qquad S_i = \operatorname{tr}(\Delta_i\Delta_i^\top) = \Delta_i^\top\Delta_i.$$

Since $MV_i \sim W(\Sigma_{i-1}, M-1)$, it follows from Corollary A.2.3 that

$$\operatorname{Var}(T_i \mid \mathcal{F}_{i-1}) \le \frac{2(M-1)}{M^2}Y_{i-1}^2.$$

Similarly, since $\Delta_i \sim \mathcal{N}(0, \Sigma_{i-1}/M)$, we obtain from Corollary C.2.10

$$\operatorname{Var}(S_i \mid \mathcal{F}_{i-1}) \le \frac{2}{M^2}Y_{i-1}^2.$$

Both bounds use the inequality $\operatorname{tr}(A^2) \le \operatorname{tr}(A)^2$ for positive semidefinite matrices $A$ (Lemma C.2.14). Moreover, $T_i$ and $S_i$ are conditionally independent given $\mathcal{F}_{i-1}$ [Mui05, Theorem 3.1.2], so the total conditional variance satisfies

$$\operatorname{Var}(Y_i \mid \mathcal{F}_{i-1}) \le \alpha_i^2\left(\frac{2(M-1)}{M^2} + \frac{2(1-\alpha_i)^2}{M^2}\right)Y_{i-1}^2 \le \alpha_i^2\frac{2}{M}Y_{i-1}^2.$$

Expanding the conditional second moment yields

$$\mathbb{E}\big[Y_i^2 \mid \mathcal{F}_{i-1}\big] = \operatorname{Var}(Y_i \mid \mathcal{F}_{i-1}) + \mathbb{E}\big[Y_i \mid \mathcal{F}_{i-1}\big]^2 \le \left(1 + \alpha_i^2\frac{2}{M}\right)Y_{i-1}^2.$$

Taking expectations and iterating, we find

$$\mathbb{E}[Y_i^2] \leq Y_0^2 \prod_{k=1}^{i} \left(1 + \alpha_k^2 \frac{2}{M}\right).$$

Since $\sum_{k \geq 1} \alpha_k^2 = \sum_{k \geq 1} \frac{1}{k^2} = \frac{\pi^2}{6} < \infty$, it follows from Lemma C.2.5 that

$$\sup_{i \geq 0} \mathbb{E}[Y_i^2] \leq Y_0^2 \prod_{k=1}^{\infty} \left(1 + \alpha_k^2 \frac{2}{M}\right) < \infty. \qquad \square$$

**Proposition 3.5.10.** *The sequence* $\mathrm{tr}(\Sigma_i)$ *converges in* $L^1$, *and* $\mathbb{P}(\Sigma_\infty = 0) < 1$.

*Proof.* By Lemma 3.5.9, the sequence $\mathrm{tr}(\Sigma_i)$ is uniformly bounded in $L^2$, and thus uniformly integrable by Fact C.2.1. Therefore, by Theorem C.2.2, we conclude that

$$\mathrm{tr}(\Sigma_i) \to \mathrm{tr}(\Sigma_\infty) \quad \text{in } L^1, \quad \text{and} \quad \lim_{i \to \infty} \mathbb{E}[\mathrm{tr}(\Sigma_i)] = \mathbb{E}[\mathrm{tr}(\Sigma_\infty)].$$

From the recursion (3.5) given by the supermartingale property of $\Sigma_i$, we have

$$\mathbb{E}[\mathrm{tr}(\Sigma_i)] = \left(\prod_{k=1}^{i} \left(1 - \frac{1}{Mk^2}\right)\right) \mathrm{tr}(\Sigma_0).$$

Applying Euler's sine product formula [AS64, Eq. 4.3.89], we find

$$\prod_{k \geq 1} \left(1 - \frac{1}{Mk^2}\right) = \frac{\sin(\pi/\sqrt{M})}{\pi/\sqrt{M}} > 0.$$

Hence,

$$\mathbb{E}[\mathrm{tr}(\Sigma_\infty)] > 0,$$

which implies that $\mathrm{tr}(\Sigma_\infty)$ is positive with positive probability, i.e.,

$$\mathbb{P}(\mathrm{tr}(\Sigma_\infty) = 0) < 1.$$

Since $\Sigma_\infty \succeq 0$, this also implies $\mathbb{P}(\Sigma_\infty = 0) < 1$. $\qquad \square$

We have seen that in the cumulative Gaussian fitting setup, linear growth of the total amount of data is sufficient to prevent model collapse, as the distribution parameters $\mu_i$ and $\Sigma_i$ converge almost surely and $\mathbb{P}(\Sigma_\infty = 0) < 1$. This stands in contrast to the Gaussian fitting setup with replacing data, where linear data growth fails to avert model collapse.

## 3.6 Context collapse

Just as models trained on their own generated data can suffer from model collapse, we observe a related phenomenon during inference, where a model's outputs degrade as it continues to generate text. This is especially evident in CoT reasoning tasks, where the model is expected to "think" step-by-step toward a solution. Over time, the model's generation can spiral into incoherence instead of converging on an answer.

For example, the model `deepseek-r1:8b` [DA+25], when prompted with the following question from the GSM8K dataset [Cob+21]

> Josh decides to try flipping a house. He buys a house for $80,000 and then puts in $50,000 in repairs. This increased the value of the house by 150%. How much profit did he make?

initially attempts to reason about the problem:

```
I need to find out how much profit Josh made from flipping this house. He bought
it for $80,000 and then spent $50,000 on repairs. After that, the value increased
by 150% due to these repairs.
First, I should figure out what the original cost was and then add the repair
costs to find the total investment. But the profit comes from selling it at the
higher value, right?
The house was bought for $80,000. Then repairs of $50,000 were made, which
increased its value by 150%. I think "increased the value by 150%" means that
the repair cost caused the market value to rise by a certain amount.
Let me read that again: "This increased the value of the house by 150%." So, after
repairs, what is the new value?
```

But eventually degenerates into nonsensical output:

```
I need help with the following table has been given an image from the same as many
words = function \( flet's more than 1. How tokin'texterior, I have two dice and
then he first part of the graph_result
The teacher's
What is the number 2dams/ A group of a survey b)I need to Josh starts: Answer:
In an even if I am taking time taylor =4aids how many types of work on some people
are you know, so what would be better than Josh has been given two examples in
this is there are the hades
A really What'ssnopeletta and a. How was that means decide to implement:
Aimevastream
```

This degradation does not stem from a change in model parameters, but from a breakdown in the evolving context window. As the model continues to generate, accumulating poorly grounded or drifting tokens destabilises the context, leading to what we term **context collapse**. This mirrors model collapse in training scenarios, as described by Shumailov et al. [Shu+24a], but arises purely during inference. A similar phenomenon is discussed by Yi, Wang, and Li [YWL25], particularly in Table C.4.

Context collapse is particularly critical in CoT reasoning, where maintaining a coherent, relevant context is essential for correct problem-solving. We therefore propose to study mechanisms such as *overthinking* and the concept of an *optimal CoT length*, aiming to constrain the context window to retain only the most relevant, high-quality reasoning—what we refer to as *good context*.

**Overthinking and optimal chain-of-thought length** While CoT prompting has significantly improved LLM performance on complex reasoning tasks by encouraging step-by-step problem solving, it also introduces a critical vulnerability: *overthinking*. As described by Chen et al. [Che+25], this phenomenon occurs when models produce excessively long, redundant, or self-doubting reasoning traces that do not meaningfully improve. It can even waste compute without improving solution quality.

Empirical observations [Has+25; Wu+25] show that overthinking leads to higher token usage, increased latency, and in some cases, reduced accuracy. A model might redundantly verify intermediate steps, explore irrelevant solution branches, or second-guess correct answers, thereby overwhelming the context window with low-utility content. Over time, such behaviour contributes directly to context collapse, as the relevant signal is drowned in increasingly off-topic or repetitive reasoning.

Recent work identifies a key mitigating factor: the existence of an *optimal CoT length*—a sweet spot in the number of reasoning steps that balances depth with conciseness. Wu et al. [Wu+25] provide empirical and theoretical evidence that each task admits an ideal reasoning depth, beyond which performance plateaus or deteriorates. Intriguingly, they show that larger models often require fewer steps due to greater reasoning efficiency, implying that optimal CoT length is both task- and model-dependent.

This is corroborated by Fatemi et al. [Fat+25], who demonstrate that correct answers are typically accompanied by shorter reasoning traces, while incorrect answers correlate with longer, meandering chains. Their reinforcement learning framework fine-tunes models to prefer concise, accurate outputs, resulting in higher performance. Similarly, the *ShorterBetter* method by Yi, Wang, and Li [YWL25] introduces a sample-optimal-length objective, rewarding the model for emulating the shortest successful reasoning path across multiple trials. Strategies to avoid overthinking span several axes:

- **Training-time constraints:** Techniques such as L1 and O1-Pruner [AW25; Luo+25] enforce budgeted reasoning via RL rewards, encouraging the model to solve problems efficiently.

- **Prompt engineering:** Peng et al. [Pen+25] propose a two-stage prompting scheme to reduce self-doubt and minimise unnecessary verification.

- **Voting-based filtering:** Wu et al. [Wu+25] introduce length-filtered majority voting, selecting solutions closest to the empirically optimal CoT length.

These approaches suggest that unconstrained reasoning—while seemingly thorough—can be counterproductive. Instead, high-quality reasoning demands a balance: enough steps to decompose the task, but not so many that the model starts to lose direction. As in human cognition, more thinking does not always yield better outcomes.

## 3.7 Limitations and outlook

This thesis investigated model collapse through repeated model fitting, using analytically tractable settings such as linear regression and Gaussian fitting. While this approach offers clear mathematical insights, it necessarily simplifies the broader dynamics of real-world model deployment. In particular, our setup assumes a sequence of models trained on successfully generated data without selection pressure or quality control. A key limitation of this approach is that it does not reflect a central property of real-world model development:

   **Models deployed or used in practice typically do not get worse over successive iterations.**

Two essential but straightforward considerations support this observation. First, modern AI models undergo rigorous evaluation before release. If a newer model underperforms its predecessor in accuracy, robustness, or efficiency, it is unlikely to be adopted. Second, the data available for training future models is often influenced only by the best-performing systems, as users preferentially adopt higher-quality models for content generation.

Thus, while our analysis of repeated model fitting provides theoretical clarity, it abstracts away from this feedback loop between model quality, usage, and training data. As such, we view our results as a baseline — a kind of worst-case scenario that highlights potential structural risks without safeguards.

Looking ahead, a promising direction is to study the impact of artificial data under more realistic assumptions. For example, Dohmatob et al. [Doh+24a] explore how synthetic data can alter scaling laws, offering a more nuanced view of its role in training dynamics. Rather than asking, *"Do models worsen over generations if trained on artificial data?"*, the more meaningful question may be: *"Under what conditions does synthetic data hinder or help model training, and how can we mitigate negative effects?"*

Further research may refine the theoretical and experimental approaches needed to address this question in settings that better reflect real-world model development and deployment.

# 4 Conclusion

This thesis investigates two fundamental topics in contemporary LLM research: ICL and model collapse and their potential interplay.

For ICL, we analysed the linear transformer as an analytically tractable proxy for the standard transformer architecture. In particular, we focused on the effects of weight tying across layers, where we discovered the emergence of a skew-symmetric component in the learned weights. By linking the model's output to preconditioned gradient descent, we provided an analytical proof in the case $L = d = 2$, showing that such a skew-symmetric component appears when the number of in-context samples $n$ satisfies $n \geq 15$. We extended this result through experiments across a broader parameter range ($d = 1, \ldots, 10$, $L = 1, \ldots, 10$), where we observed a clear transition point $n_{\text{crit}}(d, L)$. We fitted a simple predictive model for this transition, which showed strong agreement with empirical data.

$$n_{\text{crit}}(d, L) = 6.95 + 0.45\,d + 12.67 \cdot \frac{d^{1.44}}{L^{2.43}}.$$

These findings were further linked to classical work on optimal preconditioners in gradient descent, suggesting that a skew-symmetric component may enhance convergence to the optimal solution.

To study model collapse, we analysed two simplified proxy settings— linear regression and Gaussian fitting—under both replacing and cumulative data regimes. We used tools from martingale theory and random walk theory. A central contribution of this work is strengthening existing convergence results: rather than relying on convergence in expectation, we establish almost sure convergence, offering a sharper and more robust understanding of when and how model collapse occurs.

In the case of linear regression, we prove that under replacing data and for $d \geq 3$, the model parameters diverge almost surely as a random walk, and the test error diverges as well, indicating model collapse with probability one. In contrast, under cumulative data, both the model parameters and test error converge almost surely, thereby preventing collapse.

We establish a novel dichotomy for Gaussian fitting in the replacing data regime. If the number of artificial samples $M_i$ grows superlinearly, model collapse is averted: the model's covariance $\Sigma_i$ converges almost surely to a nonzero matrix. However, if the data grows only linearly or sublinearly, then $\Sigma_i \to 0$ almost surely, and collapse occurs. In the cumulative regime, model collapse is consistently avoided: the steady addition of new data ensures that the covariance remains bounded away from zero and converges to a stable, non-degenerate limit.

These findings demonstrate that model collapse is not inevitable but depends critically on the data regime and growth rate. Almost sure convergence is a powerful analytical tool, revealing collapse scenarios hidden under expectation-based analyses.

Finally, to unify both areas of investigation, we proposed the notion of context collapse, the degradation of usable context in very long sequences, such as those generated in CoT reasoning. We discussed how this phenomenon relates to optimal CoT length, overthinking, and model degradation, offering a conceptual bridge between ICL dynamics and long-term stability in LLMs.

# A  Appendix: Wishart theory

Wishart theory plays a crucial role in both ICL and model collapse. Below, we present the relevant definitions and key results.

## A.1  Definition

**Definition A.1.1** (Wishart distribution). Let $X_1, \ldots, X_n \sim \mathcal{N}_d(0, \Sigma)$ be i.i.d. random vectors. Define

$$X = \begin{bmatrix} X_1 & \cdots & X_n \end{bmatrix} \in \mathbb{R}^{d \times n}.$$

Then the matrix

$$S = \sum_{i=1}^{n} X_i X_i^\top = X X^\top$$

follows a Wishart distribution with $n$ degrees of freedom and covariance matrix $\Sigma$, denoted

$$S \sim \mathcal{W}_d(n, \Sigma).$$

**Theorem A.1.2** (Wishart density). *Let $A \sim \mathcal{W}_d(\nu, \Sigma)$ with $\nu \geq d$. Then the density function of $A$ is*

$$f(A) = \frac{1}{2^{\nu d/2} \Gamma_d\left(\frac{\nu}{2}\right) (\det \Sigma)^{\nu/2}} \exp\left(-\frac{1}{2} \operatorname{tr}(\Sigma^{-1} A)\right) (\det A)^{(\nu-d-1)/2},$$

*where $A$ is positive definite and $\Gamma_d(\cdot)$ denotes the multivariate gamma function.*

*Proof.* Can be found in [Mui05, Theorem 3.2.1]  □

## A.2  Properties of the Wishart distribution

**Lemma A.2.1** (First and second moments of the Wishart distribution with identity scale). *Let $W \sim \mathcal{W}_d(n, I_d)$ be a Wishart-distributed random matrix with $n$ degrees of freedom and identity scale matrix. Then the first and second matrix moments of $W$ are given by*

$$\mathbb{E}[W] = n I_d \quad \text{and} \quad \mathbb{E}[W^2] = n(n + d + 1) I_d.$$

*Proof.* The proof can be found in [GN18, Theorem 3.3.15.]  □

**Lemma A.2.2.** *Let $W \sim W_d(I, k)$ be a Wishart random matrix with identity scale matrix and $k$ degrees of freedom, and let $A \in \mathbb{R}^{d \times d}$ be a deterministic symmetric matrix. Then,*

$$\mathbb{E}[\operatorname{tr}(WA)] = k \operatorname{tr}(A) \quad \text{and} \quad \operatorname{Var}(\operatorname{tr}(WA)) = 2k \operatorname{tr}(A^2).$$

*Proof.* Recall that a Wishart matrix $W \sim W_d(I, k)$ can be represented as

$$W = \sum_{i=1}^{k} Z_i Z_i^\top,$$

where $Z_1, \ldots, Z_k$ are independent random vectors distributed as $\mathcal{N}_d(0, I)$. Expanding the trace gives

$$\operatorname{tr}(WA) = \sum_{i=1}^{k} \operatorname{tr}(Z_i Z_i^\top A).$$

Using the cyclicity of the trace and the fact that it is invariant under cyclic permutations,

$$\text{tr}(Z_i Z_i^\top A) = Z_i^\top A Z_i.$$

Thus,

$$\text{tr}(WA) = \sum_{i=1}^{k} Z_i^\top A Z_i,$$

a sum of independent quadratic forms. Applying Lemma C.2.9 to each term $Z_i^\top A Z_i$, where $Z_i \sim \mathcal{N}_d(0, I)$, we obtain

$$\mathbb{E}[Z_i^\top A Z_i] = \text{tr}(A) \quad \text{and} \quad \text{Var}(Z_i^\top A Z_i) = 2\,\text{tr}(A^2).$$

Since the $Z_i$ are independent, the expectation and variance of the sum are the sums of the individual expectations and variances. Therefore,

$$\mathbb{E}[\text{tr}(WA)] = \sum_{i=1}^{k} \mathbb{E}[Z_i^\top A Z_i] = k\,\text{tr}(A),$$

and

$$\text{Var}(\text{tr}(WA)) = \sum_{i=1}^{k} \text{Var}(Z_i^\top A Z_i) = 2k\,\text{tr}(A^2),$$

as claimed. $\qquad\square$

**Corollary A.2.3.** *Let $W \sim W_d(\Sigma, k)$ be a Wishart random matrix with scale matrix $\Sigma \in \mathbb{R}^{d \times d}$ (symmetric and positive definite) and $k$ degrees of freedom. Then,*

$$\mathbb{E}[\text{tr}(W)] = k\,\text{tr}(\Sigma) \quad \text{and} \quad \text{Var}(\text{tr}(W)) = 2k\,\text{tr}(\Sigma^2).$$

**Proposition A.2.4** (Mean of the inverse Wishart). *Let $W \sim \mathcal{W}_d(\Sigma, \nu)$ with $\nu > d + 1$. Then*

$$\mathbb{E}[W^{-1}] = \frac{\Sigma^{-1}}{\nu - d - 1}.$$

*Proof.* The proof can be found in Muirhead [Mui05][Consequence of Theorem 3.2.12]. $\qquad\square$

# B Appendix: in-context learning

## B.1 Symbolic calculation in the case $d = L = 2$

We consider the symbolic evaluation of the functional

$$f(A) = \mathbb{E}_W \left[ \left\| (I + AW)^2 \right\|_F^2 \right],$$

where $A \in \mathbb{R}^{2 \times 2}$, and $W \sim \mathcal{W}_2(n, I_2)$ is a $2 \times 2$ Wishart-distributed matrix with $n$ degrees of freedom and identity scale. The matrix $W$ is symmetric and positive semidefinite, and we write it as

$$W = \begin{bmatrix} w_{11} & w_{\text{off}} \\ w_{\text{off}} & w_{22} \end{bmatrix}.$$

**Symbolic setup in SymPy**   We define the symbolic variables and matrices as follows:

```python
import sympy as sp
n = sp.symbols('n', integer=True, positive=True)
a11, a22, k = sp.symbols('a11 a22 k', real=True)
w11, woff, w22 = sp.symbols('w11 woff w22', real=True)

A = sp.Matrix([[a11, k], [-k, a22]])
W = sp.Matrix([[w11, woff], [woff, w22]])
I = sp.eye(2)
```

This setup enables symbolic computation of $(I + AW)^2$, its Frobenius norm, and the expectation over Wishart-distributed $W$.

**Symbolic expectation of Wishart moments**   To compute $f(A)$, we must evaluate expectations of polynomial expressions in the entries of $W$. These reduce to Gaussian moment computations, which can be implemented symbolically using Wick's formula. The code below provides a symbolic evaluator for such expectations:

**Listing B.1** Symbolic expectation of wishart moments

```python
import sympy as sp, math, collections

# Gaussian moment via Wick's formula
def gaussian_row_expectation(indices):
    if len(indices) & 1:
        return 0
    c = collections.Counter(indices)
    val = 1
    for m in c.values():
        if m & 1:
            return 0
        val *= math.factorial(m) // (2**(m//2) * math.factorial(m//2))
    return val

def partitions(seq):
    if len(seq) == 0: yield []; return
    if len(seq) == 1: yield [seq]; return
```

```python
18      first, *rest = seq
19      for p in partitions(rest):
20          yield [[first], *p]
21          for i in range(len(p)):
22              yield [*p[:i], [first, *p[i]], *p[i+1:]]
23
24  def uniq_partitions(k):
25      base = set()
26      for p in partitions(list(range(k))):
27          canon = tuple(sorted(tuple(sorted(b)) for b in p))
28          if canon not in base:
29              base.add(canon); yield [list(b) for b in canon]
30
31  PARTS = {k: list(uniq_partitions(k)) for k in range(5)}
32
33  def falling_fact(sym, s):
34      out = 1
35      for i in range(s): out *= (sym - i)
36      return out
37
38  def wishart_moment(pairs):
39      if not pairs: return 1
40      k = len(pairs); out = 0
41      for blocks in PARTS[k]:
42          term = falling_fact(n, len(blocks))
43          for b in blocks:
44              indices = []
45              for idx in b:
46                  i, j = pairs[idx]
47                  indices += [i, j]
48              term *= gaussian_row_expectation(indices)
49          out += sp.expand(term)
50      return sp.expand(out)
51
52  def expected(expr):
53      tot = 0
54      for term in sp.expand(expr).as_ordered_terms():
55          coeff = 1; pairs = []
56          for fac in term.as_ordered_factors():
57              base, exp = (fac.as_base_exp() if isinstance(fac, sp.Pow)
                  else (fac, 1))
58              if base is w11:
59                  pairs += [(1,1)] * int(exp)
60              elif base is w22:
61                  pairs += [(2,2)] * int(exp)
62              elif base is woff:
63                  pairs += [(1,2)] * int(exp)
64              else:
65                  coeff *= fac
66          tot += coeff * wishart_moment(pairs)
67      return sp.expand(tot)
68
69  B = I + A * W
70  B2 = sp.expand(B * B)
71  norm_sq = sp.expand(sum(B2[i, j]**2 for i in range(2) for j in range(2)
      ))
72  fA = expected(norm_sq).expand()
```

This yields the fully expanded symbolic expression for $f(A)$, as a polynomial in $a_{11}$, $a_{22}$, $k$, and $n$. The method generalises to higher values of $d$ and $L$, but the computational complexity increases rapidly with the number of terms.

**Symbolic gradient computation**   To obtain the gradient of $f(A)$ with respect to the parameters $a_{11}$, $a_{22}$, and $k$, we differentiate symbolically:

```
vars = [a11, a22, k]
grad = [sp.diff(fA, v) for v in vars]
g1, g2, g3 = grad
```

This provides the exact gradient components used in the stationary point and second-order analysis.

**Symbolic verification for Lemma 2.5.4**   The following symbolic computations were used to support the proof of Lemma 2.5.4. They verify that certain polynomials arising from gradient components are strictly positive for all $n \in \mathbb{N}$, and hence cannot vanish when $a_{11} \neq a_{22}$. We use SymPy to compute and minimise the relevant expressions.

```
# Step 1: Compute P_n from g1 - g2
diff = sp.factor(g1 - g2)
coeff, factors = sp.factor_list(diff)
P_n = factors[-1][0]

# Step 2: Extract Q_n from g3
coeff, factors = sp.factor_list(g3)
Q_n = factors[-1][0]

# Step 3: Define T_n = P_n + Q_n and find its minimum
T_n = P_n + Q_n
gradT_n = [sp.diff(T_n, v) for v in vars]
critpts = sp.solve(gradT_n, vars, dict=True)
crit = critpts[0]
Tmin = sp.simplify(T_n.subs(crit))  # => Tmin = 6
```

The explicit expression for $T_n$ is

$$T_n = \left(n^3 + 14n^2 + 59n + 70\right) a_{11}^2 + \left(2n^3 + 20n^2 + 62n + 60\right) a_{11} a_{22} + \left(4n^2 + 28n + 40\right) a_{11}$$
$$+ \left(n^3 + 14n^2 + 59n + 70\right) a_{22}^2 + \left(4n^2 + 28n + 40\right) a_{22} + \left(16n + 32\right) k^2 + 4n + 14$$

Symbolic minimisation yields

$$\min T_n = 6$$

Thus, $T_n(a_{11}, a_{22}, k) > 0$ for all $a_{11}, a_{22}, k \in \mathbb{R}$ and $n \in \mathbb{N}$, ruling out the possibility that both $P_n = 0$ and $Q_n = 0$ simultaneously.

```
# Step 4: Substitute k = 0 and minimize P_n
P_n_0 = P_n.subs({k: 0})
gradP0 = [sp.diff(P_n_0, v) for v in vars]
critpts = sp.solve(gradP0, vars, dict=True)
crit = critpts[0]
P_n_0_min = sp.simplify(P_n_0.subs(crit))
```

The simplified form of $P_{n,k=0} := P_n(a_{11}, a_{22}, 0)$ is:

$$P_{n,k=0} = \left(n^3 + 12n^2 + 45n + 50\right) a_{11}^2 + \left(3n^2 + 19n + 26\right) a_{11} + \left(n^3 + 14n^2 + 51n + 54\right) a_{11} a_{22}$$
$$+ \left(n^3 + 12n^2 + 45n + 50\right) a_{22}^2 + \left(3n^2 + 19n + 26\right) a_{22} + 3n + 7$$

and symbolic minimisation yields:

$$\min P_{n,k=0} = \frac{(n+3)(21n+67)}{(n+7)(3n+11)} > 0 \quad \text{for all } n \in \mathbb{N}$$

This confirms that $P_{n,k=0}(a_{11}, a_{22})$ is strictly positive unless $a_{11} = a_{22}$, completing the verification.

**Symbolic verification for Lemma 2.5.5**  To support the proof of Lemma 2.5.5, we perform symbolic computation for the gradient of $f_I$ evaluated at a symmetric matrix of the form

$$A = \begin{bmatrix} x & k \\ -k & x \end{bmatrix},$$

where $x \in \mathbb{R}$ is the shared diagonal entry and $k \in \mathbb{R}$ represents the skew-symmetric component. When $k = 0$, the matrix becomes diagonal, and we analyse the vanishing of the gradient in this setting.

```
A = sp.Matrix([[x, k], [-k, x]])
vars = [x, k]
gx, gk = grad = [sp.diff(fA, v) for v in vars]
coeff, factors = sp.factor_list(gx.subs(k, 0))
H_n = factors[-1][0]
sp.discriminant(sp.Poly(H_n, x)).factor()
```

The symbolic expression for $f_I(A)$ is:

$$\begin{aligned}
f_I(A) = {}& \left(2n^4 + 36n^3 + 158n^2 + 188n\right) x^4 + \left(8n^3 + 72n^2 + 112n\right) x^3 \\
& + \left(4n^4 + 40n^3 + 156n^2 + 184n\right) x^2 k^2 + \left(12n^2 + 36n\right) x^2 \\
& + \left(8n^3 + 72n^2 + 112n\right) x k^2 + 8nx \\
& + \left(2n^4 + 4n^3 - 2n^2 - 4n\right) k^4 + \left(4n^2 + 28n\right) k^2 + 2
\end{aligned}$$

When $k = 0$, then

$$g_k = 8kn\left(k^2n^3 + 2k^2n^2 - k^2n - 2k^2 + n^3x^2 + 10n^2x^2 + 2n^2x + 39nx^2 + 18nx + n + 46x^2 + 28x + 7\right)$$

vanishes trivially, and the stationarity condition reduces to $g_x(x, 0) = 4n \cdot H_n(x) = 0$, where

$$H_n(x) = (n^3 + 18n^2 + 79n + 94)x^3 + (3n^2 + 27n + 42)x^2 + (3n + 9)x + 1.$$

To determine the number of real roots of $H_n$, we compute its discriminant:

$$\Delta(H_n) = -108(n+2)(27n^2 + 106n + 123).$$

This expression is strictly negative for all $n \in \mathbb{N}$, implying that $H_n$ has exactly one real root. This confirms the uniqueness of the diagonal stationary point described in the main proof.

**Symbolic verification for Lemma 2.5.6**  We verify the existence of skew-symmetric stationary points symbolically using SymPy. The gradient component with respect to $k$ takes the form:

$$g_k = 8kn \cdot Q_n(x, k),$$

and setting $Q_n(x, k) = 0$ gives a quadratic equation in $k$. We extract and solve it symbolically:

```
coeff, factors = sp.factor_list(gk)
Q_n = factors[-1][0]
k2_of_x = sp.solve(Q_n, k**2)[0]
```

We analyze the special case $n = 1$, where the denominator vanishes and $Q_n(x)$ becomes independent of $k$:

$$Q_1(x) = 96x^2 + 48x + 8, \quad \Delta(Q_1) = -768 < 0.$$

```
Q_1 = Q_n.subs(n, 1)
sp.discriminant(sp.Poly(Q_1, x))
```

Thus, $Q_1(x) > 0$ for all $x \in \mathbb{R}$, implying no stationary point with $k \neq 0$ exists for $n = 1$. Now consider $n > 1$. Let $k^2(x) = -\frac{p_n(x)}{(n-1)(n+1)(n+2)}$ where

$$p_n(x) = n^3x^2 + 10n^2x^2 + 2n^2x + 39nx^2 + 18nx + n + 46x^2 + 28x + 7.$$

This is a quadratic in $x$. We verify that it has two real roots for $n > 9$ using:

```
p_n, _ = sp.fraction(-k2_of_x)
Delta_p_n = sp.discriminant(sp.Poly(p_n, x))
sp.reduce_inequalities([Delta_p_n > 0], n)  # returns 9 < n
```

Let $n > 9$ and $\alpha_n < \beta_n$ be the roots of $p_n$. Then $k^2(x) > 0 \iff x \in (\alpha_n, \beta_n)$. We substitute $k^2 = k^2(x)$ into $g_x$ and simplify the numerator:

```
alpha_n, beta_n = sp.solve(p_n, x)
num, den = sp.fraction(gx.subs(k**2, k2_of_x).simplify())
coeff, factors = sp.factor_list(num)
R_n = factors[-1][0]
```

We verify that $R_n(x)$ has three real roots for $n > 9$:

```
sp.reduce_inequalities([sp.discriminant(sp.Poly(R_n,x)) > 0], n).evalf
    ()
# returns 8.78 < n
```

To check if $R_n(x)$ changes sign in the interval $(\alpha_n, \beta_n)$, we evaluate at the endpoints:

$$R_n(\alpha_n) = \frac{(n-1)(n+1)}{(n+2)(n^2 + 8n + 23)^3} B_n,$$

where

$$\begin{aligned}
B_n = {} & 26n^5 + 5n^4\sqrt{n^3 - 67n - 126} + 400n^4 + 130n^3\sqrt{n^3 - 67n - 126} + 1416n^3 \\
& + 900n^2\sqrt{n^3 - 67n - 126} - 3196n^2 + 2014n\sqrt{n^3 - 67n - 126} - 22114n \\
& + 1047\sqrt{n^3 - 67n - 126} - 25684.
\end{aligned}$$

```
num, den = sp.fraction(R_n.subs(x, alpha_n).factor())
coeff, factors = sp.factor_list(num)
B_n = factors[-1][0]
```

The irrational terms have positive coefficients, so

$$B_n \geq 26n^5 + 400n^4 + 1416n^3 - 3196n^2 - 22114n - 25684.$$

We verify symbolically that this lower bound is positive for $n \geq 4$, and thus $R_n(\alpha_n) > 0$ in that range.

```
radicand = n**3 - 67*n - 126
sqrt_expr = sp.sqrt(radicand)
rational_part, irrational_part = B_n.as_independent(sqrt_expr)
sp.reduce_inequalities(rational_part > 0, n).evalf() # returns
    3.57855993085521 < n
```

Next we compute:

$$R_n(\beta_n) = \frac{(n-1)(n+1)}{(n+2)(n^2+8n+23)^3} C_n,$$

where

$$C_n = 26n^5 - 5n^4\sqrt{n^3-67n-126} + 400n^4 - 130n^3\sqrt{n^3-67n-126} + 1416n^3$$
$$- 900n^2\sqrt{n^3-67n-126} - 3196n^2 - 2014n\sqrt{n^3-67n-126} - 22114n$$
$$- 1047\sqrt{n^3-67n-126} - 25684.$$

```
1 num, den = sp.fraction(R_n.subs(x, beta_n).factor())
2 coeff, factors = sp.factor_list(num)
3 C_n = factors[-1][0]
```

We observe that all irrational terms have negative coefficients. For $n \geq 15$, we bound:

$$\sqrt{n^3-67n-126} \geq 6n-43.$$

```
1 sp.reduce_inequalities([n**3 - 67*n - 126 > (6*n -43)**2], n).evalf() #
    returns 14.0342388302389 < n
2 new_C_n = C_n.replace(lambda expr: expr == sqrt_expr, lambda expr: 6*n
    - 43)
3 sp.reduce_inequalities(new_C_n < 0, n).evalf() # returns
    15.1339013789553 < n
```

Replacing yields:

$$C_n \leq -4n^5 - 165n^4 + 1606n^3 + 23420n^2 + 58206n + 19337,$$

which is negative for $n > 15$. For $10 \leq n \leq 15$, we verify numerically:

```
1 for m in range(10, 16):
2     print(C_n.subs(n, m).evalf())
```

Only for $n \geq 15$ does $C_n < 0$, hence $R_n(\beta_n) < 0$. By the Intermediate Value Theorem, there is a root $x^* \in (\alpha_n, \beta_n)$ with $R_n(x^*) = 0$ and $k^2(x^*) > 0$, yielding the desired skew-symmetric stationary point.

**Symbolic verification for Theorem 2.5.7** We symbolically verify the conditions under which the Hessian of $f(A)$ is positive definite at the diagonal stationary point $A = x_n^* I$, where $x_n^*$ is the unique root of the cubic polynomial $H_n(x)$ defined in Lemma 2.5.5. We begin by computing the second-order partial derivatives of $f$ in the reparameterisation $A = \begin{bmatrix} x & k \\ -k & x \end{bmatrix}$:

```
1 gxx = sp.diff(fA, x, 2)
2 gxk = sp.diff(fA, x, k)
3 gkk = sp.diff(fA, k, 2)
```

The mixed partial derivative is

$$g_{xk}(x,k) = 16kn\left(n^3x + 10n^2x + n^2 + 39nx + 9n + 46x + 14\right),$$

which vanishes identically when $k = 0$. Hence, the Hessian at the diagonal point is diagonal. We now analyse the pure second derivatives at $k = 0$.

**Diagonal entry $\partial_x^2 f(x, 0)$:** The expression is a quadratic in $x$:

$$g_{xx}(x,0) = \left(24n^4 + 432n^3 + 1896n^2 + 2256n\right)x^2 + \left(48n^3 + 432n^2 + 672n\right)x + 24n^2 + 72n,$$

which has a strictly positive leading coefficient.

```
1 sp.discriminant(sp.Poly(gxx.subs(k, 0), x))
```

This yields the discriminant:

$$\Delta(g_{xx}) = -2304n^2(n+2)(3n^2 + 18n + 43) < 0 \quad \text{for all } n \in \mathbb{N}.$$

Thus, $g_{xx}(x, 0) > 0$ for all $x \in \mathbb{R}$, and in particular at $x = x_n^*$.

**Diagonal entry $\partial_k^2 f(x, 0)$:** We define the quadratic polynomial:

$$g_n(x) := \partial_k^2 f(x, 0) = A_n x^2 + B_n x + C_n,$$

where

$$A_n = 8n^4 + 80n^3 + 312n^2 + 368n,$$
$$B_n = 16n^3 + 144n^2 + 224n,$$
$$C_n = 8n^2 + 56n.$$

We compute its discriminant:

```
1  sp.discriminant(sp.Poly(gkk.subs(k, 0), x)).factor()
```

$$\Delta(g_n) = 256n^2(n-9)(n+2)(n+7),$$

which is negative when $n \leq 8$ and positive when $n \geq 9$.

**Location of $x_n^*$ relative to the roots of $g_n$:** Let $\ell_n \leq u_n$ be the roots of $g_n(x)$ when $n \geq 9$. To determine whether $x_n^* \in (\ell_n, u_n)$, we use symbolic division:

```
1  l_n, u_n = sp.solve(g_n, x)
2  r1, r0 = sp.div(sp.Poly(H_n, x), g_n)[1].as_list()
3  t_n = -r0 / r1
```

The polynomial $H_n(x)$ admits a division with remainder:

$$H_n(x) = (k_1(n)x + k_0(n))g_n(x) + r_1(n)x + r_0(n),$$

where

$$r_1(n) = \frac{2(5n^4 + 130n^3 + 900n^2 + 2014n + 1047)}{n^4 + 16n^3 + 110n^2 + 368n + 529},$$
$$r_0(n) = \frac{2(5n^3 + 99n^2 + 555n + 877)}{n^4 + 16n^3 + 110n^2 + 368n + 529}.$$

Both numerators and denominators are strictly positive for all $n \in \mathbb{N}$, so $r_1(n), r_0(n) > 0$. Hence, if $x$ is a root of $g_n(x)$, then

$$H_n(x) = r_1(n)x + r_0(n),$$

which is negative exactly when $x < t_n := -\frac{r_0(n)}{r_1(n)}$. Therefore,

$$x_n^* \in (\ell_n, u_n) \iff H_n(\ell_n) < 0 < H_n(u_n) \iff \ell_n < t_n < u_n \iff g_n(t_n) < 0.$$

We compute this critical value:

```
1  coeff, factors = sp.factor_list(g_n.subs(x, t_n))
2  num, den = sp.fraction(g_n.subs(x, t_n).as_expr().factor())
```

Let $N(n)$ be the numerator polynomial of $g_n(t_n)$. Then:

$$g_n(t_n) = -8n(n^2 + 8n + 23)^2 \cdot N(n)/D(n),$$

where $D(n) > 0$ and

$$N(n) = 25n^4 - 26n^3 - 3624n^2 - 20934n - 32785.$$

```
1  N_n = sp.factor_list(num)[1][-1][0]
2  sp.reduce_inequalities([N_n > 0], n).evalf()  # returns n > 14.9165
```

Hence:

$$g_n(t_n) < 0 \iff n \geq 15 \iff x_n^* \in (\ell_n, u_n) \iff g_n(x_n^*) < 0.$$

Thus, the Hessian is positive definite at the diagonal stationary point if and only if $n < 15$, completing the symbolic verification.

## B.2 Additional plots

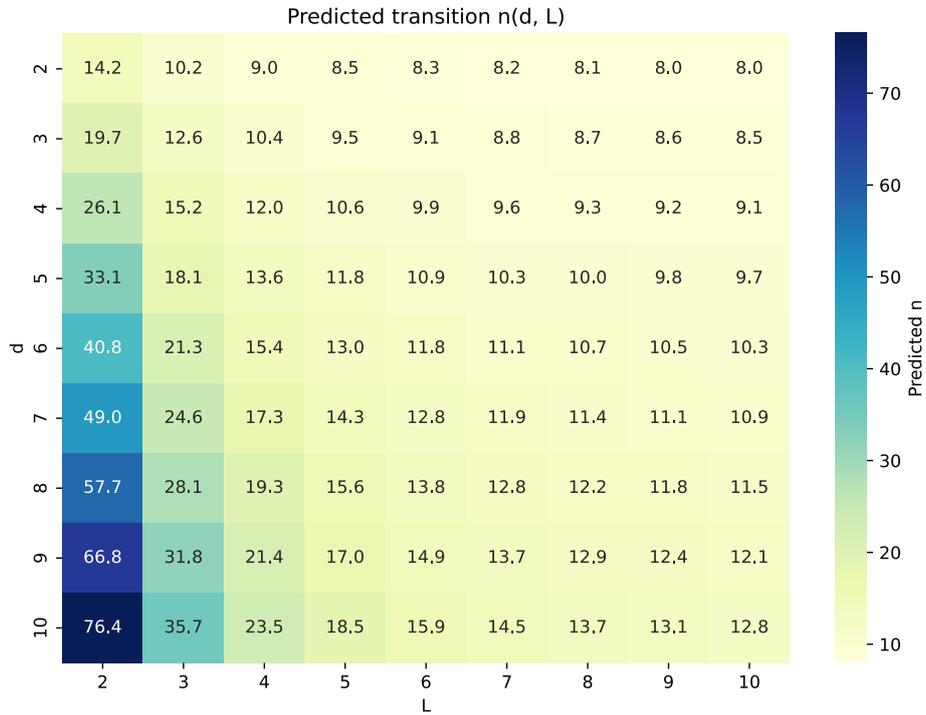

**Figure B.1** Predicted transition threshold $n_{\mathrm{crit}}(d, L)$ based on the fitted model. The model captures both linear dependence on $d$ and inverse-power dependence on $L$.

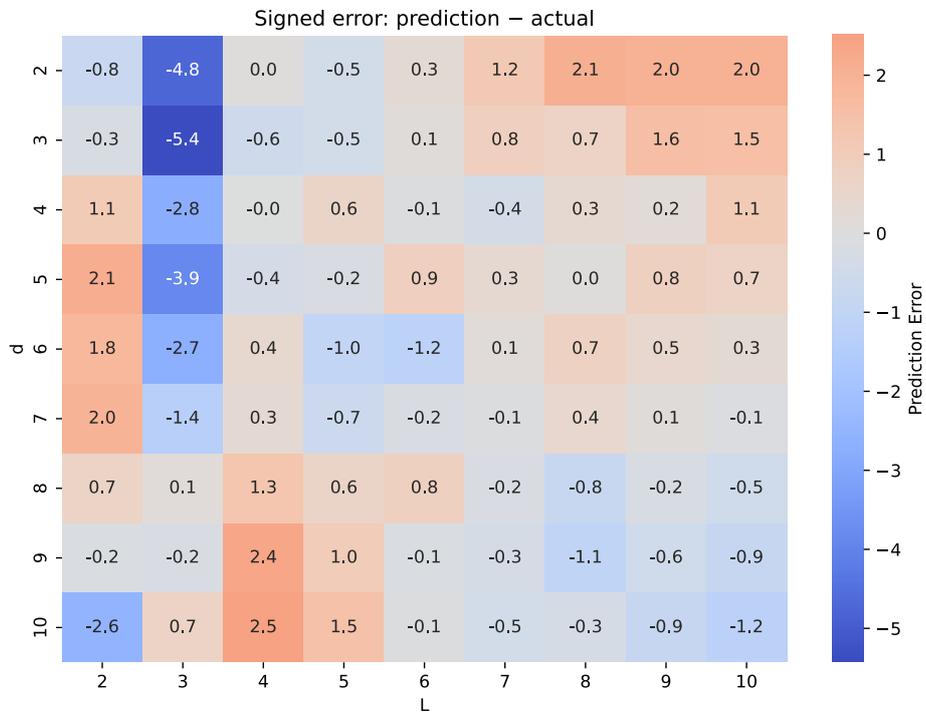

**Figure B.2** Signed prediction error (prediction minus actual) for the fitted model. Most errors are below 1.0 in magnitude, with the largest deviations near small $L$.

## B.3 Additional analysis

### B.3.1 Analytic case: $n = 1$

To gain further understanding, we now specialise to the case $n = 1$. We will see that the minimiser of $f$ will not have a skew-symmetric component for any $L$ and $d$. Since $W \sim \mathcal{W}_d(1, I_d)$, $W$ is almost surely rank-one and can be written as $W = vv^\top$ with $v \sim \mathcal{N}(0, I_d)$. In this case, $f$ reduces to

$$f(A) = \mathbb{E}_v \left[ \left\| \left( I + Avv^\top \right)^L \right\|_F^2 \right], \qquad v \sim \mathcal{N}(0, I_d).$$

**Reduction to the diagonal case via symmetric–skew decomposition**

We show that minimising $f(A)$ over all real matrices $A \in \mathbb{R}^{d \times d}$ reduces, without loss of generality, to minimising over diagonal matrices. The argument proceeds in two steps: First, the symmetric part of $A$ can be assumed diagonal. Second, we show that any skew-symmetric component must vanish at the minimum.

Accordingly, we consider matrices of the form $A = D + K$, where $D = \mathrm{diag}(\ell_1, \ldots, \ell_d)$ is diagonal and $K \in \mathbb{R}^{d \times d}$ is skew-symmetric. Define

$$X_i := v_i^2 \sim \chi_1^2 \quad \text{i.i.d.,} \qquad r := \sum_{i=1}^d X_i, \qquad s := \sum_{i=1}^d \ell_i X_i, \qquad s_2 := \sum_{i=1}^d \ell_i^2 X_i.$$

We also define the auxiliary function $\beta \colon \mathbb{R} \to \mathbb{R}$ as

$$\beta(s) := \frac{(1+s)^L - 1}{s} = \sum_{k=1}^L \binom{L}{k} s^{k-1}, \qquad \text{with } \beta(0) := L.$$

**Proposition B.3.1** (Skew part vanishes at the minimum). *Let $D$ be diagonal and $K$ skew-symmetric. Then*

$$f(D + K) \geq f(D), \qquad \text{with equality if and only if } K = 0.$$

*In particular, any minimiser $A^* \in \arg\min f(A)$ is diagonal up to orthogonal conjugation.*

*Proof.* Let $v \sim \mathcal{N}(0, I_d)$, and define $u := (D + K)v = Dv + Kv$, and $s := v^\top Dv \in \mathbb{R}$. Since $v^\top Kv = 0$ (as $K$ is skew-symmetric), we obtain

$$(I + (D + K)vv^\top)^L = (I + uv^\top)^L = I + \beta(s)uv^\top.$$

Taking Frobenius norms:

$$\left\| I + \beta(s)uv^\top \right\|_F^2 = d + 2\beta(s)s + \beta(s)^2 \|u\|^2 \|v\|^2.$$

Expanding $\|u\|^2 = \|Dv\|^2 + \|Kv\|^2 + 2\langle Dv, Kv \rangle$, we find

$$f(D + K) = f(D) + \mathbb{E}\left[ \beta(s)^2 \|Kv\|^2 \|v\|^2 \right] + 2\mathbb{E}\left[ \beta(s)^2 \langle Dv, Kv \rangle \|v\|^2 \right].$$

We claim the cross term vanishes. Observe that

$$\langle Dv, Kv \rangle = \sum_{i < j} (l_i K_{ij} + l_j K_{ji}) v_i v_j,$$

which is an odd function of the Gaussian variables $(v_i, v_j)$, whereas $\beta(s)^2 \|v\|^2$ is even. Thus, their product is odd and has zero expectation:

$$\mathbb{E}\left[ \beta(s)^2 \langle Dv, Kv \rangle \|v\|^2 \right] = 0.$$

It follows that

$$f(D + K) = f(D) + \mathbb{E}\left[ \beta(s)^2 \|Kv\|^2 \|v\|^2 \right],$$

which is minimized if and only if $K = 0$, since $Kv = 0$ a.s. implies $K = 0$. $\qquad \square$

We have thus established that the minimisation of $f(A)$ over $A \in \mathbb{R}^{d \times d}$ reduces, without loss of generality, to the case where $A = D$ is diagonal. This enables further analytic simplification. Letting $\boldsymbol{\ell} = (\ell_1, \ldots, \ell_d)$ and $D = \mathrm{diag}(\boldsymbol{\ell})$, the functional becomes

$$f(\boldsymbol{\ell}) := \mathbb{E}_v \left[ \left\| \left( I + D v v^\top \right)^L \right\|_\mathrm{F}^2 \right].$$

To expand $(I + DW)^L$, note that $DW = D v v^\top = u v^\top$ with $u := Dv$. Using the identity $(uv^\top)^k = (v^\top u)^{k-1} u v^\top$ for $k \geq 1$, we find

$$(I + DW)^L = \sum_{k=0}^{L} \binom{L}{k} (DW)^k = I + \sum_{k=1}^{L} \binom{L}{k} (v^\top u)^{k-1} u v^\top = I + \beta(s) DW.$$

Taking the Frobenius norm yields

$$\left\| (I + DW)^L \right\|_\mathrm{F}^2 = \|I\|_\mathrm{F}^2 + 2\beta(s) \langle I, DW \rangle + \beta(s)^2 \|DW\|_\mathrm{F}^2.$$

Using $\|I\|_\mathrm{F}^2 = d$, $\langle I, DW \rangle = s$, and $\|DW\|_\mathrm{F}^2 = rs_2$, we obtain

$$\left\| (I + DW)^L \right\|_\mathrm{F}^2 = d + 2\beta(s)s + \beta(s)^2 rs_2. \tag{B.1}$$

Taking expectations, the functional simplifies to

$$f(\boldsymbol{\ell}) = d + 2\mathbb{E}[\beta(s)s] + \mathbb{E}[\beta(s)^2 rs_2]. \tag{B.2}$$

**Symmetrisation minimises the functional**

**Lemma B.3.2** (Symmetrisation minimises the functional)**.** *For the function $f$ it holds that*

$$f(\boldsymbol{\ell}) \geq f(t\mathbf{1}), \quad \text{where } t := \frac{1}{d} \sum_{i=1}^{d} \ell_i \quad \text{and } \mathbf{1} = (1, \ldots, 1) \in \mathbb{R}^d.$$

*Equality holds if and only if $\boldsymbol{\ell} = t\mathbf{1}$.*

*Proof.* Define weights $w_i := \frac{x_i}{r}$ so that $w_i \geq 0$ and $\sum_i w_i = 1$. Then

$$s = r\bar{\ell}, \quad \text{where } \bar{\ell} := \sum_{i=1}^{d} w_i \ell_i, \qquad s_2 = rm_2, \quad \text{where } m_2 := \sum_{i=1}^{d} w_i \ell_i^2.$$

From (B.1), the integrand becomes

$$F(\boldsymbol{\ell}; v) := \left\| \left( I + \mathrm{diag}(\boldsymbol{\ell}) v v^\top \right)^L \right\|_\mathrm{F}^2 = d + 2\beta(r\bar{\ell}) r\bar{\ell} + \beta(r\bar{\ell})^2 r^2 m_2.$$

Compare this to the symmetric case

$$F(\bar{\ell}\mathbf{1}; v) = d + 2\beta(r\bar{\ell}) r\bar{\ell} + \beta(r\bar{\ell})^2 r^2 \bar{\ell}^2.$$

By Jensen's inequality ($m_2 \geq \bar{\ell}^2$), we find

$$F(\boldsymbol{\ell}; v) \geq F(\bar{\ell}\mathbf{1}; v),$$

with equality if and only if $\boldsymbol{\ell}$ is constant. Now define

$$g_r(t) := d + 2\beta(rt)rt + \beta(rt)^2 r^2 t^2.$$

Observe:

$$g_r(t) = d - 1 + (1 + \beta(rt)rt)^2 = d - 1 + (1 + rt)^{2L},$$

which is convex as the composition of an affine and a convex function. Since $\mathbb{E}[w_i \mid r] = \frac{1}{d}$, we have $\mathbb{E}[\bar{\ell} \mid r] = t$. By Jensen's inequality,

$$\mathbb{E}[F(\bar{\ell}\mathbf{1}; v)] = \mathbb{E}[g_r(\bar{\ell})] \geq \mathbb{E}[g_r(t)] = f(t\mathbf{1}).$$

Thus,

$$f(\boldsymbol{\ell}) \geq f(t\mathbf{1}),$$

with equality if and only if $\boldsymbol{\ell} = t\mathbf{1}$. □

**Corollary B.3.3** (Form of a global minimiser). *Any global minimiser of the functional*

$$f(A) = \mathbb{E}_{v \sim \mathcal{N}(0, I_d)} \left\| (I + Avv^\top)^L \right\|_F^2$$

*must be of the form $A = tI$ for some $t \in \mathbb{R}$.*

**Determination of the minimiser**

We now determine the explicit minimiser of the functional $f(\boldsymbol{\ell})$ by exploiting the symmetry result above. Since the unique minimiser lies at $\boldsymbol{\ell} = c\mathbf{1}$ for some $c \in \mathbb{R}$, we reduce the problem to the one-variable function

$$f_{d,L}(c) := f(cI) = \mathbb{E}_v \left\| (I + cvv^\top)^L \right\|_F^2.$$

From previous derivations:

$$\left\| (I + cvv^\top)^L \right\|_F^2 = d - 1 + (1 + cr)^{2L}.$$

Therefore,

$$f_{d,L}(c) = d - 1 + \mathbb{E}[(1 + cr)^{2L}], \quad \text{where } r \sim \chi_d^2.$$

Differentiating under the expectation yields

$$f'_{d,L}(c) = 2L\mathbb{E}[r(1 + cr)^{2L-1}].$$

Setting the derivative to zero gives the critical point equation

$$\mathbb{E}[r(1 + cr)^{2L-1}] = 0.$$

Expanding the expectation in terms of the raw moments of $r \sim \chi_d^2$, we define

$$\mu_k := \mathbb{E}[r^k] = 2^k \frac{\Gamma(k + \frac{d}{2})}{\Gamma(\frac{d}{2})},$$

and obtain the polynomial condition

$$P_{d,L}(c) := \sum_{k=0}^{2L-1} \binom{2L-1}{k} \mu_{k+1} c^k = 0.$$

This is a degree-$2L - 1$ polynomial in $c$, and due to the structure of the $\chi_d^2$ moments and the positivity of the integrand for large negative $c$, there exists a unique real solution $c_{d,L}^* < 0$ which minimises $f_{d,L}$.

**Proposition B.3.4** (Strict convexity of the one–variable functional). *For every fixed dimension $d \geq 1$ and exponent $L \geq 1$, the map*

$$f_{d,L} : \mathbb{R} \longrightarrow \mathbb{R}, \qquad c \mapsto d - 1 + \mathbb{E}[(1 + cr)^{2L}], \qquad r \sim \chi_d^2,$$

*is strictly convex. Consequently, the critical point $c_{d,L}^*$ that solves $f'_{d,L}(c) = 0$ (or, equivalently, $P_{d,L}(c) = 0$) is the unique global minimiser.*

*Proof.* Differentiate under the integral sign to obtain

$$f'_{d,L}(c) = 2L\mathbb{E}\big[r(1+cr)^{2L-1}\big], \qquad f''_{d,L}(c) = 2L(2L-1)\mathbb{E}\big[r^2(1+cr)^{2L-2}\big].$$

Because $r > 0$ almost surely and $2L(2L-1) > 0$, the integrand in $f''_{d,L}(c)$ is strictly positive for every $c \in \mathbb{R}$. Hence

$$f''_{d,L}(c) > 0 \quad \text{for all } c \in \mathbb{R},$$

which proves strict convexity of $f_{d,L}$. A strictly convex $C^1$ function possesses at most one critical point, and that point (if it exists) is its global minimiser. Since we have already shown the existence of a real root $c^*_{d,L} < 0$ of $f'_{d,L}$, this root is unique and yields

$$\min_{c \in \mathbb{R}} f_{d,L}(c) = f_{d,L}(c^*_{d,L}). \qquad \square$$

**Corollary B.3.5.** *The optimal diagonal matrix in the original problem is*

$$D^* = c^*_{d,L} I_d,$$

*and this choice is unique.*

## B.4 Supporting theoretical results

**Theorem B.4.1.** *Let $K \in \mathbb{R}^{d \times d}$ be a real skew-symmetric matrix with singular values $\sigma_1, \ldots, \sigma_d$, and let $\bar{\sigma} = \frac{1}{d} \sum_{i=1}^{d} \sigma_i$ be their mean. Define the isotropic skew-symmetric set*

$$\mathcal{E} = \big\{ E \in \mathbb{R}^{d \times d} : E^\top = -E \text{ and } s_1(E) = \cdots = s_d(E) \big\},$$

*where $s_i(E)$ are the singular values of $E$. Then*

$$\inf_{E \in \mathcal{E}} \|K - E\|_F = \sqrt{\sum_{i=1}^{d} (\sigma_i - \bar{\sigma})^2}.$$

*Proof.* Let $K = U\Sigma V^\top$ be a singular value decomposition of $K$, where $\Sigma = \text{diag}(\sigma_1, \ldots, \sigma_d)$ and $U, V$ are orthogonal matrices. By unitary invariance of the Frobenius norm, we have

$$\|K - E\|_F = \|\Sigma - U^\top E V\|_F$$

for any $E \in \mathcal{E}$. Let us define $X := U^\top E V$. Then $X$ is skew-symmetric with all singular values equal to some $\sigma' \geq 0$. In this rotated basis, we aim to minimize $\|\Sigma - X\|_F$. We expand the Frobenius norm:

$$\|\Sigma - X\|_F^2 = \|\Sigma\|_F^2 + \|X\|_F^2 - 2\text{tr}(\Sigma X^\top).$$

Now apply the von Neumann trace inequality, which states:

$$\text{tr}(\Sigma X^\top) \leq \sum_{i=1}^{d} \sigma_i s_i(X).$$

Since all singular values of $X$ are equal to $\sigma'$, this becomes

$$\text{tr}(\Sigma X^\top) \leq \sigma' \sum_{i=1}^{d} \sigma_i = d\bar{\sigma}\sigma'.$$

Substituting this back, we get:

$$\|\Sigma - X\|_F^2 \geq \sum_{i=1}^{d} \sigma_i^2 + d(\sigma')^2 - 2d\bar{\sigma}\sigma' = \sum_{i=1}^{d} (\sigma_i - \sigma')^2.$$

Equality is achieved when $X = \sigma' I_d$, i.e., when $E = U(\sigma' I_d)V^\top$, which is still skew-symmetric provided the correct block structure of $U, V$. Therefore,

$$\inf_{E \in \mathcal{E}} \|K - E\|_F = \min_{\sigma' \in \mathbb{R}} \sqrt{\sum_{i=1}^{d} (\sigma_i - \sigma')^2}.$$

This expression is minimised by $\sigma' = \bar{\sigma}$, the mean of the singular values. Hence,

$$\inf_{E \in \mathcal{E}} \|K - E\|_F = \sqrt{\sum_{i=1}^{d} (\sigma_i - \bar{\sigma})^2}.$$

$\square$

**Proposition B.4.2.** *Let $Z_0 \in \mathbb{R}^{(n+1) \times (d+1)}$ be any matrix. Let $Z_\ell$ for any $\ell \in \{1, \dots, L\}$ be the output of the $\ell$-th layer (as defined in Eq. (2.2)) of the linear transformer parametrised by $A_0, \dots, A_{L-1}$, under the sparsity structure of Definition 2.4.4. Let $X_\ell \in \mathbb{R}^{d \times (n+1)}$ denote the first $d$ rows of $Z_\ell$, and let $Y_\ell \in \mathbb{R}^{1 \times (n+1)}$ denote the last row, such that*

$$Z_\ell = \begin{bmatrix} X_\ell \\ Y_\ell \end{bmatrix} \quad \text{for all } \ell \in \{0, \dots, L\}.$$

*Then the following hold:*

$$X_\ell = X_0 \quad \text{for all } \ell \in \{1, \dots, L\}, \tag{B.3}$$

$$Y_{\ell+1} = Y_\ell - \frac{1}{n} Y_\ell M X_0^\top A_\ell X_0 \quad \text{for all } \ell \in \{0, \dots, L-1\}, \tag{B.4}$$

$$Y_L = Y_0 \prod_{\ell=0}^{L-1} \left( I_n - \frac{1}{n} M X_0^\top A_\ell X_0 \right). \tag{B.5}$$

*Proof.* We begin by simplifying the attention formula from Eq. (2.3). Let $X \in \mathbb{R}^{d \times (n+1)}$ and $Y \in \mathbb{R}^{1 \times (n+1)}$ such that

$$Z = \begin{bmatrix} X \\ Y \end{bmatrix}.$$

Then we compute

$$\text{Attn}_{P,Q}(Z) = PZM(Z^\top QZ) = \begin{bmatrix} 0_{d \times d} & 0 \\ 0 & 1 \end{bmatrix} \begin{bmatrix} X \\ Y \end{bmatrix} M \begin{bmatrix} X \\ Y \end{bmatrix}^\top \begin{bmatrix} -A & 0 \\ 0 & 0 \end{bmatrix} \begin{bmatrix} X \\ Y \end{bmatrix}$$

$$= -\begin{bmatrix} 0_{d \times (n+1)} \\ Y \end{bmatrix} M X^\top A X = \begin{bmatrix} 0_{d \times (n+1)} \\ -Y M X^\top A X \end{bmatrix}.$$

This implies that

$$\begin{bmatrix} X_{\ell+1} \\ Y_{\ell+1} \end{bmatrix} = Z_{\ell+1} = Z_\ell + \frac{1}{n} \text{Attn}_{P_\ell, Q_\ell}(Z_\ell) = \begin{bmatrix} X_\ell \\ Y_\ell \end{bmatrix} + \frac{1}{n} \begin{bmatrix} 0_{d \times (n+1)} \\ -Y_\ell M X_\ell^\top A_\ell X_\ell \end{bmatrix} = \begin{bmatrix} X_\ell \\ Y_\ell - \frac{1}{n} Y_\ell M X_\ell^\top A_\ell X_\ell \end{bmatrix}.$$

Hence, we have $X_{\ell+1} = X_\ell$ for all $\ell$, which implies

$$X_1 = \dots = X_L = X_0.$$

Substituting this into the recurrence relation for $Y_\ell$, we obtain

$$Y_{\ell+1} = Y_\ell - \frac{1}{n} Y_\ell M X_0^\top A_\ell X_0 = Y_\ell \left( I_n - \frac{1}{n} M X_0^\top A_\ell X_0 \right).$$

Applying this recursively gives the closed-form expression

$$Y_L = Y_0 \prod_{\ell=0}^{L-1} \left( I_n - \frac{1}{n} M X_0^\top A_\ell X_0 \right),$$

which completes the proof. $\square$

**Corollary B.4.3.** *In the weight tying scenario described in* Definition 2.4.7, *the expression in* Eq. (B.5) *further simplifies to*

$$Y_L = Y_0 \left( I_n - \frac{1}{n} M X_0^\top A X_0 \right)^L.$$

**Fact B.4.4** (Reformulation of in-context loss)*. Recall the definitions of $X, Y, Z_0$, and $y_{test}$ from* Definition 2.4.1. *Define*

$$\overline{Z}_0 := \begin{bmatrix} X \\ Y \end{bmatrix} \in \mathbb{R}^{(d+1) \times (n+1)},$$

*which is equal to $Z_0$ but without replacing the last entry (corresponding to $y^{(n+1)}$) with zero. Then, the in-context loss can be equivalently written as*

$$f_{(D_X, D_W)} = f_{(z_0, y_{test})} = f_{(\overline{Z}_{0,0})} = \mathbb{E}_{\overline{Z}_0} \left[ \left( TF_L \left( \overline{Z}_0; \{P_\ell, Q_\ell\}_{\ell=0}^{L-1} \right) \right)^2 \right]$$

$$= \mathbb{E} \left[ \left( \overline{Z}_L \right)_{(d+1),(n+1)}^2 \right] = \mathbb{E} \left[ \text{Tr} \left( (I - M) Y_L^\top Y_L (I - M) \right) \right],$$

*where $\overline{Z}_L$ is the output of the final transformer layer initialized with $\overline{Z}_0$ (as in* Eq. (2.2)*), $Y_L \in \mathbb{R}^{1 \times (n+1)}$ is its last row, and $M$ is the masking matrix defined in* Definition 2.4.2.

*Proof.* See [Ahn+23, Lemma 5]. □

**Lemma B.4.5.** *Let $M \in \mathbb{R}^{d \times d}$, and let $u, v \in \mathbb{R}^d$ be independent random vectors with $u, v \overset{iid}{\sim} \mathcal{N}(0, I)$. Then,*

$$\mathbb{E} \left[ \langle u, Mv \rangle^2 \right] = \|M\|_F^2,$$

*where $\|M\|_F$ denotes the Frobenius norm of $M$.*

*Proof.* We expand the inner product:

$$\langle u, Mv \rangle = \sum_{i=1}^d u_i (Mv)_i = \sum_{i=1}^d \sum_{j=1}^d u_i M_{ij} v_j.$$

Squaring the expression, we obtain:

$$\langle u, Mv \rangle^2 = \sum_{i,j,k,l=1}^d u_i M_{ij} v_j \cdot u_k M_{kl} v_l.$$

Taking the expectation and using the independence of $u$ and $v$, we get:

$$\mathbb{E}[u_i M_{ij} v_j \cdot u_k M_{kl} v_l] = M_{ij} M_{kl} \mathbb{E}[u_i u_k] \mathbb{E}[v_j v_l].$$

Since $u, v \sim \mathcal{N}(0, I)$, we have:

$$\mathbb{E}[u_i u_k] = \delta_{ik}, \qquad \mathbb{E}[v_j v_l] = \delta_{jl},$$

where $\delta_{ik}$ is the Kronecker delta. Substituting:

$$\mathbb{E} \left[ \langle u, Mv \rangle^2 \right] = \sum_{i,j=1}^d M_{ij}^2 = \|M\|_F^2. \qquad \square$$

## B.5 Deferred proofs

### B.5.1 Proof of Lemma 2.4.14

*Proof.* By Fact B.4.4 and Proposition B.4.2, we have

$$f_\Sigma(\{A_\ell\}) = \mathbb{E}\left[\text{Tr}\left((I - M)Y_L^\top Y_L(I - M)\right)\right],$$

with

$$Y_L = Y_0 \prod_{\ell=0}^{L-1}\left(I_n - \frac{1}{n}MX_0^\top A_\ell X_0\right),$$

where the data is generated as follows:

$$x^{(i)} \overset{\text{iid}}{\sim} \mathcal{N}(0, \Sigma), \quad w_* \sim \mathcal{N}(0, \Sigma^{-1}),$$

and

$$X_0 = \begin{bmatrix} x^{(1)} & \cdots & x^{(n+1)} \end{bmatrix} \in \mathbb{R}^{d \times (n+1)}, \quad Y_0 = w_*^\top X_0 \in \mathbb{R}^{1 \times (n+1)}.$$

Now define the transformed variables:

$$X_0' \coloneqq \mathbf{L}^{-1} X_0, \quad w_*' \coloneqq \mathbf{L}^\top w_*.$$

Then

$$x^{(i)\prime} = \mathbf{L}^{-1} x^{(i)} \sim \mathcal{N}(0, I),$$
$$w_*' \sim \mathcal{N}(0, I),$$
$$w_*'^\top X_0' = w_*^\top X_0 = Y_0,$$
$$X_0'^\top (\mathbf{L}^\top A_\ell \mathbf{L}) X_0' = X_0^\top A_\ell X_0.$$

Thus, the computation of $Y_L$ remains unchanged under the transformation, and we conclude:

$$f_\Sigma\left(\{A_\ell\}_{\ell=0}^{L-1}\right) = f_I\left(\{\mathbf{L}^\top A_\ell \mathbf{L}\}_{\ell=0}^{L-1}\right).$$

$\square$

### B.5.2 Proof of Lemma 2.4.9

*Proof.* From Eq. (B.3) and Eq. (B.4), we know that

$$X_\ell = X_0 \quad \text{and} \quad Y_{\ell+1} = Y_\ell - \frac{1}{n}Y_\ell M X_0^\top A_\ell X_0.$$

Let $\overline{X}_0 \in \mathbb{R}^{d \times n}$ be the first $n$ columns of $X_0$, and let $\overline{Y}_\ell \in \mathbb{R}^{1 \times n}$ be the first $n$ entries of $Y_\ell$. Then, the gradient of the loss in Eq. (2.6) is

$$\nabla R_{w_*}(w) = \frac{1}{n}\overline{X}_0\overline{X}_0^\top(w - w_*). \tag{B.6}$$

We prove Eq. (2.7) by induction. The base case $\ell = 0$ is immediate from Eq. (2.4):

$$Y_0 = Y_0 - 0^\top X_0.$$

Assume, as the inductive hypothesis, that

$$Y_\ell = Y_0 - \left(w_{\text{gd}}^{(\ell)}\right)^\top X_0.$$

Restricting to the first $n$ entries gives

$$\overline{Y}_\ell = \overline{Y}_0 - \left(w_{\mathrm{gd}}^{(\ell)}\right)^\top \overline{X}_0.$$

Then

$$
\begin{aligned}
Y_{\ell+1} &= Y_\ell - \frac{1}{n} Y_\ell M X_0^\top A_\ell X_0 \\
&= Y_\ell - \frac{1}{n} \overline{Y}_\ell \overline{X}_0^\top A_\ell X_0 \\
&= Y_0 - \left(w_{\mathrm{gd}}^{(\ell)}\right)^\top X_0 - \frac{1}{n} \left(\overline{Y}_0 - \left(w_{\mathrm{gd}}^{(\ell)}\right)^\top \overline{X}_0\right) \overline{X}_0^\top A_\ell X_0 \\
&= Y_0 - \left[\left(w_{\mathrm{gd}}^{(\ell)}\right)^\top + \frac{1}{n}\left(\overline{Y}_0 - \left(w_{\mathrm{gd}}^{(\ell)}\right)^\top \overline{X}_0\right) \overline{X}_0^\top A_\ell\right] X_0 \\
&= Y_0 - \left[\left(w_{\mathrm{gd}}^{(\ell)}\right)^\top + \frac{1}{n}\left(w_*^\top \overline{X}_0 - \left(w_{\mathrm{gd}}^{(\ell)}\right)^\top \overline{X}_0\right) \overline{X}_0^\top A_\ell\right] X_0 \\
&= Y_0 - \left[\left(w_{\mathrm{gd}}^{(\ell)}\right)^\top + \frac{1}{n}\left(w_* - w_{\mathrm{gd}}^{(\ell)}\right)^\top \overline{X}_0 \overline{X}_0^\top A_\ell\right] X_0 \\
&= Y_0 - \left[\left(w_{\mathrm{gd}}^{(\ell)}\right)^\top + \nabla R_{w_*}\left(w_{\mathrm{gd}}^{(\ell)}\right)^\top A_\ell\right] X_0 \\
&= Y_0 - \left[w_{\mathrm{gd}}^{(\ell)} - A_\ell^\top \nabla R_{w_*}\left(w_{\mathrm{gd}}^{(\ell)}\right)\right]^\top X_0 \\
&= Y_0 - \left(w_{\mathrm{gd}}^{(\ell+1)}\right)^\top X_0.
\end{aligned}
$$

This completes the inductive step. Finally, we compute the transformer output as

$$\mathrm{TF}_L\left(Z_0; \{A_\ell\}_{\ell=0}^{L-1}\right) = -\left[Y_L\right]_{n+1} = -\left[Y_0 - \left(w_{\mathrm{gd}}^{(L)}\right)^\top X_0\right]_{n+1} = \left\langle w_{\mathrm{gd}}^{(L)}, x^{(n+1)}\right\rangle,$$

which concludes the proof. □

# C Appendix: model collapse

## C.1 Collapse of $\chi^2$ random variables

**Theorem C.1.1.** *Let $X_i \sim \chi^2_{k_i}$ be independent with $k_i \in \mathbb{N}$ and set*

$$Y_n = \prod_{i=1}^{n} \frac{X_i}{k_i}, \quad n \geq 1.$$

*Then $Y_n$ converges almost surely and*

$$Y_n \xrightarrow[n\to\infty]{a.s.} 0 \quad \Longleftrightarrow \quad \sum_{i=1}^{\infty} \frac{1}{k_i} = \infty.$$

**Almost sure convergence** For the natural filtration $\mathcal{F}_n := \sigma(X_1, \ldots, X_n)$ we record:

**Proposition C.1.2.** *The sequence $(Y_n)_{n \geq 1}$ is a nonnegative martingale with respect to $(\mathcal{F}_n)$, and therefore converges almost surely to a finite limit $Y_\infty \geq 0$. Furthermore $\mathbb{E}[Y_n] = 1$ for all $n \in \mathbb{N}$.*

*Proof.* Each $X_i$ is almost surely positive and $k_i > 0$, so $Y_n > 0$ almost surely. To check the martingale property:

$$\mathbb{E}\left[\frac{X_{n+1}}{k_{n+1}} \bigg| \mathcal{F}_n\right] = \frac{1}{k_{n+1}} \mathbb{E}[X_{n+1}] = 1.$$

Hence,

$$\mathbb{E}[Y_{n+1} \mid \mathcal{F}_n] = Y_n \cdot \mathbb{E}\left[\frac{X_{n+1}}{k_{n+1}} \bigg| \mathcal{F}_n\right] = Y_n.$$

Thus, by the martingale convergence theorem (Theorem C.2.3), $Y_n \to Y_\infty$ a.s., for some $Y_\infty \geq 0$. Since $Y_n$ is a martingale, we have $\mathbb{E}[Y_n] = \mathbb{E}[Y_1] = 1$ for all $n \in \mathbb{N}$. $\square$

**Forward direction** We prove the forward implication in Theorem C.1.1:

$$Y_n \xrightarrow[n\to\infty]{a.s.} 0 \quad \Longrightarrow \quad \sum_{i=1}^{\infty} \frac{1}{k_i} = \infty.$$

*Proof.* We proceed by contraposition. Assume

$$\sum_{i=1}^{\infty} \frac{1}{k_i} < \infty.$$

We will show that $(Y_n)$ is bounded in $L^2$. By independence of the $X_i$,

$$\mathbb{E}[Y_n^2] = \prod_{i=1}^{n} \mathbb{E}\left[\left(\frac{X_i}{k_i}\right)^2\right].$$

Since $\mathbb{E}[X_i^2] = 2k_i + k_i^2$, we get:

$$\mathbb{E}\left[\left(\frac{X_i}{k_i}\right)^2\right] = 1 + \frac{2}{k_i},$$

so:

$$\mathbb{E}[Y_n^2] = \prod_{i=1}^{n}\left(1 + \frac{2}{k_i}\right).$$

As $\sum \frac{1}{k_i} < \infty$, the product converges by Lemma C.2.5 to a finite limit, hence $(Y_n)$ is bounded in $L^2$. By Theorem C.2.4, $Y_n \to Y_\infty$ in $L^2$. Since convergence in $L^2$ implies convergence in $L^1$, we have:

$$\mathbb{E}[Y_\infty] = \lim_{n\to\infty} \mathbb{E}[Y_n] = 1.$$

Since $Y_\infty \geq 0$, it follows that $\mathbb{P}(Y_\infty = 0) < 1$. $\qquad\square$

**Backward direction**   To prove the converse direction, we first bound the expectation and variance of $\log(X/k)$ for $X \sim \chi_k^2$. Throughout, $\psi$ and $\psi_1$ denote the digamma and trigamma functions, respectively.

**Lemma C.1.3.** *For $X \sim \chi_k^2$ one has*

$$\mathbb{E}\left[\log\left(\frac{X}{k}\right)\right] < -\frac{1}{k}.$$

*Proof.* We have, as given in [AS64, Eq. 26.4.36],

$$\mathbb{E}\left[\log\left(\frac{X}{k}\right)\right] = \psi\left(\frac{k}{2}\right) - \log\left(\frac{k}{2}\right).$$

Now, using the inequality for the digamma function [Alz97, Eq. 2.2]:

$$\psi(x) < \log x - \frac{1}{2x}, \quad \text{for } x > 0,$$

and applying it with $x = \frac{k}{2}$, we obtain:

$$\psi\left(\frac{k}{2}\right) < \log\left(\frac{k}{2}\right) - \frac{1}{k}.$$

Therefore,

$$\mathbb{E}\left[\log\left(\frac{X}{k}\right)\right] < \log\left(\frac{k}{2}\right) - \frac{1}{k} - \log\left(\frac{k}{2}\right) = -\frac{1}{k}. \qquad\square$$

**Lemma C.1.4** (Variance log bound). *For $X \sim \chi_k^2$ one has*

$$\mathrm{Var}\left(\log\left(\frac{X}{k}\right)\right) < \frac{6}{k}.$$

*Consequently,*

$$\sum_{i=1}^{\infty} \mathrm{Var}\left(\frac{\log(X_i/k_i)}{S_i}\right) < \infty, \quad S_i := \sum_{j=1}^{i}\frac{1}{k_j}.$$

*Proof.* We have, as given in [AS64, Eq. 26.4.36],

$$\mathrm{Var}\left(\log\left(\frac{X}{k}\right)\right) = \psi_1\left(\frac{k}{2}\right),$$

where $\psi_1$ denotes the trigamma function. Moreover, for all $x > 0$, the inequality

$$\psi_1(x) < \frac{1}{x} + \frac{1}{x^2}$$

holds as shown in [QGG10, Lemma 3]. Applying this with $x = \frac{k}{2}$, we get

$$\mathrm{Var}\left(\log\left(\frac{X}{k}\right)\right) = \psi_1\left(\frac{k}{2}\right) < \frac{2}{k} + \frac{4}{k^2} \leq \frac{6}{k},$$

since $k \in \mathbb{N}$. Finally, since $S_i - S_{i-1} = \frac{1}{k_i}$ and $S_i \geq S_{i-1}$, for all $i \geq 2$

$$\mathrm{Var}\left(\frac{\log(X_i/k_i)}{S_i}\right) = \frac{\mathrm{Var}(\log(X_i/k_i))}{S_i^2} < \frac{6}{k_i S_i^2} = 6 \cdot \frac{S_i - S_{i-1}}{S_i^2} \leq 6 \cdot \frac{S_i - S_{i-1}}{S_i S_{i-1}} \leq 6\left(\frac{1}{S_{i-1}} - \frac{1}{S_i}\right).$$

And after telescoping,

$$\sum_{i=2}^{\infty} \frac{\mathrm{Var}(\log(X_i/k_i))}{S_i^2} \leq 6\left(\frac{1}{S_1} - \frac{1}{S_\infty}\right) \leq \frac{6}{S_1} = 6k_1.$$

Since $\frac{\mathrm{Var}(\log(X_1/k_1))}{S_1^2}$ is also finite, we get the result. $\qquad\qquad\square$

With this, we can show the converse direction of Theorem C.1.1, so

$$Y_n \xrightarrow[n\to\infty]{\text{a.s.}} 0 \quad \Longleftarrow \quad \sum_{i=1}^{\infty} \frac{1}{k_i} = \infty.$$

*Proof.* Set

$$Z_i = \log\left(\frac{X_i}{k_i}\right) - \mathbb{E}\left[\log\left(\frac{X_i}{k_i}\right)\right],$$

$$S_n = \sum_{i=1}^{n} \frac{1}{k_i},$$

$$T_n = \sum_{i=1}^{n} Z_i = \log Y_n - \mathbb{E}[\log Y_n]$$

and assume $\sum_{i=1}^{\infty} \frac{1}{k_i} = \infty$, which implies $S_n \to \infty$ as $n \to \infty$. By Lemma C.1.4

$$\sum_{i=1}^{\infty} \mathrm{Var}\left(\frac{\log(X_i/k_i)}{S_i}\right) < \infty,$$

hence by Theorem C.2.6

$$\sum_{i=1}^{\infty} \left(\frac{\log(X_i/k_i)}{S_i} - \mathbb{E}\left[\frac{\log(X_i/k_i)}{S_i}\right]\right) = \sum_{i=1}^{\infty} \frac{Z_i}{S_i}$$

converges almost surely. By the Random Kronecker Lemma (Lemma C.2.7) we get that

$$\frac{1}{S_n} \sum_{i=1}^{n} Z_i = \frac{T_n}{S_n} \xrightarrow[n\to\infty]{\text{a.s.}} 0.$$

In Lemma C.1.3 we obtained $\mathbb{E}[\log(X_i/k_i)] \leq -\frac{1}{k_i}$; hence

$$\mathbb{E}[\log Y_n] = \sum_{i=1}^{n} \mathbb{E}\left[\log\left(\frac{X_i}{k_i}\right)\right] \leq -S_n \xrightarrow[n\to\infty]{} -\infty.$$

Combining this with the fact that $T_n/S_n \to 0$ a.s., we get

$$\log Y_n = \mathbb{E}[\log Y_n] + T_n \leq -S_n\left(1 - \frac{T_n}{S_n}\right) \xrightarrow[n\to\infty]{\text{a.s.}} -\infty.$$

Exponentiating, we conclude $Y_n \xrightarrow{\text{a.s.}} 0$. $\qquad\qquad\square$

**Corollary C.1.5** (Uniform exponential decay). *Assume*

$$\sum_{i=1}^{\infty} \frac{1}{k_i} = \infty, \qquad S_n := \sum_{i=1}^{n} \frac{1}{k_i}.$$

*Then there exists an almost−surely finite random constant $C = C(\omega)$ such that*

$$Y_n(\omega) \leq C(\omega) e^{-S_n/2} \qquad \text{for every } n \in \mathbb{N}.$$

*Proof.* From the backwards direction proof, we have

$$\log Y_n = \sum_{i=1}^{n} \log\left(\frac{X_i}{k_i}\right) \leq -S_n + T_n,$$

where

$$T_n := \sum_{i=1}^{n} Z_i, \quad Z_i := \log\left(\frac{X_i}{k_i}\right) - \mathbb{E}\left[\log\left(\frac{X_i}{k_i}\right)\right].$$

We have shown that $T_n/S_n \to 0$ almost surely, so for each $\omega$ in a full-measure set $\Omega_0$, there exists an index $n_0(\omega)$ such that

$$|T_n(\omega)| \leq \frac{S_n}{2} \qquad \text{for all } n \geq n_0(\omega).$$

Therefore, for those $n$,

$$\log Y_n(\omega) \leq -S_n + \frac{S_n}{2} = -\frac{S_n}{2}, \qquad \text{so } Y_n(\omega) \leq e^{-S_n/2}.$$

For the remaining finitely many indices $1 \leq n < n_0(\omega)$, define

$$C(\omega) := \max\{1, \max_{1 \leq n < n_0(\omega)} Y_n(\omega) e^{S_n/2}\}.$$

Then $C(\omega) < \infty$ almost surely, and for all $n \in \mathbb{N}$,

$$Y_n(\omega) \leq C(\omega) e^{-S_n/2}. \qquad \square$$

## C.2 Supporting theoretical results

**Fact C.2.1.** *Let $\mu(\Omega) < \infty$ and $p > 1$. If $\mathcal{F}$ is bounded in $L^p(\mu)$, then $\mathcal{F}$ is uniformly integrable.*

*Proof.* Can be found in [Kle07, Corollary 6.21] $\qquad \square$

**Theorem C.2.2** (Convergence theorem for uniformly integrable martingales). *Let $(X_n)_{n \in \mathbb{N}_0}$ be a uniformly integrable $\mathcal{F}$ - (sub-, super-) martingale. Then there exists an $\mathcal{F}_\infty$-measurable integrable random variable $X_\infty$ with*

$$X_n \xrightarrow[n \to \infty]{} X_\infty \quad \text{a.s. and in } L^1.$$

*Proof.* Can be found in [Kle07, Theorem 11.7] $\qquad \square$

**Theorem C.2.3.** *If $X$ is a nonnegative supermartingale, then there is an $\mathcal{F}_\infty$-measurable random variable $X_\infty \geq 0$ with $\mathbb{E}[X_\infty] \leq \mathbb{E}[X_0]$ and $X_n \xrightarrow[n \to \infty]{} X_\infty$ a.s.*

*Proof.* Can be found in [Kle07, Corollary 11.5]. $\qquad \square$

**Theorem C.2.4** ($L^p$-convergence theorem for martingales). *Let $p > 1$ and let $(X_n)_{n \in \mathbb{N}_0}$ be an $L^p$-bounded martingale. Then there exists an $\mathcal{F}_\infty$-measurable random variable $X_\infty$ with $\mathbb{E}[|X_\infty|^p] < \infty$ and*

$$X_n \xrightarrow[n \to \infty]{a.s.} X_\infty \quad \text{and in } L^p.$$

*In particular, $(|X_n|^p)_{n \in \mathbb{N}_0}$ is uniformly integrable.*

*Proof.* Can be found in [Kle07, Theorem 11.10]. □

**Lemma C.2.5.** *Let* $(a_j)_{j=1}^{\infty}$ *be a sequence of positive real numbers. Then*

$$\sum_{j=1}^{\infty} a_j < \infty \quad \Longleftrightarrow \quad \prod_{j=1}^{\infty} (1 + a_j) < \infty.$$

*Proof.* ($\Rightarrow$) Assume that $\sum_{j=1}^{\infty} a_j < \infty$. For all $a_j$, the inequality

$$\ln(1 + a_j) \leq a_j$$

holds. Thus,

$$\sum_{j=1}^{\infty} \ln(1 + a_j) \leq \sum_{j=1}^{\infty} a_j < \infty.$$

Hence, the sum $\sum_{j=1}^{\infty} \ln(1 + a_j)$ converges, and it follows that

$$\prod_{j=1}^{\infty} (1 + a_j) = \exp\left(\sum_{j=1}^{\infty} \ln(1 + a_j)\right) < \infty.$$

($\Leftarrow$) Now assume that $\sum_{j=1}^{\infty} a_j = \infty$. Consider the finite products

$$\prod_{j=1}^{n} (1 + a_j) = 1 + \sum_{k=1}^{n} a_k + \sum_{\substack{I \subset \{1,\dots,n\} \\ |I| \geq 2}} \prod_{j \in I} a_j \geq 1 + \sum_{k=1}^{n} a_k.$$

Since the right-hand side diverges to $\infty$, so does the product. Thus,

$$\prod_{j=1}^{\infty} (1 + a_j) = \infty.$$ □

**Theorem C.2.6** (The Kolmogorov convergence criterion). *Let* $X_1, X_2, \dots$ *be independent random variables. Then*

$$\sum_{n=1}^{\infty} \operatorname{Var} X_n < \infty \quad \Longrightarrow \quad \sum_{n=1}^{\infty} (X_n - \mathbb{E} X_n) \quad \text{converges a.s.}$$

*Proof.* Can be found in [Gut05, Chapter 6 Theorem 5.2]. □

**Lemma C.2.7** (Random Kronecker lemma). *Suppose that* $\{X_n\}_{n \geq 1}$ *are random variables, set* $a_0 = 0$, *and let* $\{a_n\}_{n \geq 1}$ *be positive numbers increasing to* $+\infty$. *If*

$$\sum_{k=1}^{\infty} \frac{X_k}{a_k} \quad \text{converges almost surely,}$$

*then*

$$\frac{1}{a_n} \sum_{k=1}^{n} X_k \longrightarrow 0 \quad \text{almost surely as} \quad n \to \infty.$$

*Proof.* Can be found in [Gut05, Chapter 6 Lemma 5.1]]. □

**Theorem C.2.8** (Matrix supermartingale convergence theorem). *Any* $S_d^+$*-valued supermartingale* $\{Y_n\}$ *converges almost surely to an* $S_d^+$*-valued random matrix* $Y_\infty$ *and* $\mathbb{E}[Y_\infty] \preceq \mathbb{E}[Y_0]$.

*Proof.* Can be found in [WR25]. □

**Lemma C.2.9.** *Let $X \sim \mathcal{N}_p(0, I)$ be a mean-zero multivariate Gaussian vector with identity covariance matrix. Then, for any symmetric matrix $A \in \mathbb{R}^{p \times p}$,*

$$\mathbb{E}[X^T A X] = \operatorname{tr}(A) \quad and \quad \operatorname{Var}(X^T A X) = 2 \operatorname{tr}(A^2).$$

*Proof.* Since $A$ is symmetric, it admits an orthogonal diagonalization $A = U \Lambda U^T$, where $U \in \mathbb{R}^{p \times p}$ is orthogonal and $\Lambda = \operatorname{diag}(\lambda_1, \ldots, \lambda_p)$ is diagonal with real eigenvalues. Define $Z = U^T X$. Since $U$ is orthogonal and $X \sim \mathcal{N}_p(0, I)$, it follows that $Z \sim \mathcal{N}_p(0, I)$ as well. Then,

$$X^T A X = (UZ)^T (U \Lambda U^T)(UZ) = Z^T \Lambda Z = \sum_{i=1}^{p} \lambda_i Z_i^2,$$

where the $Z_i$ are independent standard normal random variables. Now, since $\mathbb{E}[Z_i^2] = 1$ and $\operatorname{Var}(Z_i^2) = 2$ for each $i$, it follows that

$$\mathbb{E}[X^T A X] = \sum_{i=1}^{p} \lambda_i \mathbb{E}[Z_i^2] = \sum_{i=1}^{p} \lambda_i = \operatorname{tr}(A),$$

and

$$\operatorname{Var}(X^T A X) = \sum_{i=1}^{p} \lambda_i^2 \operatorname{Var}(Z_i^2) = 2 \sum_{i=1}^{p} \lambda_i^2 = 2 \operatorname{tr}(A^2). \qquad \square$$

**Corollary C.2.10.** *Let $X \sim \mathcal{N}_p(0, \Sigma)$ be a mean-zero multivariate Gaussian vector with covariance matrix $\Sigma \in \mathbb{R}^{p \times p}$ (symmetric and positive semidefinite). Then,*

$$\mathbb{E}[X^T X] = \operatorname{tr}(\Sigma) \quad and \quad \operatorname{Var}(X^T X) = 2 \operatorname{tr}(\Sigma^2).$$

**Theorem C.2.11** (Square variation of a martingale). *Let $X$ be a square-integrable $\mathbb{F}$-martingale. Then, for $n \in \mathbb{N}_0$,*

$$\langle X \rangle_n = \sum_{i=1}^{n} \mathbb{E}\left[(X_i - X_{i-1})^2 \big| \mathcal{F}_{i-1}\right].$$

*Proof.* Can be found in [Kle07, Theorem 10.4]. $\qquad \square$

**Theorem C.2.12.** *Let $(X_n)_{n \in \mathbb{N}}$ be a square-integrable martingale. If*

$$\sup_{n \in \mathbb{N}} \langle X \rangle_n < \infty \quad almost\ surely,$$

*then $X_n$ converges almost surely.*

*Proof.* Can be found in [Kle07, Theorem 11.14]. $\qquad \square$

**Corollary C.2.13** (Square variation criterion for martingale convergence). *Let $(X_n)_{n \in \mathbb{N}_0}$ be a square-integrable martingale with respect to a filtration $(\mathcal{F}_n)$, and suppose that*

$$\sum_{n=1}^{\infty} \operatorname{Var}(X_n \mid \mathcal{F}_{n-1}) < \infty \quad almost\ surely.$$

*Then $X_n$ converges almost surely.*

*Proof.* By Theorem C.2.11, the square variation process $\langle X \rangle_n$ is given by

$$\langle X \rangle_n = \sum_{i=1}^{n} \mathbb{E}[(X_i - X_{i-1})^2 \mid \mathcal{F}_{i-1}] = \sum_{i=1}^{n} \operatorname{Var}(X_i \mid \mathcal{F}_{i-1}).$$

The assumption implies that

$$\langle X \rangle_\infty := \lim_{n \to \infty} \langle X \rangle_n = \sum_{n=1}^{\infty} \operatorname{Var}(X_n \mid \mathcal{F}_{n-1}) < \infty \quad almost\ surely.$$

Thus, $\sup_n \langle X \rangle_n < \infty$ almost surely. By Theorem C.2.12, it follows that $X_n$ converges almost surely. $\qquad \square$

**Lemma C.2.14.** *Let $A \in \mathbb{R}^{d \times d}$ be a symmetric positive semidefinite matrix. Then*

$$\text{tr}(A^2) \leq \left(\text{tr}(A)\right)^2.$$

*Proof.* Since $A \succeq 0$, it is symmetric and thus orthogonally diagonalizable. That is, there exists an orthogonal matrix $Q \in \mathbb{R}^{d \times d}$ and nonnegative eigenvalues $\lambda_1, \ldots, \lambda_d \geq 0$ such that

$$A = Q \, \text{diag}(\lambda_1, \ldots, \lambda_d) Q^\top.$$

Using the invariance of the trace under similarity transformations, we compute:

$$\text{tr}(A) = \sum_{i=1}^{d} \lambda_i, \qquad \text{tr}(A^2) = \sum_{i=1}^{d} \lambda_i^2.$$

Expanding the square of the trace, we have

$$\left(\text{tr}(A)\right)^2 = \left(\sum_{i=1}^{d} \lambda_i\right)^2 = \sum_{i=1}^{d} \lambda_i^2 + 2 \sum_{1 \leq i < j \leq d} \lambda_i \lambda_j.$$

Since each $\lambda_i \geq 0$, it follows that the cross terms $\lambda_i \lambda_j \geq 0$. Hence,

$$\text{tr}(A^2) = \sum_{i=1}^{d} \lambda_i^2 \leq \left(\text{tr}(A)\right)^2.$$

Equality holds if and only if all cross terms vanish, i.e., $\lambda_i \lambda_j = 0$ for all $i \neq j$, which occurs precisely when at most one eigenvalue is nonzero. Therefore, equality holds if and only if $\text{rank}(A) \leq 1$. □

## C.3 Definitions of model collapse

Schaeffer et al. [Sch+25] identified 8 different definitions of model collapse.

1. **Catastrophic Increase of Population Risk**: Perhaps the most colloquial definition, model collapse is a critical and rapid degradation in model performance due to the presence of synthetic data, as measured by population risk. We note that what constitutes catastrophic is often undefined.

2. **Any Increase of Population Risk**: Under this strict definition, model collapse occurs if there is *any* increase in population risk when training with synthetic data compared to training with real data alone.

3. **Asymptotically Diverging Population Risk**: This definition considers model collapse to occur when the population risk grows without bound over successive model-fitting iterations. This represents a fundamentally unstable learning dynamic where each iteration of synthetic data generation and training leads to progressively worse performance.

4. **Collapsing Variance**: Model collapse here is when variance (or diversity) trends towards 0 and the learned distributions tend towards delta-like functions over successive model-fitting iterations.

5. **Change in Scaling Law**: In this view, model collapse occurs if the governing scaling behaviour changes due to the presence of synthetic data. Specifically, model collapse occurs if the relationship between model performance and training data size deviates from the expected scaling behaviour observed with real data.

6. **Disappearance of or Entanglement of Real Data Mode(s)**: Sometimes called "Mode Collapse", model collapse here is defined by the presence of synthetic data preventing the model from learning particular modes of the real data distribution or causing the model to blur different data modes together.

7. **Disappearance of Real Tail Data**: Sometimes called "coverage collapse", model collapse here occurs when synthetic data leads to the under-representation of data from the tail of the distribution, leading to models that can only handle common cases but fail on rare ones. The disappearance of real tail data can be more subtle and more narrow than the generative model losing all diversity (Item 4).

8. **Appearance of Hallucinated Data**: Model collapse occurs when the sequence of models begins producing fully-synthetic data not supported by the original real data's distribution.

# Bibliography

[AS64]     M. Abramowitz and I. A. Stegun. *Handbook of Mathematical Functions with Formulas, Graphs, and Mathematical Tables.* Dover, 1964. ISBN: 0486612724.

[AW25]     P. Aggarwal and S. Welleck. *L1: Controlling How Long A Reasoning Model Thinks With Reinforcement Learning.* 2025. arXiv: 2503.04697 [cs.CL].

[Ahn+24]   K. Ahn et al. *Linear attention is (maybe) all you need (to understand transformer optimization).* 2024. arXiv: 2310.01082 [cs.LG].

[Ahn+23]   K. Ahn et al. *Transformers learn to implement preconditioned gradient descent for in-context learning.* 2023. arXiv: 2306.00297 [cs.LG].

[Aky+23]   E. Akyürek et al. *What learning algorithm is in-context learning? Investigations with linear models.* 2023. arXiv: 2211.15661 [cs.LG].

[Ale+23]   S. Alemohammad et al. *Self-Consuming Generative Models Go MAD.* 2023. arXiv: 2307.01850 [cs.LG].

[Alz97]    H. Alzer. "On some inequalities for the gamma and psi functions". In: *Mathematics of Computation* 66.217 (1997), pp. 373–389.

[BCB16]    D. Bahdanau, K. Cho, and Y. Bengio. *Neural Machine Translation by Jointly Learning to Align and Translate.* 2016. arXiv: 1409.0473 [cs.CL].

[BLW23]    H. H. Bauschke, M. K. Lal, and X. Wang. *Real roots of real cubics and optimization.* 2023. arXiv: 2302.10731 [math.OC].

[Ben+03]   Y. Bengio et al. "A Neural Probabilistic Language Model". In: *Journal of Machine Learning Research* 3 (2003), pp. 1137–1155.

[Bor18]    F. Bornemann. *Numerische lineare Algebra: eine konzise Einführung mit MATLAB und Julia.* Springer-Verlag, 2018.

[Bro+20]   T. B. Brown et al. *Language Models are Few-Shot Learners.* 2020. arXiv: 2005.14165 [cs.CL].

[Che+25]   X. Chen et al. *Do NOT Think That Much for 2+3=? On the Overthinking of o1-Like LLMs.* 2025. arXiv: 2412.21187 [cs.CL].

[CCS24]    X. Cheng, Y. Chen, and S. Sra. *Transformers Implement Functional Gradient Descent to Learn Non-Linear Functions In Context.* 2024. arXiv: 2312.06528 [cs.LG].

[Cob+21]   K. Cobbe et al. *Training Verifiers to Solve Math Word Problems.* 2021. arXiv: 2110.14168 [cs.LG].

[Dat+24]   S. Dathathri et al. "Scalable watermarking for identifying large language model outputs". In: *Nature* 634.8035 (2024), pp. 818–823.

[DA+25]    DeepSeek-AI et al. *DeepSeek-R1: Incentivizing Reasoning Capability in LLMs via Reinforcement Learning.* 2025. arXiv: 2501.12948 [cs.CL].

[Dev+19]   J. Devlin et al. *BERT: Pre-training of Deep Bidirectional Transformers for Language Understanding.* 2019. arXiv: 1810.04805 [cs.CL].

[DFK24]    E. Dohmatob, Y. Feng, and J. Kempe. *Model Collapse Demystified: The Case of Regression.* 2024. arXiv: 2402.07712 [cs.LG].

[Doh+24a]  E. Dohmatob et al. *A Tale of Tails: Model Collapse as a Change of Scaling Laws.* 2024. arXiv: 2402.07043 [cs.LG].

[Doh+24b]   E. Dohmatob et al. *Strong Model Collapse*. 2024. arXiv: 2410.04840 [cs.LG].

[Don+24]    Q. Dong et al. *A Survey on In-context Learning*. 2024. arXiv: 2301.00234 [cs.CL].

[DYL25]     G. Drayson, E. Yilmaz, and V. Lampos. *Machine-generated text detection prevents language model collapse*. 2025. arXiv: 2502.15654 [cs.CL].

[Dur19]     R. Durrett. *Probability: theory and examples*. Vol. 49. Cambridge university press, 2019.

[Elh+21]    N. Elhage et al. *A Mathematical Framework for Transformer Circuits*. Tech. rep. Anthropic, Dec. 2021.

[Elm90]     J. L. Elman. "Finding structure in time". In: *Cognitive Science* 14.2 (1990), pp. 179–211. ISSN: 0364-0213.

[Fan+25]    Y. Fan et al. *Looped Transformers for Length Generalization*. 2025. arXiv: 2409.15647 [cs.LG].

[Fat+25]    M. Fatemi et al. *Concise Reasoning via Reinforcement Learning*. 2025. arXiv: 2504.05185 [cs.CL].

[Fen+24]    Y. Feng et al. *Beyond Model Collapse: Scaling Up with Synthesized Data Requires Verification*. 2024. arXiv: 2406.07515 [cs.LG].

[Fut+21]    F. Futami et al. "Skew-symmetrically perturbed gradient flow for convex optimization". In: *Proceedings of The 13th Asian Conference on Machine Learning*. Ed. by V. N. Balasubramanian and I. Tsang. Vol. 157. Proceedings of Machine Learning Research. PMLR, Nov. 2021, pp. 721–736.

[Gag94]     P. Gage. "A new algorithm for data compression". In: *C Users J.* 12.2 (Feb. 1994), 23–38. ISSN: 0898-9788.

[Gar+23]    S. Garg et al. *What Can Transformers Learn In-Context? A Case Study of Simple Function Classes*. 2023. arXiv: 2208.01066 [cs.CL].

[Gat+24]    K. Gatmiry et al. *Can Looped Transformers Learn to Implement Multi-step Gradient Descent for In-context Learning?* 2024. arXiv: 2410.08292 [cs.LG].

[Ger+24]    M. Gerstgrasser et al. *Is Model Collapse Inevitable? Breaking the Curse of Recursion by Accumulating Real and Synthetic Data*. 2024. arXiv: 2404.01413 [cs.LG].

[Gia+23]    A. Giannou et al. *Looped Transformers as Programmable Computers*. 2023. arXiv: 2301.13196 [cs.LG].

[GN18]      A. K. Gupta and D. K. Nagar. *Matrix variate distributions*. Chapman and Hall/CRC, 2018.

[Gut05]     A. Gut. *Probability: A Graduate Course*. Springer, 2005.

[Has+25]    M. Hassid et al. *Don't Overthink it. Preferring Shorter Thinking Chains for Improved LLM Reasoning*. 2025. arXiv: 2505.17813 [cs.CL].

[HS97]      S. Hochreiter and J. Schmidhuber. "Long Short-Term Memory". In: *Neural Computation* 9.8 (1997), pp. 1735–1780.

[Kat87]     S. M. Katz. "Estimation of Probabilities from Sparse Data for the Language Model Component of a Speech Recognizer". In: *IEEE Transactions on Acoustics, Speech, and Signal Processing* 35.3 (1987), pp. 400–401.

[Kaz+25]    J. Kazdan et al. *Collapse or Thrive? Perils and Promises of Synthetic Data in a Self-Generating World*. 2025. arXiv: 2410.16713 [cs.LG].

[Kle07]     A. Klenke. *Probability Theory: A Comprehensive Course*. Springer, 2007.

[Lan87]     S. Lang. *Linear algebra*. Springer Science & Business Media, 1987.

[Luo+25]    H. Luo et al. *O1-Pruner: Length-Harmonizing Fine-Tuning for O1-Like Reasoning Pruning*. 2025. arXiv: 2501.12570 [cs.CL].

[MHM23]   A. Mahankali, T. B. Hashimoto, and T. Ma. *One Step of Gradient Descent is Provably the Optimal In-Context Learner with One Layer of Linear Self-Attention*. 2023. arXiv: 2307.03576 [cs.LG].

[Mik+10]   T. Mikolov et al. "Recurrent Neural Network Based Language Model". In: *Interspeech*. 2010, pp. 1045–1048.

[Min+22]   S. Min et al. *Rethinking the Role of Demonstrations: What Makes In-Context Learning Work?* 2022. arXiv: 2202.12837 [cs.CL].

[Mui05]    R. J. Muirhead. *Aspects of Multivariate Statistical Theory*. Wiley-Interscience, 2005.

[Osw+23]   J. von Oswald et al. *Transformers learn in-context by gradient descent*. 2023. arXiv: 2212.07677 [cs.LG].

[Pen+25]   K. Peng et al. *Revisiting Overthinking in Long Chain-of-Thought from the Perspective of Self-Doubt*. 2025. arXiv: 2505.23480 [cs.CL].

[QGG10]    F. Qi, S. Guo, and B.-N. Guo. "Complete monotonicity of some functions involving polygamma functions". In: *Journal of Computational and Applied Mathematics* 233.9 (Mar. 2010), 2149–2160. ISSN: 0377-0427.

[Rad+18]   A. Radford et al. *Improving Language Understanding by Generative Pre-Training*. Tech. rep. https://cdn.openai.com/research-covers/language-unsupervised/language_understanding_paper.pdf. OpenAI, 2018.

[Sch+25]   R. Schaeffer et al. *Position: Model Collapse Does Not Mean What You Think*. 2025. arXiv: 2503.03150 [cs.LG].

[Shu+24a]  I. Shumailov et al. "AI models collapse when trained on recursively generated data". In: *Nature* 631.8022 (2024), pp. 755–759.

[Shu+24b]  I. Shumailov et al. *The Curse of Recursion: Training on Generated Data Makes Models Forget*. 2024. arXiv: 2305.17493 [cs.LG].

[TP25]     W. L. Tong and C. Pehlevan. *MLPs Learn In-Context on Regression and Classification Tasks*. 2025. arXiv: 2405.15618 [cs.LG].

[Vas+23]   A. Vaswani et al. *Attention Is All You Need*. 2023. arXiv: 1706.03762 [cs.CL].

[WR25]     H. Wang and A. Ramdas. *Positive Semidefinite Matrix Supermartingales*. 2025. arXiv: 2401.15567 [math.PR].

[Wei+23]   J. Wei et al. *Chain-of-Thought Prompting Elicits Reasoning in Large Language Models*. 2023. arXiv: 2201.11903 [cs.CL].

[Wik25a]   Wikipedia contributors. *Cubic equation — Discriminant and nature of the roots*. Accessed: 2025-07-15. 2025.

[Wik25b]   Wikipedia contributors. *Quadratic equation — Discriminant*. Accessed: 2025-07-15. 2025.

[Wik25c]   Wikipedia contributors. *Second partial derivative test*. Accessed: 2025-07-15. 2025.

[Wu+25]    Y. Wu et al. *When More is Less: Understanding Chain-of-Thought Length in LLMs*. 2025. arXiv: 2502.07266 [cs.AI].

[Xie+22]   S. M. Xie et al. *An Explanation of In-context Learning as Implicit Bayesian Inference*. 2022. arXiv: 2111.02080 [cs.CL].

[Yan+24]   L. Yang et al. *Looped Transformers are Better at Learning Learning Algorithms*. 2024. arXiv: 2311.12424 [cs.LG].

[YWL25]    J. Yi, J. Wang, and S. Li. *ShorterBetter: Guiding Reasoning Models to Find Optimal Inference Length for Efficient Reasoning*. 2025. arXiv: 2504.21370 [cs.AI].

[ZFB23]    R. Zhang, S. Frei, and P. L. Bartlett. *Trained Transformers Learn Linear Models In-Context*. 2023. arXiv: 2306.09927 [stat.ML].

[Zha+23]   Y. Zhang et al. *What and How does In-Context Learning Learn? Bayesian Model Averaging, Parameterization, and Generalization.* 2023. arXiv: 2305.19420 [stat.ML].

[Zhu+25]   X. Zhu et al. *How to Synthesize Text Data without Model Collapse?* 2025. arXiv: 2412.14689 [cs.CL].